
\pdfoutput=1  

\documentclass[runningheads]{llncs}

\let\arxivbegin\begin
\let\arxivdocumentclass\documentclass
\def\begin#1{\endinput}
\def\documentclass[#1]#2{}

\title{Counting Contributions, Not Views:\texorpdfstring{\\}{ }Occlusion-Aware Densification\texorpdfstring{\\}{ }for 3D Gaussian Splatting} 

\titlerunning{Counting Contributions, Not Views}

\author{Marcel Rogge\inst{1,2} \and
René Schuster\inst{1,2} \and
Didier Stricker\inst{1,2}}

\authorrunning{M.~Rogge et al.}

\institute{University Kaiserslautern-Landau -- RPTU \and
German Research Center for Artificial Intelligence -- DFKI
}

\maketitle

%

\begin{abstract}
  3D Gaussian Splatting grows a dense scene from a sparse initialization through adaptive density control (ADC), driven by per-Gaussian statistics gathered during rendering. 
  These statistics pass a frustum-based visibility filter blind to occlusion: fully occluded Gaussians still count as observed, diluting the averaged gradient signal and suppressing densification precisely where geometry is occluded. 
  We correct this with an occlusion-aware filter that restricts densification statistics to views in which a Gaussian actually contributes, and replace the criterion's raw view-count normalizer with an average-contribution one. 
  Integrated into a StopThePop-based framework, the two together improve every image-quality metric over the occlusion-blind baseline on occlusion-heavy scenes (up to $6\%$ lower LPIPS), at less than half the Gaussians a view-count normalizer requires.
  Under the longer schedules these large scenes require, the count stays bounded where most criteria diverge, and on smaller benchmarks our method stays competitive at a fraction of the competitors' count. 
\end{abstract}

\begin{center}
    \begin{minipage}[c]{0.32\linewidth}
        \centering
        \includegraphics[width=\linewidth]{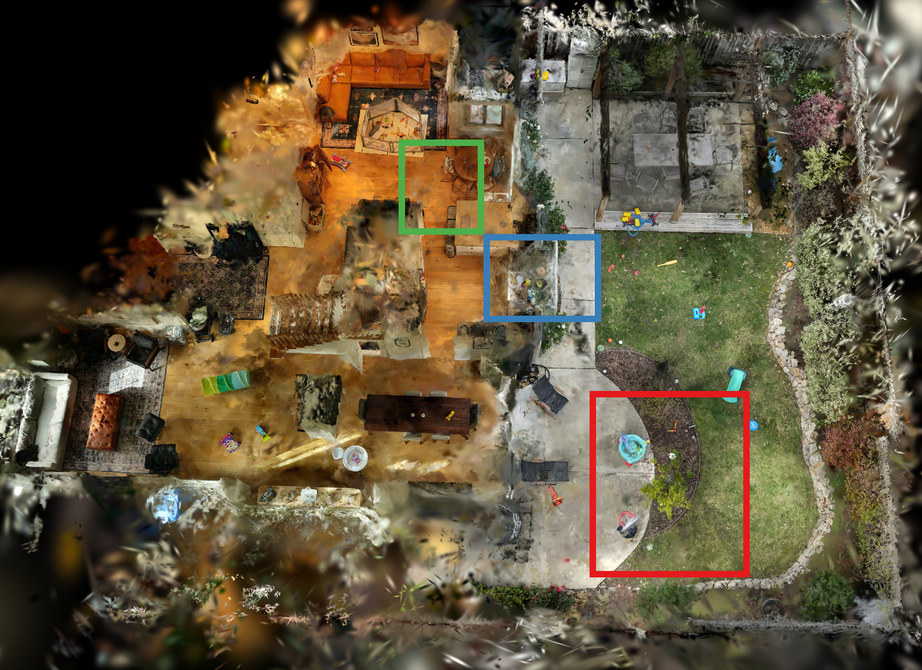}
    \end{minipage}\hspace{0.006\linewidth}%
    \begin{minipage}[c]{0.6325\linewidth}
        \centering
        \setlength{\tabcolsep}{1pt}
        \begin{tabular}{ccc}
            \scriptsize Ground truth & \scriptsize StopThePop & \scriptsize Ours \\
            \teaserbox{teasergreen}{\includegraphics[width=\dimexpr0.3245\linewidth-2\tbrule\relax,trim=27 88 628 333,clip]{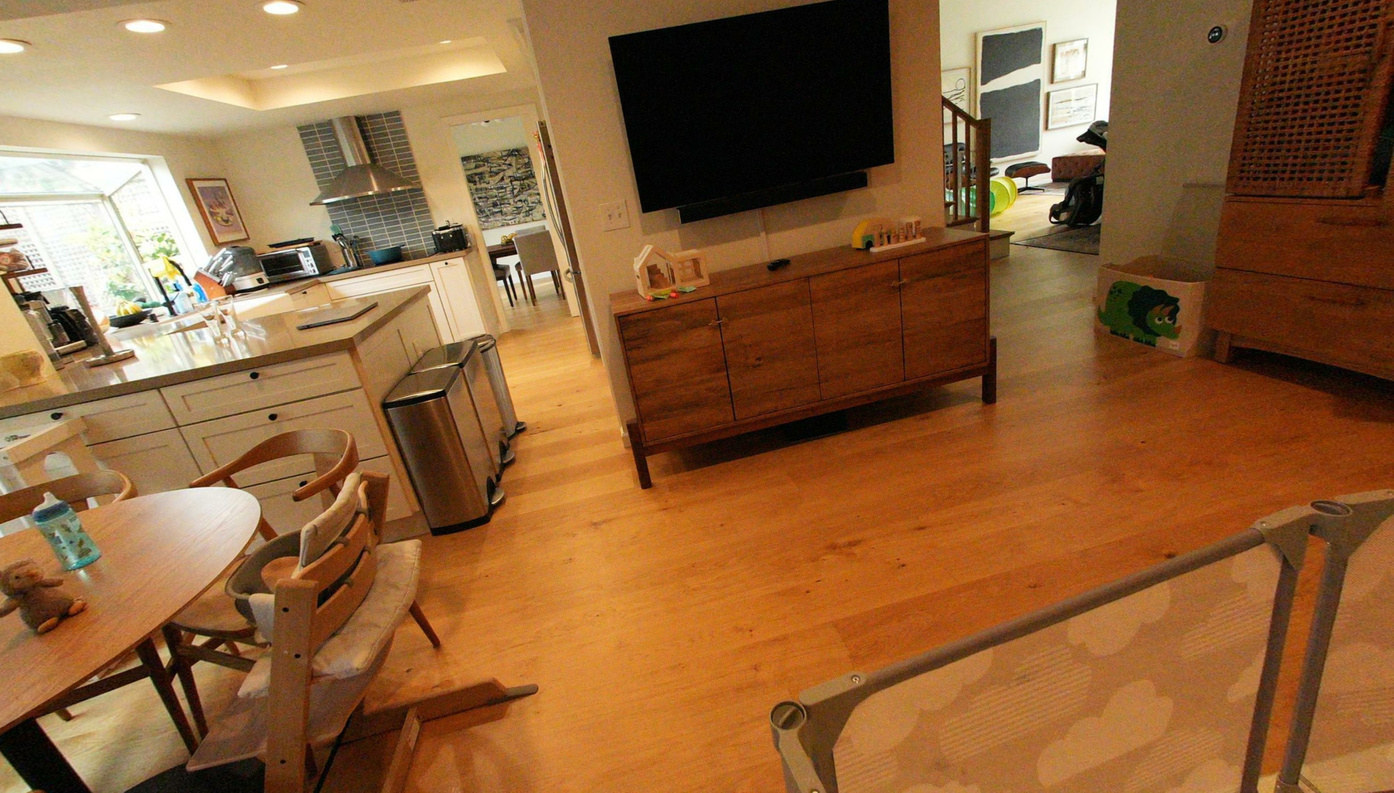}} &
            \teaserbox{teasergreen}{\includegraphics[width=\dimexpr0.3245\linewidth-2\tbrule\relax,trim=27 88 628 333,clip]{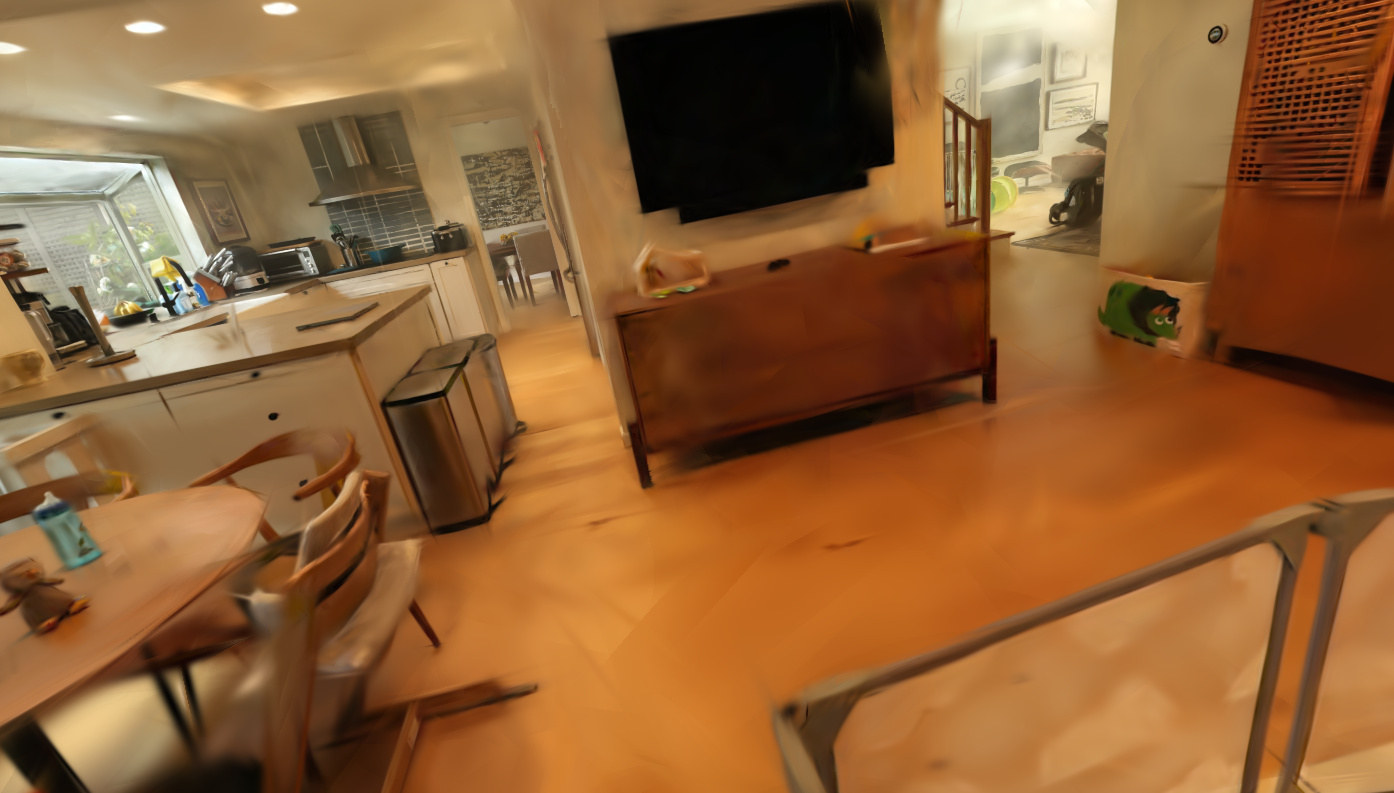}} &
            \teaserbox{teasergreen}{\includegraphics[width=\dimexpr0.3245\linewidth-2\tbrule\relax,trim=27 88 628 333,clip]{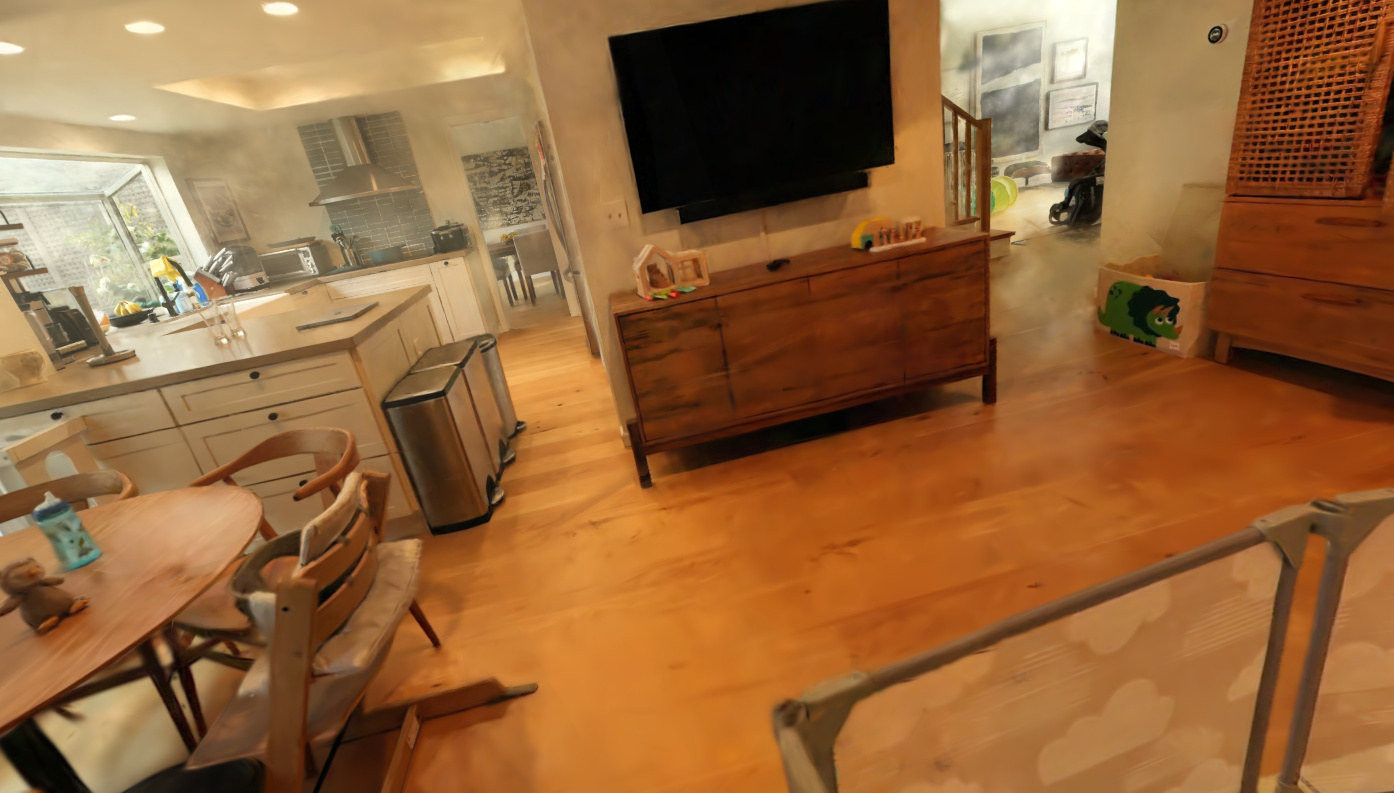}} \\
            \teaserbox{teaserblue}{\includegraphics[width=\dimexpr0.3245\linewidth-2\tbrule\relax,trim=529 96 112 317,clip]{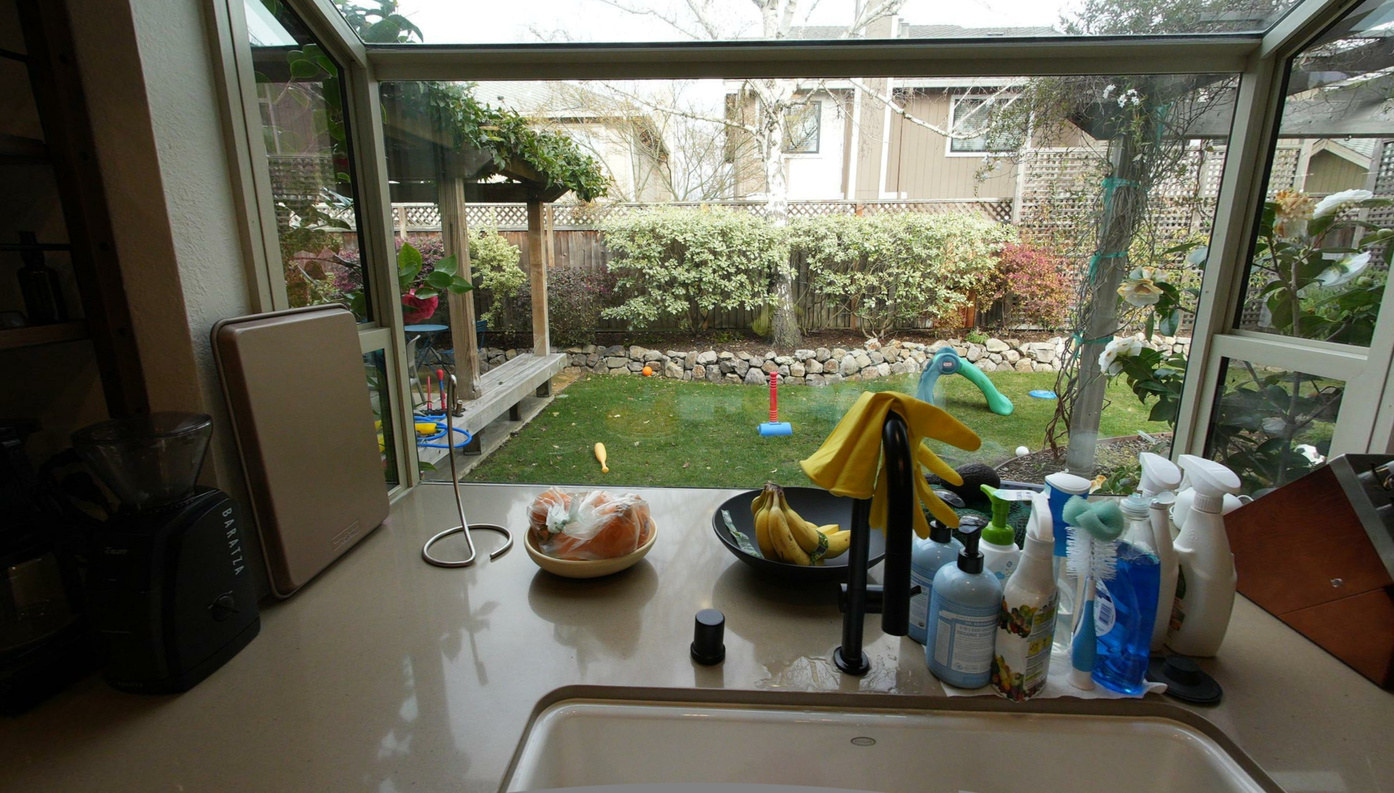}} &
            \teaserbox{teaserblue}{\includegraphics[width=\dimexpr0.3245\linewidth-2\tbrule\relax,trim=529 96 112 317,clip]{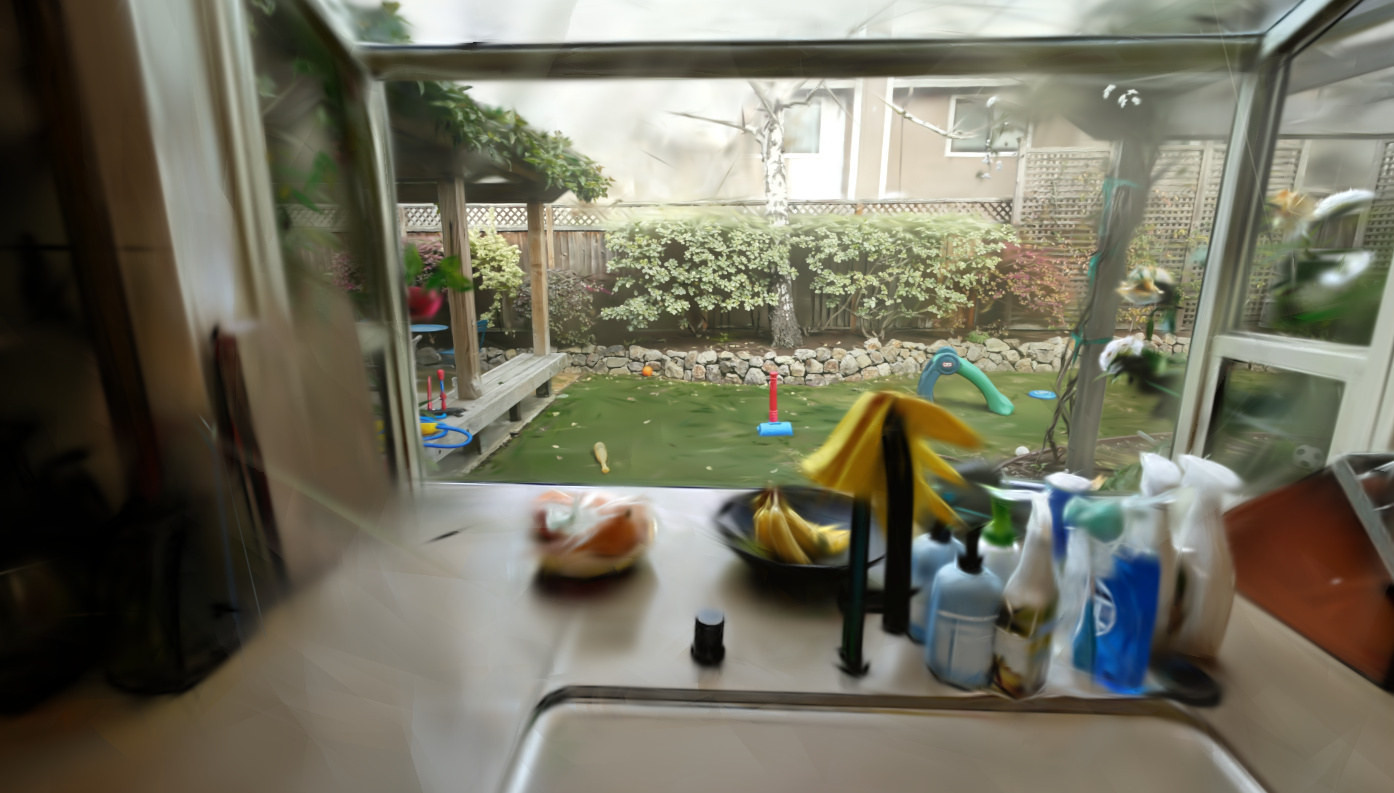}} &
            \teaserbox{teaserblue}{\includegraphics[width=\dimexpr0.3245\linewidth-2\tbrule\relax,trim=529 96 112 317,clip]{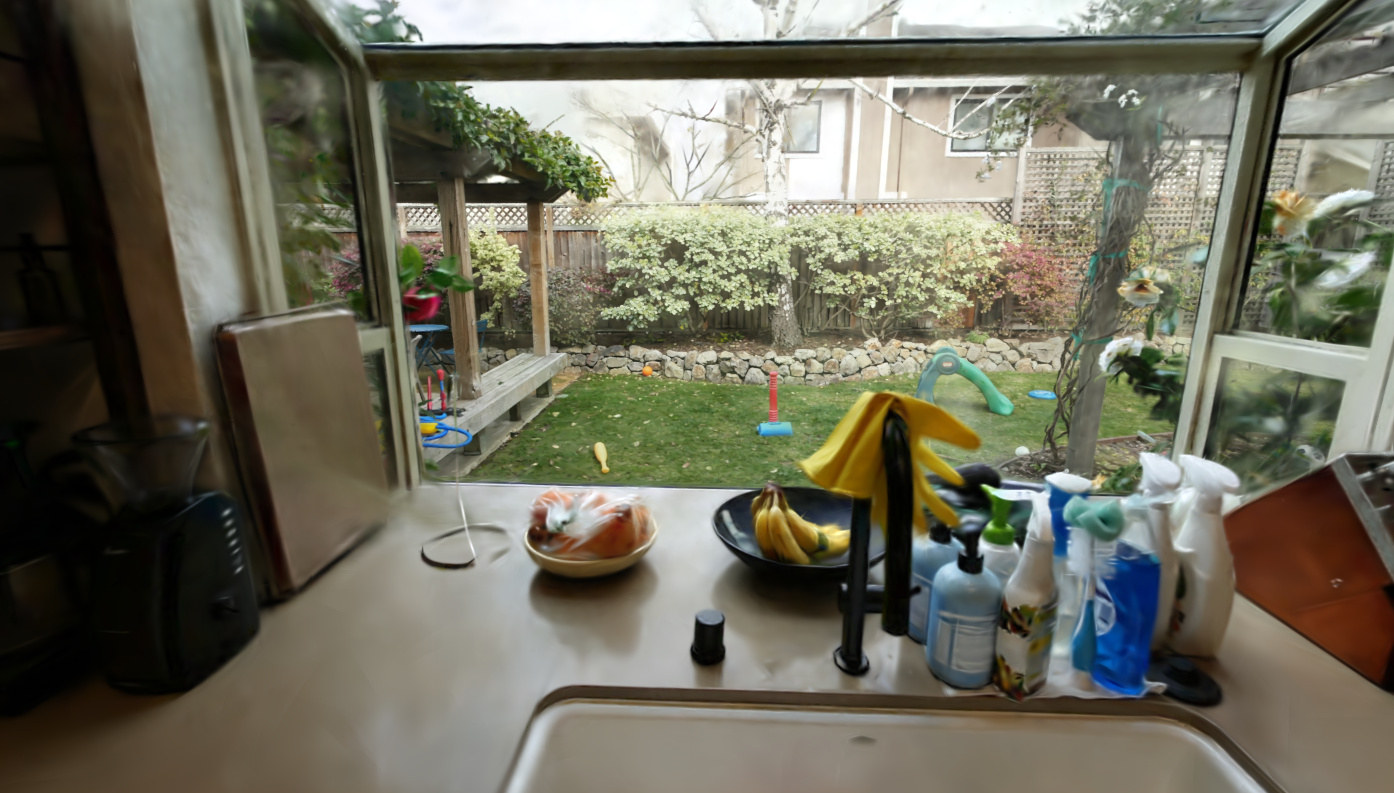}} \\
            \teaserbox{teaserred}{\includegraphics[width=\dimexpr0.3245\linewidth-2\tbrule\relax,trim=390 175 251 237,clip]{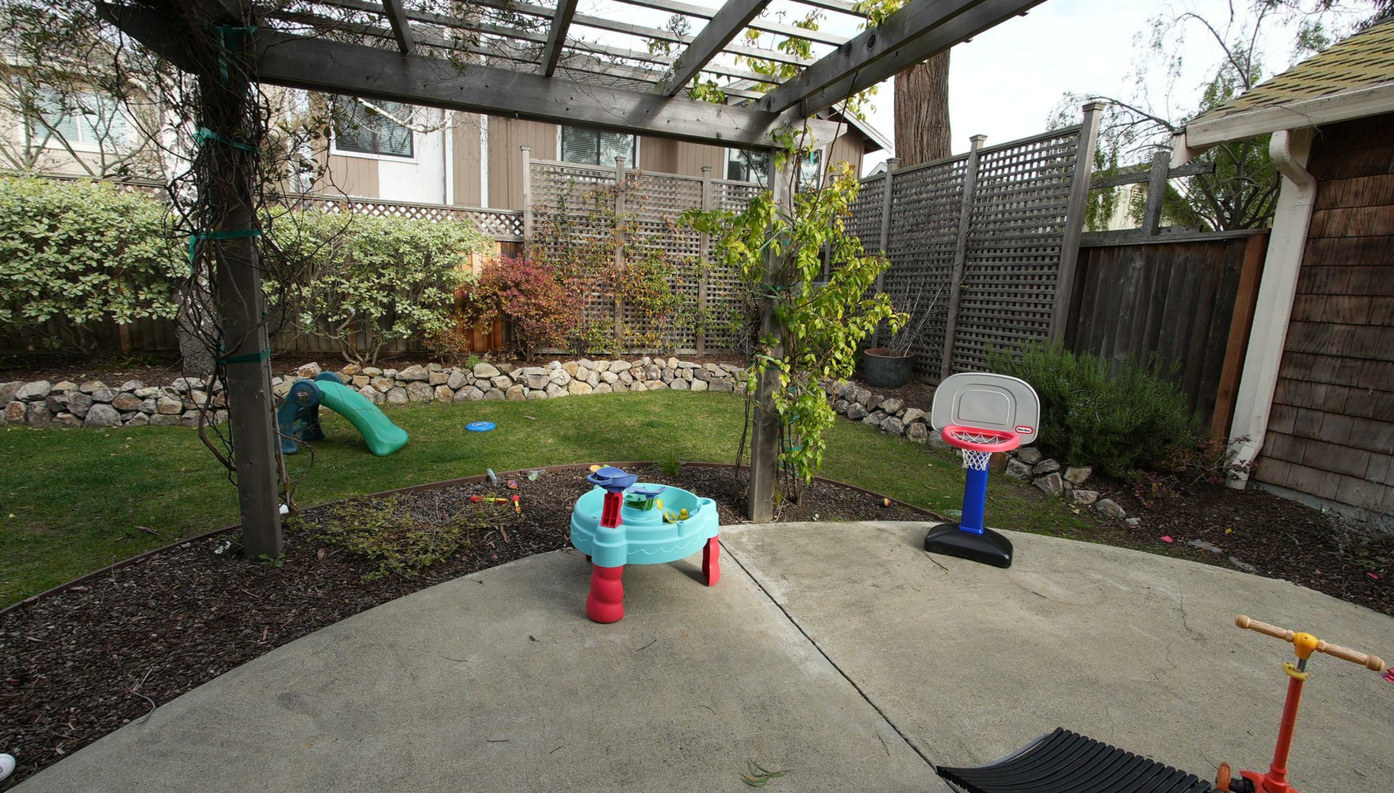}} &
            \teaserbox{teaserred}{\includegraphics[width=\dimexpr0.3245\linewidth-2\tbrule\relax,trim=390 175 251 237,clip]{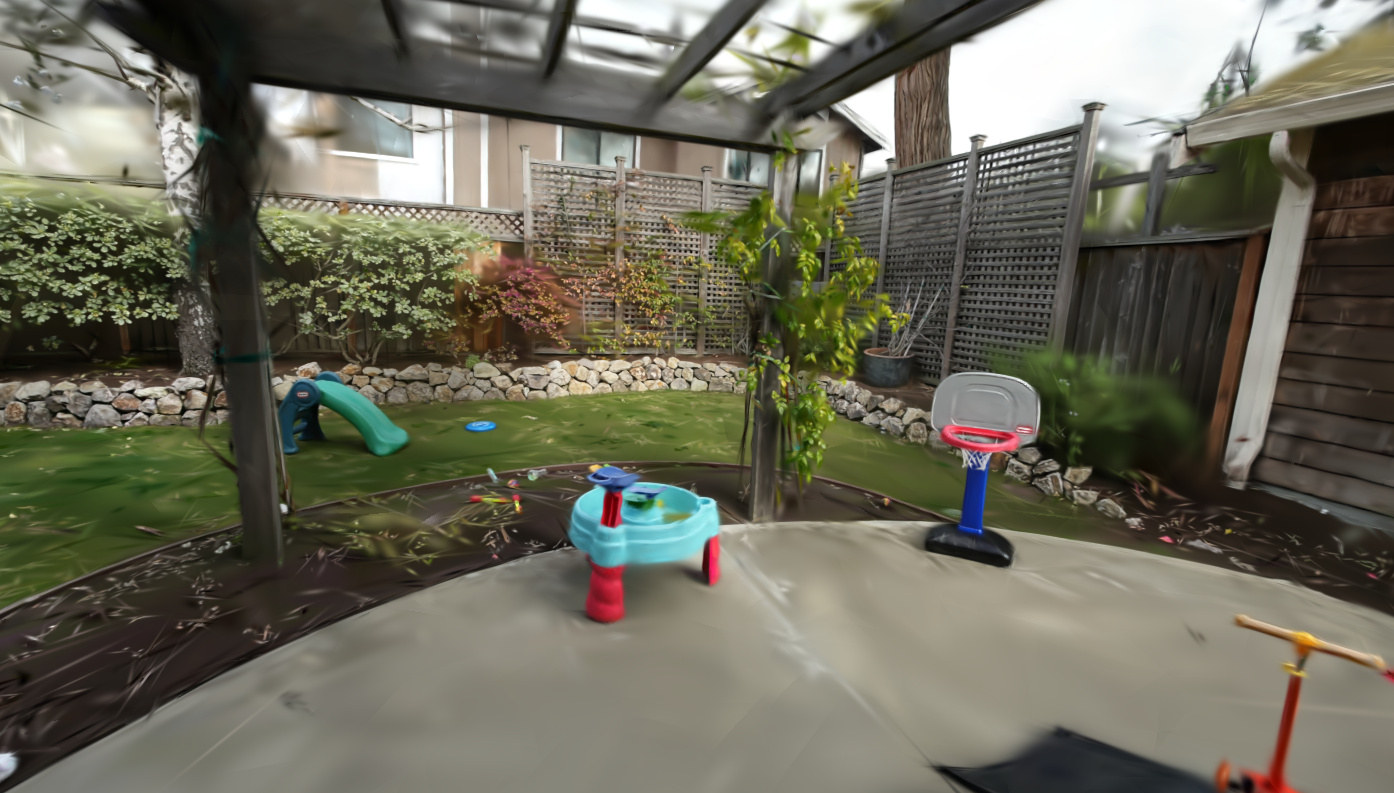}} &
            \teaserbox{teaserred}{\includegraphics[width=\dimexpr0.3245\linewidth-2\tbrule\relax,trim=390 175 251 237,clip]{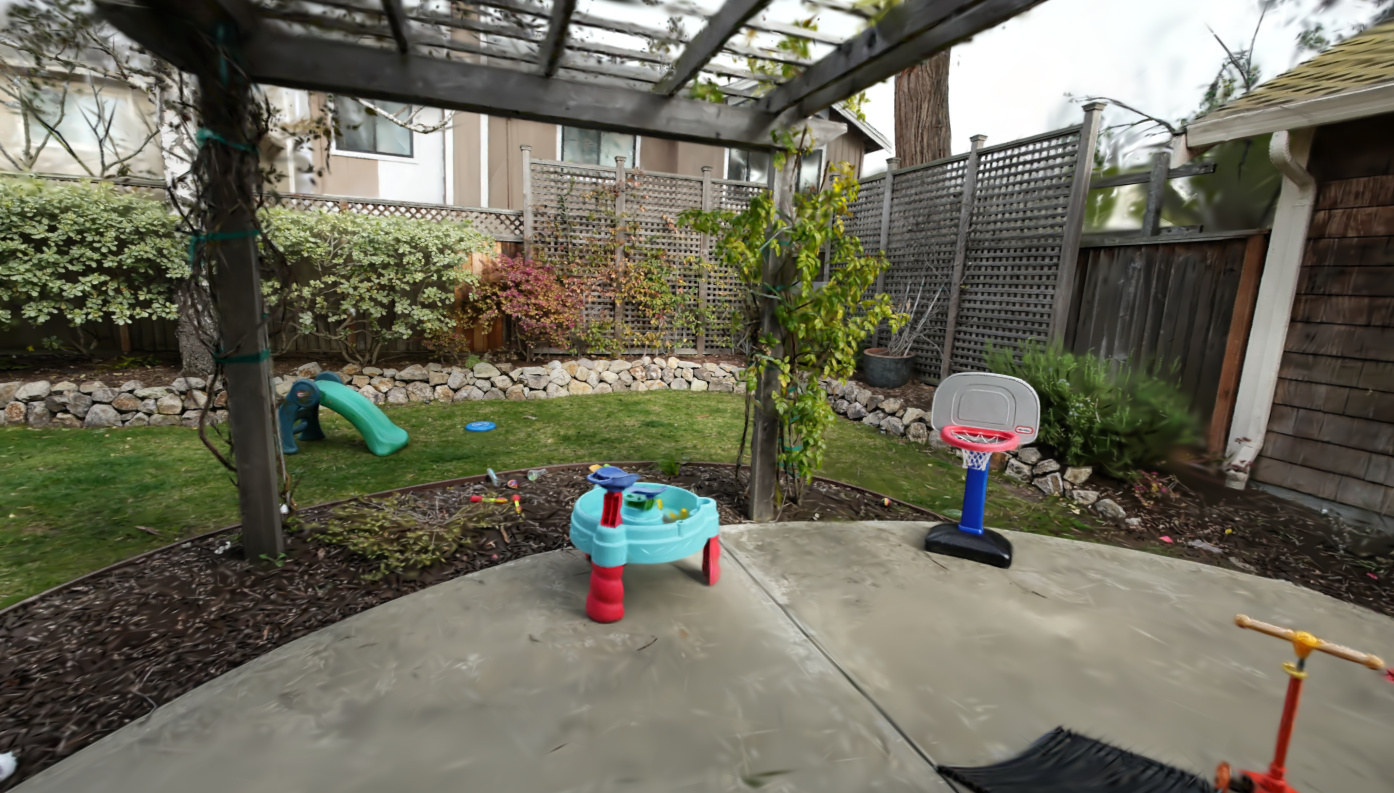}} \\
        \end{tabular}
    \end{minipage}
\end{center}
\vspace{-6pt}
\captionof{figure}{\textbf{Occlusion-aware densification.} Left: top-down overview of the large alameda scene (modified reconstruction; for orientation only). Most geometry is occluded from any given position. Right: crops of the marked locations. Occlusion-blind densification (such as StopThePop) systematically \textit{starves} occluded regions of Gaussians; our occlusion-aware approach resolves the missing detail.} \label{fig:teaser}

\section{Introduction} \label{sec:intro}
3D Gaussian Splatting (3DGS)~\cite{kerbl3Dgaussians} represents a milestone in 3D scene reconstruction and novel-view synthesis. 
It introduced a large multi-component system, with a tile-based rasterizer and adaptive density control (ADC) purpose-built for the new representation. 
Follow-up works improve these components largely in isolation, many focusing on ADC, which is essential for growing a dense reconstruction from the sparse initialization. 
However, we find that the accumulation of the densification statistics at the root of all ADC decisions has a fundamental flaw: 
the frustum-based visibility filter that gathers them tests only whether a Gaussian could appear in a view, not whether it actually contributes to the rendered image. 
A fully occluded Gaussian yields zero gradients, yet the filter still counts its view in the criterion's denominator, dragging the average down and stalling densification wherever geometry is occluded. 
The resulting under-densification grows with the degree of occlusion and degrades quality most where high detail and occlusion co-occur. 
Large, occlusion-heavy scenes such as multi-room apartments are affected most severely (\cref{fig:teaser}). 
We correct this with an occlusion-aware visibility filter that updates a Gaussian's statistics only from views in which it is actually rendered; densification then reaches occluded regions whenever their level of detail requires it. 

We further find that normalizing the accumulated gradient by a raw view count does not match the contribution-weighted nature of the tracked gradients. 
We derive a more natural normalizer that accumulates a Gaussian's average contribution to each rendered view. 
To validate our occlusion-aware densification, we build a framework around StopThePop~\cite{radl2024stopthepop} that consolidates established densification improvements from separate prior methods into a strong baseline. 
We evaluate both changes on this baseline and directly on StopThePop. 
On the occlusion-heavy Zip-NeRF~\cite{barron2023zipnerf} scenes, the resulting method outperforms StopThePop on all image-quality metrics, reducing LPIPS by up to $12\%$. 
Its Gaussian count also stays bounded under the longer schedules these large scenes require, where most competing criteria diverge. 
On smaller benchmarks, it performs comparably to the state of the art with up to $63\%$ fewer Gaussians. 
Our contributions are as follows:
\begin{itemize}
    \item To the best of our knowledge, we are the first to replace the visibility filter itself: the frustum-based test that underlies \emph{all} densification statistics. Our occlusion-aware filter updates a Gaussian's statistics only from views in which it is actually rendered, independent of the criterion built on top.
    \item We introduce a contribution-based densification criterion that is more closely related to the gradient signal used by ADC, leading to better densification decisions, demonstrated on small- and large-scale scenes.
    \item We combine both contributions with established densification improvements and other refinements into a strong framework that does not require per-scene tuning and has fewer hyperparameters (\eg eliminating the unstable scene extent). We will make the code available to the research community.
\end{itemize}

\section{Background and Related Work} \label{sec:related}

\subsection{Adaptive Density Control} \label{sec:preliminaries} \label{sec:preliminaries:adc}
3D Gaussian Splatting~\cite{kerbl3Dgaussians} represents a scene as a set of 3D Gaussians, each defined by a 3D mean position, a 3D covariance matrix (parameterized by a rotation and per-axis scale), an opacity $o$, and a view-dependent color $c$ derived from spherical harmonics~\cite{yu2022plenoxels}. 
For rendering, each Gaussian is projected onto the image as a 2D footprint with a screen-space mean and radius. 
Starting from a sparse point cloud, 3DGS grows and prunes its set of Gaussians via adaptive density control (ADC), executed every 100 iterations after a 500-iteration warm-up and disabled after the first half of training. 
A scene extent $E$, computed from the training cameras, provides the reference scale for several ADC decisions.

\paragraph{Criterion.}
During rendering, a frustum-based visibility filter accumulates per-Gaussian statistics: the 2D screen-space positional gradient $g_{i,v}$ of Gaussian $i$ in view $v$, the screen-space radius, and a view count $M$. 
At each ADC event, a Gaussian is marked for densification if its accumulated gradient magnitude, averaged over the view count, exceeds a predefined threshold ($0.0002$ by default).

\paragraph{Densification.}
To resolve the problem of an insufficient number of Gaussians capturing a region's textural or structural detail (from here on described as under-densification), a marked Gaussian is densified in one of two modes: 
if it is large relative to the scene extent, it is \emph{split} into two downscaled children sampled from the parent's distribution, otherwise it is \emph{cloned}, \ie duplicated, in place.

\paragraph{Pruning.}
Redundant or noisy Gaussians are removed using pruning mechanisms. 
An opacity prune removes every Gaussian whose opacity $o_i$ falls below a predefined threshold ($0.005$). 
A size-based prune removes Gaussians whose screen-space radius or world-space scale exceeds a fixed threshold. 
To clear floater artifacts, 3DGS additionally resets all opacities to near zero ($0.01$) every $3000$ iterations; important Gaussians recover quickly, floaters stay dim and fall to the opacity prune. 
StopThePop~\cite{radl2024stopthepop} offers a gradual opacity \emph{decay} as an alternative to the drastic reset. 
We evaluate both as the (Reset) and (Decay) variants.

\subsection{Adaptive Density Control Improvements}
All of the following ADC improvements inherit the frustum-based visibility filter described above, which counts every in-frustum Gaussian, occluded or not. 
We make this filter occlusion-aware (\cref{sec:method:occlusion}).

\paragraph{Densification Criterion.}
AbsGS~\cite{ye2024absgs} and GOF~\cite{Yu2024GOF} counter gradient cancellation in the accumulation, which otherwise causes under-densification of high-detail regions. 
PixelGS~\cite{zhang2024pixelgs} instead questions the normalization by the observation count and proposes a pixel-count-weighted mean. 
Grubert \etal~\cite{Grubert2025improving} adopt the PixelGS strategy with an ascending threshold that shifts densification earlier. 
Taming3DGS~\cite{mallick2024taming3dgs} replaces the threshold with a per-Gaussian multi-term score under a fixed densification budget. 
However, budget-controlled methods require a target count as input, a value that is not known and that a good densification criterion should find. 
We propose to normalize a Gaussian's accumulated gradient by the accumulated average contributions of its rendered views (\cref{sec:method:criterion}), rather than the observation count of 3DGS or the visible-pixel count of PixelGS. 
We also adopt GOF's absolute-gradient signal against cancellation.

\paragraph{Densification Step (Split \& Clone).}
GOF observes that clones remain clustered and instead samples the child's position from the parent's distribution, while \bulo\etal~\cite{bulo2024revising} show that the clone operation introduces an opacity bias. 
EFA-GS~\cite{wang2025lowfrequencyfirsteliminatingfloating} and Deng \etal~\cite{deng2025improvingdensification3dgaussian} drop the clone entirely, and Deng \etal additionally propose a geometrically more meaningful split procedure. 
Grubert \etal address inconsistencies in the scene-extent computation that governs some ADC decisions by also accounting for the input point cloud. 
Other works replace the densification step more radically~\cite{kheradmand2024mcmc,cheng2024gaussianpro,fang2024minisplatting,wang2025steepgs}. 
Our experiments also support dropping the clone: 
we notice large floater clusters due to excessive cloning on scenes where the scene extent is not well matched to the size of detail (\cref{tab:exp:ablation-zip}). 
We therefore follow the split-only approach of EFA-GS and Deng \etal, including the latter's split procedure. 
This removes one dependence on the scene extent.

\paragraph{Pruning.}
Grubert \etal restrict the opacity prune to Gaussians that also contribute little; EAGLES~\cite{girish2024eagles} and OC2DGS~\cite{RoggeOC2DGS2025} prune low- or non-contributing Gaussians at intervals long enough to observe every training view; ReLaGS~\cite{xiearafa2026relags} and RadSplat~\cite{niemeyer2025radsplat} prune by maximum contribution over a full pass, at isolated points of the optimization. 
We prune by maximum contribution, as in ReLaGS and RadSplat, but throughout the optimization, as in EAGLES and OC2DGS. 

\subsection{Large-Scale Scenes}
Large-scale scenes harbor a range of problems for Gaussian Splatting: 
Kerbl \etal~\cite{hierarchicalgaussians24} partition them into chunks optimized individually after a coarse full-scene pass. 
They note problems with the regular densification regime and adopt the maximum 2D screen-space gradient as the criterion. 
OccluGaussian~\cite{liu2025occlugaussian} makes the partitioning occlusion-aware, clustering by what each camera sees so that no chunk contains heavy occlusions, and optimizes chunks individually before merging them. 
RadSplat observes that lighting variations in the Zip-NeRF dataset~\cite{barron2023zipnerf} cause 3DGS to struggle and produce overly smooth results even with an exposure-handling module. 
SMERF~\cite{duckworth2023smerf} performs extensive densification-related hyperparameter tuning on Zip-NeRF yet reports low Gaussian counts, indicating under-densification. 
Lowering the gradient threshold raises the count only at the cost of more floaters. 
Neither work identifies the root cause. 
We argue it is occlusion-blind densification (a cause that exposure correction and hyperparameter tuning leave unaddressed) and provide supporting evidence in \cref{sec:exp:occlusion}. 
Unlike OccluGaussian, we do not avoid this problem by optimizing occluded regions as individual small scenes, but instead make the densification occlusion-aware, enabling a single full-scene optimization.

\section{Method} \label{sec:method}
We build on StopThePop~\cite{radl2024stopthepop}, a state-of-the-art Gaussian Splatting renderer whose exact per-pixel depth ordering removes popping artifacts and keeps occlusion boundaries intact. 
Hybrid transparency~\cite{hahlbohm2025htgs}, in contrast, sorts only the front-most Gaussians and composites all remaining ones without enforcing occlusion ordering. 
Leaving even fully occluded Gaussians with a nonzero contribution stands in opposition to our motivation of occlusion-aware densification. 
However, our method is not dependent on the per-pixel sorting of StopThePop: it relies only on the per-pixel blending weights that virtually every Gaussian Splatting rasterizer already computes. 
Our contributions can therefore transfer to other bases (evaluated on 2DGS~\cite{Huang2DGS2024} and 3DGS~\cite{kerbl3Dgaussians} in Sec.~A.2 of the supplementary material). 

\paragraph{Differentiable rasterization.}
Projected onto the image plane, a Gaussian $i$ forms a 2D footprint $G_{i,p}\in(0,1]$ over pixels $p$~\cite{zwicker2001ewa}, yielding the per-pixel opacity $\alpha_{i,p}=o_i G_{i,p}$. 
3DGS renders with a tile-based differentiable rasterizer: a preprocessing step culls Gaussians against the camera view frustum, maps the remaining ones to the image tiles they overlap, and sorts them by view-space depth. 
Each pixel $p$ is rendered by front-to-back alpha compositing of the depth-sorted Gaussians that overlap it, yielding its color $C_p$:
\begin{equation} \label{eq:compositing}
    C_p = \sum_i c_i\, \alpha_{i,p}\, T_{i,p} + c_{bg}\, T_{final,p}, \qquad T_{i,p} = \prod_{j<i}(1-\alpha_{j,p}),
\end{equation}
where the transmittance $T_{i,p}$ is the fraction of light that reaches Gaussian $i$ at pixel $p$ past the Gaussians in front of it. 
Rendering terminates early once $T_{i,p}$ saturates (falls below $10^{-4}$), so Gaussians behind that point are never rendered. 
The remaining transmittance $T_{final,p}$ is filled with a constant background color $c_{bg}$. 
The product $w_{i,p}=\alpha_{i,p}\,T_{i,p}$ is the blending weight with which Gaussian $i$ contributes to pixel $p$: its per-pixel opacity $\alpha_{i,p}$ weighted by its visibility $T_{i,p}$.

The complete integrated framework, which assembles established densification improvements around them, is described in \cref{sec:exp:implementation}.

\subsection{Occlusion-Awareness} \label{sec:method:occlusion}
As discussed in \cref{sec:preliminaries:adc}, densification-related statistics are gathered via a visibility filter based on a view-frustum test. 
This reflects potential, not actual, observations of a Gaussian. 
From here on, we refer to Gaussians that are not rendered at all as \emph{occluded}: 
transmittance saturates before reaching them at every pixel they cover. 
With a frustum-based filter, occluded Gaussians thus enter the accumulated densification statistics despite not participating in the rendered image. 
Although their gradients are zero, every occluded view still increases the criterion's observation count and updates the maximum-radius tracking. 
Increasing the observation count results in a downward pressure on the densification criterion of any Gaussian that lies within the frustum while being obstructed. 
The consequence is under-densification in occlusion-heavy scenes such as the multi-room apartments of Zip-NeRF~\cite{barron2023zipnerf}, as also observed by prior work~\cite{duckworth2023smerf,niemeyer2025radsplat}.

We propose an occlusion-aware visibility filter that restricts the influence of rendering statistics to those Gaussians that actually participate in rendering the image. 
For the filter itself, a binary signal suffices: 
a Gaussian counts as observed in a view only if it contributes to at least one pixel. 
We obtain this signal from the total contribution the Gaussian accumulates during rendering: 
the sum of its blending weights $w_{i,p}$ over all pixels of the view. 
This quantity is reused by our criterion in \cref{sec:method:criterion}. 
If the total contribution is zero, the Gaussian did not participate in the rendering and is therefore occluded (\cref{fig:method:occlusion-selection}). 
Replacing the original frustum-based visibility filter with our occlusion-aware filter thus ensures occlusion-aware densification statistics.

\begin{figure}[t]
    \centering
    \begin{minipage}{0.32\linewidth}\centering
        \includegraphics[width=\linewidth]{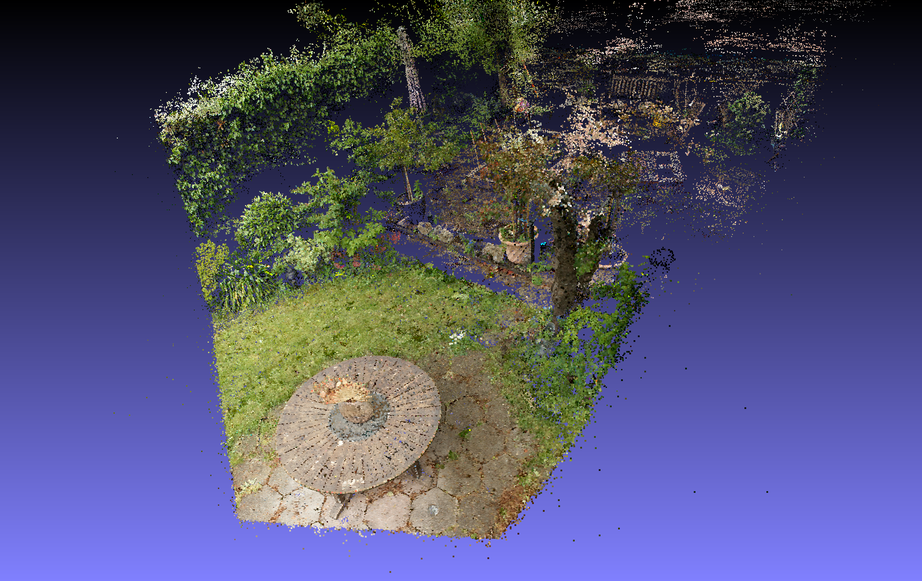}\\
        \small (a) Frustum-visible
    \end{minipage}\hfill
    \begin{minipage}{0.32\linewidth}\centering
        \includegraphics[width=\linewidth]{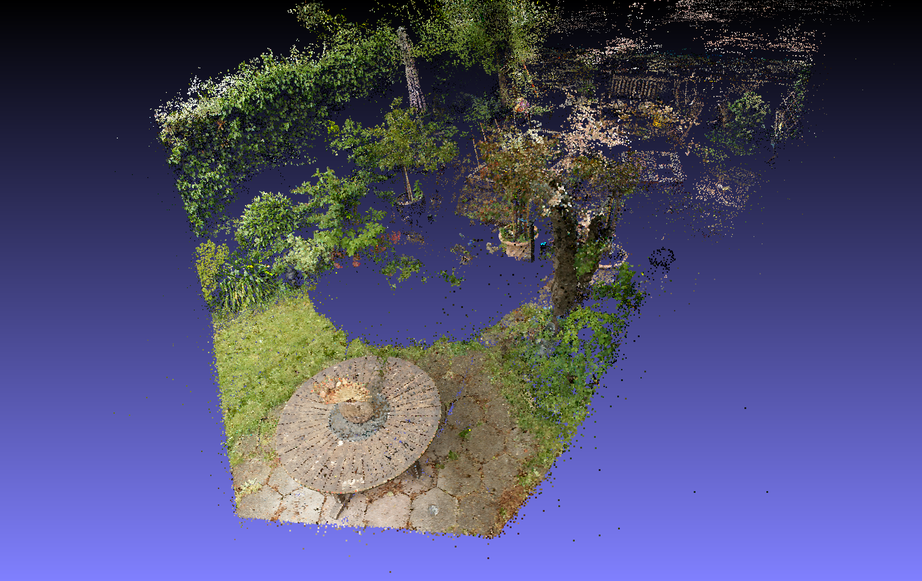}\\
        \small (b) Contributing
    \end{minipage}\hfill
    \begin{minipage}{0.32\linewidth}\centering
        \includegraphics[width=\linewidth]{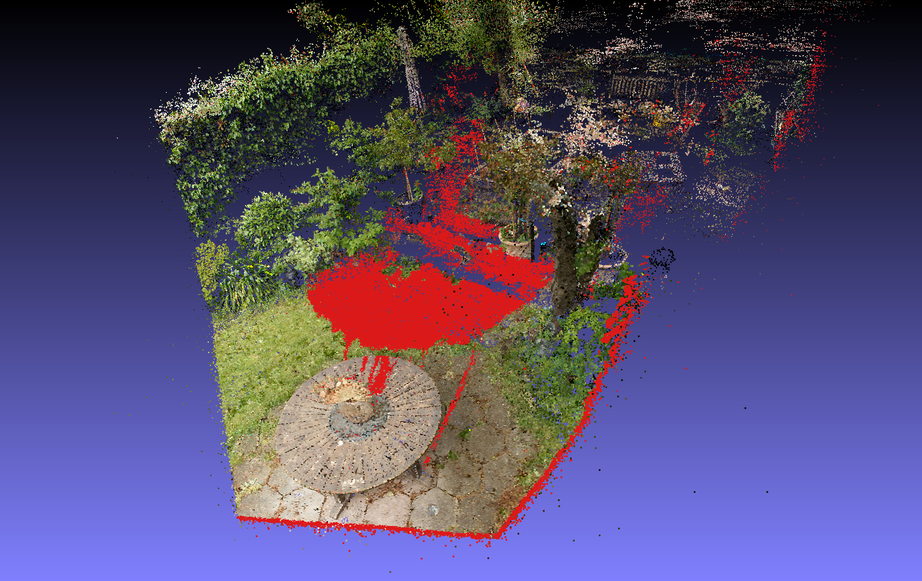}\\
        \small (c) Occluded (red)
    \end{minipage}
    \caption{Occlusion-aware versus frustum-based visibility filter. Showing all Gaussians in front of a camera from the Mip-NeRF360 garden scene, which (a) pass the frustum test or (b) actually contribute to the rendered image. (c) shows their difference in red.}
    \label{fig:method:occlusion-selection}
\end{figure}

The occlusion signal could equally come from a Boolean rendered flag as computed in OC2DGS~\cite{RoggeOC2DGS2025} or a rendered-pixel count as in PixelGS~\cite{zhang2024pixelgs}. 
However, it is essential that the filter gating the densification statistics is derived from such a rendering signal rather than remaining based on the view frustum.

\subsection{Average Contribution Criterion} \label{sec:method:criterion}
3DGS decides densification from a Gaussian's positional gradients in screen space, averaged in magnitude across its observed views. 
For a Gaussian $i$ observed in $M$ views, let $g_{i,v}=\partial L / \partial \mu_i$ denote the gradient of the training loss $L$ with respect to the Gaussian's projected screen-space mean $\mu_i$ in view $v$. 
A Gaussian is densified (\ie split or cloned) when
\begin{equation} \label{eq:criterion:3dgs}
    \gamma_{i}^{\text{3DGS}}=\frac{1}{M} \sum_{v}\|g_{i,v}\|
\end{equation}
exceeds a threshold. 
The denominator $M$ weights every view equally: a view in which the Gaussian is barely visible counts the same as a fully visible one. 
As discussed in \cref{sec:method:occlusion}, fully occluded views inflate $M$ while contributing nothing to the gradient sum, biasing the average downward.

To identify a more faithful measure of observation, we examine how the gradient is formed. 
The per-view gradient $g_{i,v}=(g_{i,v,x}, g_{i,v,y})$ is accumulated over pixels to which Gaussian $i$ contributes ($w_{i,p}> 0$), and the contribution of each pixel $p$ is gated by the Gaussian's blending weight $w_{i,p}$. 
Per pixel, the gradient factorizes into the scalar $w_{i,p}$ and a weight-independent vector $s_{i,p}$:
\begin{equation} \label{eq:gradient}
    g_{i,v}=\sum_p w_{i,p} s_{i,p}, \hspace{20px} w_{i,p}=\alpha_{i,p}\,T_{i,p},
\end{equation}
where the per-pixel vector $s_{i,p}$ combines the photometric residual, the geometric steering direction, and the local contrast of the Gaussian's color against what lies behind it, all independent of the Gaussian's visibility (detailed derivation in Sec.~A.1). 
The blending weight $w_{i,p}$ is thus simultaneously the quantity with which the rasterizer composites the Gaussian's color and the factor that scales every term of its positional gradient. 
A pixel where the Gaussian is mostly occluded ($T_{i,p}\rightarrow0$) or barely present ($\alpha_{i,p} \rightarrow0$) contributes almost nothing to $g_{i,v}$, irrespective of that pixel's error. 
The gradient is thus already visibility-weighted at the pixel level, which motivates measuring a Gaussian's observation by its blending weights $w_{i,p}$ rather than by a binary view count.

We summarize a Gaussian's participation in view $v$ by its mean blending weight $\bar{w}_{i,v}$ over the $N_{i,v}$ pixels in which it contributes as
\begin{equation}
    \bar{w}_{i,v} = \frac{1}{N_{i,v}} \sum_p w_{i,p} \in [0,1)
\end{equation}
and replace the view count in \cref{eq:criterion:3dgs} with the accumulated $\bar{w}_{i,v}$:
\begin{equation} \label{eq:criterion:ours}
    \gamma_{i}^{\text{Ours}}=\frac{1}{1+\sum_v \bar{w}_{i,v}}\sum_v \|g_{i,v}\|.
\end{equation}
We normalize by the mean per-view contribution $\bar{w}_{i,v}$ rather than the total $W_{i,v}=\sum_p w_{i,p}=N_{i,v}\bar{w}_{i,v}$. 
Any denominator that scales with the pixel count $N_{i,v}$ ties densification to screen-space footprint area and suppresses exactly the large Gaussians that are most likely to require it. 
In contrast, the mean measures how visible the Gaussian is where it is rendered, independently of coverage. 
Moreover, because $\bar{w}_{i,v}$ contains the transmittance $T_{i,p}$, the criterion is occlusion-aware without an explicit visibility test: 
a view in which the Gaussian is only faintly visible cannot dilute the average, unlike a view-count normalizer. 
Whether such a view pushes toward densification is then decided by its gradient alone: 
a faint view with a large residual contributes its full signal with minimal normalization cost, whereas a faint view with little error simply fades out.

This incidental occlusion-awareness is shared by the pixel-count normalizer of PixelGS~\cite{zhang2024pixelgs}: 
a fully occluded view renders no pixels and drops out of its denominator as well. 
Beyond that, however, a visible-pixel count entangles view visibility with screen-space footprint, over-densifying near-camera Gaussians. 
PixelGS counteracts this with a separate distance scaling term based on the scene extent $E$. 
Our per-view mean $\bar{w}_{i,v}$ avoids this entanglement entirely. 
Additionally, through the transmittance $T_{i,p}$, it attenuates partial occlusion, grazing observations, and low-opacity views that leave a pixel count unchanged.

The constant in the denominator of \cref{eq:criterion:ours} is a fixed floor: without it, rarely observed Gaussians ($\sum_v \bar{w}_{i,v} \rightarrow 0$) would receive unbounded scores where evidence is weakest. 
The floor ensures that scores increase smoothly and monotonically with accumulated evidence, placing rarely and frequently observed Gaussians on a common scale. 
A sweep in Sec.~F.2 of the supplementary material confirms this: 
without the floor training diverges, while other floor values mainly re-scale the threshold. 
The updated criterion operates on a new scale and therefore requires a one-time recalibration of the densification threshold. 
A threshold sweep on a subset of the Mip-NeRF360 dataset shows $0.002$ to be the operating point. 
We keep this value fixed across all experiments.

Finally, the normalization in \cref{eq:criterion:ours} depends only on the per-pixel blending weights $w_{i,p}$, not on how the 2D gradients are reduced to a scalar value. 
Therefore, it applies unchanged to the absolute-gradient formulations of AbsGS~\cite{ye2024absgs} and GOF~\cite{Yu2024GOF}. 
We apply the same criterion computation to all gradient terms.

\section{Experiments} \label{sec:experiments}
\subsection{Datasets and Metrics}
We evaluate on the common Gaussian Splatting scenes: Mip-NeRF360~\cite{barron2022mipnerf360} (9 scenes), Tanks \& Temples~\cite{Knapitsch2017tanks} (2 scenes), and Deep Blending~\cite{DeepBlending2018} (2 scenes). 
Additionally, we evaluate on Zip-NeRF~\cite{barron2023zipnerf}, whose 4 large apartment scenes (alameda, berlin, london, nyc) exhibit heavy occlusions between rooms, with an occlusion rate of $0.63$ against $0.15$--$0.48$ on the other datasets (definition and per-scene values in Sec.~D). 
Zip-NeRF is the established stress test for this regime: prior work reports 3DGS under-densification and over-smoothing on these scenes despite extensive tuning~\cite{duckworth2023smerf,niemeyer2025radsplat}. 
For Zip-NeRF, we use images at half their original resolution, mirroring common 3DGS target resolutions, and hold out every 8th image as the validation set, following the standard protocol~\cite{kerbl3Dgaussians}. 
We report peak signal-to-noise ratio (PSNR), structural similarity (SSIM)~\cite{wang2004ssim}, learned perceptual image patch similarity (LPIPS; VGG backbone)~\cite{zhang2018lpips}, the final number of Gaussians, and the training time. 
Large-scale scenes are prone to over-smoothing due to under-densification~\cite{niemeyer2025radsplat,duckworth2023smerf}. 
We therefore also report the gradient magnitude similarity deviation (GMSD)~\cite{xue2014gmsd}, a metric specifically sensitive to blur. 
We read these metrics jointly: 
perceptual ones such as LPIPS credit reconstructed detail that \eg PSNR penalizes when not pixel-aligned. 
Likewise, a lower Gaussian count is only better at comparable quality: 
a starved model is not more efficient.

\subsection{Implementation} \label{sec:exp:implementation}
We embed our contributions in the integrated framework: 
the publicly available StopThePop~\cite{radl2024stopthepop}, extended with the following improvements proposed by related work, which we do not claim as contributions. 
Unless stated otherwise, all runs use the opacity \emph{decay} variant from StopThePop, and reported numbers are averaged over three training runs. 
All experiments run on an Nvidia A40 GPU.

\paragraph{Absolute gradient.} 
We include the absolute-gradient condition of GOF~\cite{Yu2024GOF}, which improves high-frequency detail at an increased Gaussian count (\cref{tab:exp:ablation-zip}).
\paragraph{Gaussian densification.}
We follow a split-only strategy, as also adopted by EFA-GS~\cite{wang2025lowfrequencyfirsteliminatingfloating} and Deng \etal~\cite{deng2025improvingdensification3dgaussian}. 
In our setting this removes a dependence on the scene extent $E$ and avoids the excessive cloning that leads to pronounced floaters or total failure (\cref{tab:exp:ablation-zip}). 
We also adopt Deng \etal's geometrically meaningful long-axis split, which slightly reduces the count at competitive quality.
\paragraph{Pruning.}
Following the recommendation of Grubert \etal~\cite{Grubert2025improving}, we disable size-based pruning, which we find has no significant impact on quality or Gaussian count (\cref{tab:exp:ablation-zip}). 
Disabling it also removes another dependence on the scene extent $E$. 
We additionally prune Gaussians whose maximum blending contribution $w_{max}$ stays below a fixed threshold ($0.005$), as discussed by ReLaGS~\cite{xiearafa2026relags} and RadSplat~\cite{niemeyer2025radsplat}. 
We track $w_{max}$ on the fly and prune throughout training~\cite{RoggeOC2DGS2025,girish2024eagles} on an automatic, hyperparameter-free schedule synchronized with the training queue. 
A single cheap pass over all training views at the end guarantees that the final model contains no fully occluded Gaussians (scheduling details in Sec.~E.1, occluded-Gaussian counts per method in Sec.~E.2). 
\paragraph{Densification statistics.}
Original 3DGS accumulates densification statistics already during warm-up, where the largely transparent scene does not yet reflect its geometry. 
We exclude the warm-up period, which affects only the first densification event and is confirmed safe (\cref{tab:exp:ablation-zip}).
\paragraph{Competitor comparison.}
We replicate PixelGS~\cite{zhang2024pixelgs} and Grubert \etal~\cite{Grubert2025improving} on the same StopThePop renderer in two settings: their full methods on vanilla StopThePop for the state-of-the-art comparison (\cref{tab:exp:sota}), and their criterion alone inside the integrated framework, for the criterion comparison (\cref{tab:method:criterion}). 

\subsection{Impact of Occlusion-Awareness} \label{sec:exp:occlusion}
Integrating the adopted densification components into StopThePop already yields a strong but still occlusion-blind baseline. 
It surpasses StopThePop on all image-quality metrics on both datasets (Zip-NeRF: LPIPS by $6\%$, PSNR by $0.57$\,dB; StopThePop comparison in Sec.~B). 
On top of this baseline, we isolate the occlusion-aware visibility filter (\cref{sec:method:occlusion}), leaving every other component unchanged. 
The top two rows of \cref{tab:method:occlusion-zip} compare the occlusion-blind baseline against its occlusion-aware counterpart. 
Making the filter occlusion-aware improves all image-quality metrics while enabling the densification that occlusion-blindness suppresses (LPIPS improves by a further $9\%$; the count rises from $0.56$M to $2.57$M). 
Merely lowering the occlusion-blind threshold is no equivalent control: 
it densifies indiscriminately rather than where occlusion suppressed it, and SMERF~\cite{duckworth2023smerf} reports that this produces floaters on these scenes. 
Occlusion-blind densification is thus a primary cause of under-densification on large, occlusion-heavy scenes. 
On Mip-NeRF360, where occlusions are far lighter, the same pairing confirms the corrected filter does no harm: 
LPIPS improves by $4\%$ at a moderately higher count (Sec.~F.3). 
\begin{table}[t]
    \centering
    \caption{Occlusion-aware filter and densification criteria on the occlusion-heavy Zip-NeRF scenes (Mip-NeRF360 counterpart in Tab.~S12). The occlusion-blind baseline, then varying only the criterion: the occlusion-aware observation count, the PixelGS pixel-count-weighted mean without and with distance scaling (+ DS), and our average contribution. Counts are in millions of Gaussians; time in minutes.} 
    \label{tab:method:occlusion-zip}\label{tab:method:criterion}
    \setlength{\tabcolsep}{3pt}
    \begin{tabular}{l*{6}{c}}
        \toprule
        Method & PSNR$\uparrow$ & SSIM$\uparrow$ & LPIPS$\downarrow$ & GMSD$\downarrow$ & Count$\downarrow$ & Time$\downarrow$ \\
        \midrule
        Occlusion-blind count & \hl{40}{25.52} & \hl{12}{0.818} & \hl{12}{0.352} & \hl{12}{0.117} & \hl{60}{0.557} & \hl{60}{22.3} \\
        \addlinespace
        Occlusion-aware count & \hl{56}{25.75} & \hl{60}{0.834} & \hl{52}{0.319} & \hl{60}{0.103} & \hl{44}{2.567} & \hl{45}{36.6} \\
        PixelGS & \hl{12}{25.11} & \hl{45}{0.829} & \hl{60}{0.312} & \hl{53}{0.105} & \hl{12}{6.607} & \hl{12}{67.0} \\
        PixelGS + DS & \hl{45}{25.59} & \hl{30}{0.824} & \hl{25}{0.341} & \hl{29}{0.112} & \hl{53}{1.443} & \hl{53}{28.8} \\
        Ours & \hl{60}{25.81} & \hl{48}{0.830} & \hl{38}{0.330} & \hl{50}{0.106} & \hl{55}{1.133} & \hl{52}{29.9} \\
        \bottomrule
    \end{tabular}
\end{table}

\subsection{Impact of the Average Contribution Criterion}
We now compare densification criteria directly, holding the occlusion-aware filter and all other components fixed and varying only the criterion. 
We examine four criteria: 
the occlusion-aware observation count, the PixelGS~\cite{zhang2024pixelgs} pixel-count-weighted mean without and with distance scaling (PixelGS, PixelGS + DS), and the average contribution (Ours). 
All criteria run at a single fixed threshold across every scene and dataset: 
the competitors at their published defaults, ours calibrated once on the Mip-NeRF360 sweep subset (Sec.~F.1) and left unchanged elsewhere. 
\Cref{tab:method:criterion} reports the four criteria on the occlusion-heavy Zip-NeRF scenes (Mip-NeRF360 counterpart in Tab.~S12). 
On the smaller Mip-NeRF360 scenes the criterion barely affects quality (across the four criteria, PSNR spans $0.13$\,dB and LPIPS $3\%$), so compactness is decisive: 
our average-contribution criterion produces the fewest Gaussians (roughly $55$--$75\%$ of the other criteria's counts) at matched quality. 
The criteria differ sharply on the large Zip-NeRF scenes. 
The raw pixel-count criterion is poorly behaved on its own: 
it over-densifies to $6.6$M Gaussians, by far the largest count, yet reaches the worst PSNR. 
Under a long training schedule it diverges catastrophically (\cref{sec:exp:largescale}). 
PixelGS therefore couples the criterion with distance scaling as a guard rail against runaway densification. 
This bounds the count, but on these large multi-room scenes the guard suppresses densification that is actually needed: 
quality degrades to the worst SSIM, LPIPS, and GMSD of the four. 
We attribute this to the reliance of distance scaling on the scene extent. 
We find the scene extent to be unreliable in general but especially on large scenes (\cref{sec:exp:implementation}). 
Our method is instead fully independent of it. 
Our average-contribution criterion needs no guard rail: 
it is intrinsically self-limiting, giving the fewest Gaussians of the four at competitive quality. 
The occlusion-aware count is $0.5\%$, $3\%$, and $3\%$ better on SSIM, LPIPS, and GMSD and $0.06$\,dB worse on PSNR, at $2.3\times$ our count. 
This demand is structural: 
given more densification opportunity, it keeps growing until training diverges, whereas ours stays bounded (\cref{sec:exp:largescale}).
\begin{table}[t]
    \centering
    \caption{Densification criteria on the Zip-NeRF scenes under longer schedules ($100$k and $210$k iterations; single runs). Runs that crash from over-densification are marked as divergent. Counts are in millions of Gaussians; time in minutes.}
    \label{tab:exp:largescale-zip}
    \begin{tabular}{cl*{6}{c}}
        \toprule
        & Method & PSNR$\uparrow$ & SSIM$\uparrow$ & LPIPS$\downarrow$ & GMSD$\downarrow$ & Count$\downarrow$ & Time$\downarrow$ \\
        \midrule
        \multirow{4}{*}{\rotatebox[origin=c]{90}{$100$k}} & Occlusion-aware count & \hl{12}{25.90} & \hl{12}{0.842} & \hl{60}{0.292} & \hl{12}{0.095} & \hl{12}{4.792} & \hl{12}{178.5} \\
        & PixelGS & \multicolumn{6}{c}{diverges ($>16.8$M Gaussians, training crashes)} \\
        & PixelGS + DS & \hl{49}{26.35} & \hl{22}{0.843} & \hl{12}{0.302} & \hl{24}{0.094} & \hl{44}{2.843} & \hl{44}{129.5} \\
        & Ours & \hl{60}{26.48} & \hl{60}{0.847} & \hl{46}{0.295} & \hl{60}{0.091} & \hl{60}{1.890} & \hl{60}{105.6} \\
        \midrule
        \multirow{3}{*}{\rotatebox[origin=c]{90}{$210$k}} & Occlusion-aware count & \multicolumn{6}{c}{diverges on 2 of 4 scenes (alameda, nyc)} \\
        & PixelGS + DS & \hl{12}{26.46} & \hl{12}{0.848} & \hl{12}{0.288} & \hl{12}{0.090} & \hl{12}{4.020} & \hl{12}{328.4} \\
        & Ours & \hl{60}{26.60} & \hl{60}{0.852} & \hl{60}{0.282} & \hl{60}{0.087} & \hl{60}{2.495} & \hl{60}{262.9} \\
        \bottomrule
    \end{tabular}
\end{table}

\begin{figure}[t]
    \centering
    \setlength{\tabcolsep}{1pt}
    \begin{tabular}{cccccc}
        \scriptsize Test view (GT) & \scriptsize GT crop & \scriptsize StopThePop & \scriptsize PixelGS & \scriptsize Grubert \etal & \scriptsize Ours \\
        \includegraphics[width=0.245\linewidth]{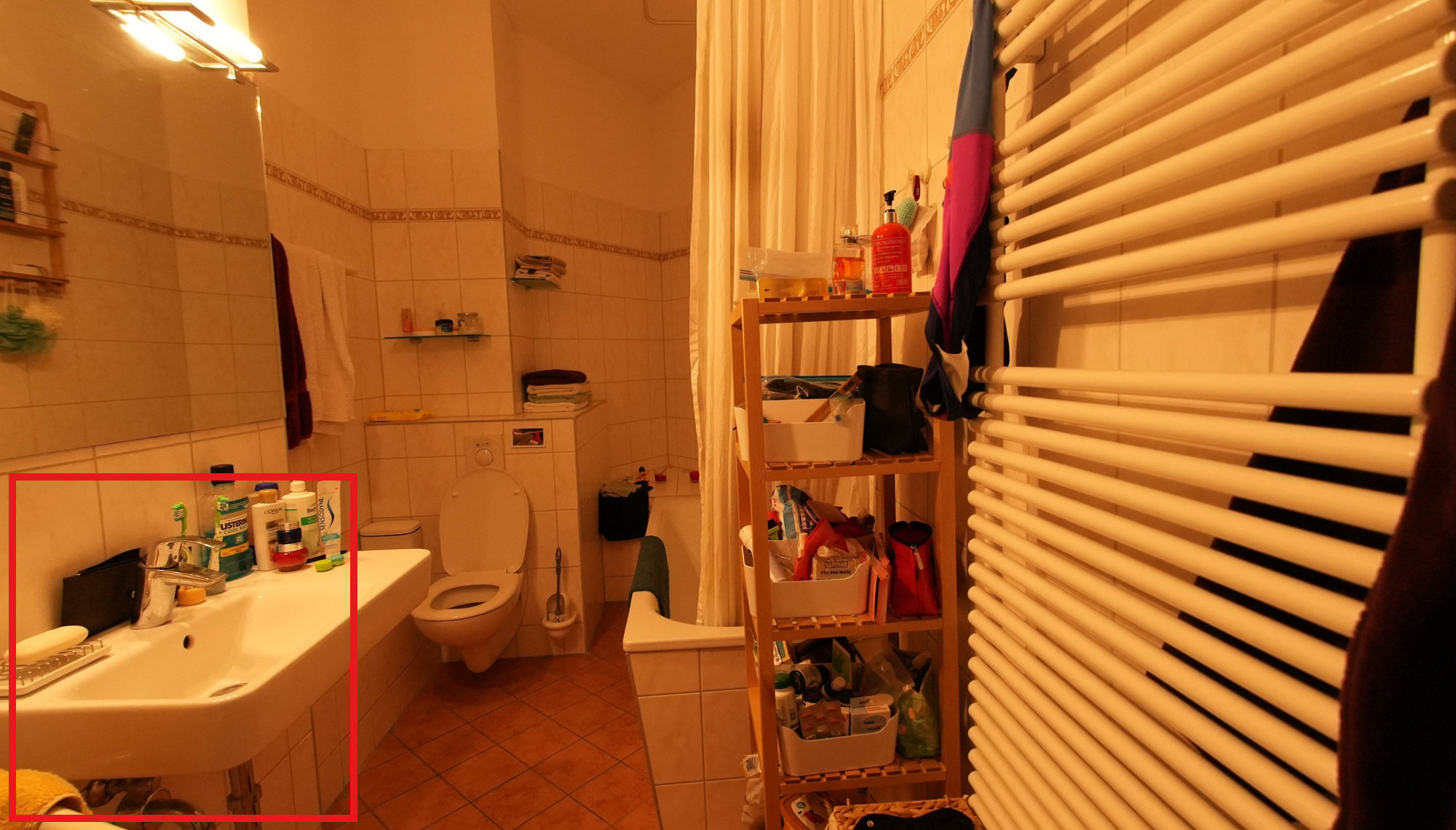} &
        \includegraphics[width=0.142\linewidth,trim=12 12 1510 650,clip]{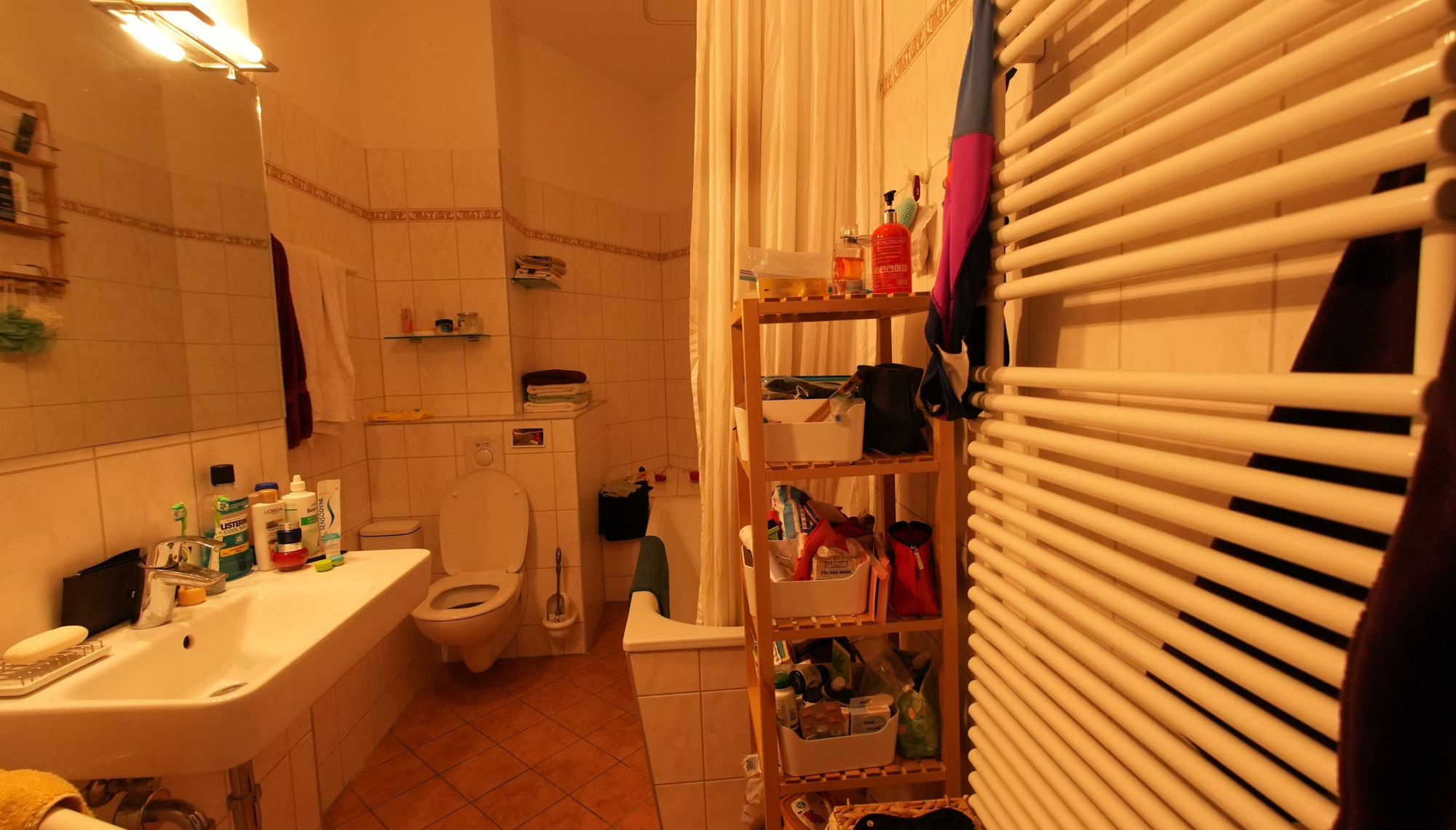} &
        \includegraphics[width=0.142\linewidth,trim=12 12 1510 650,clip]{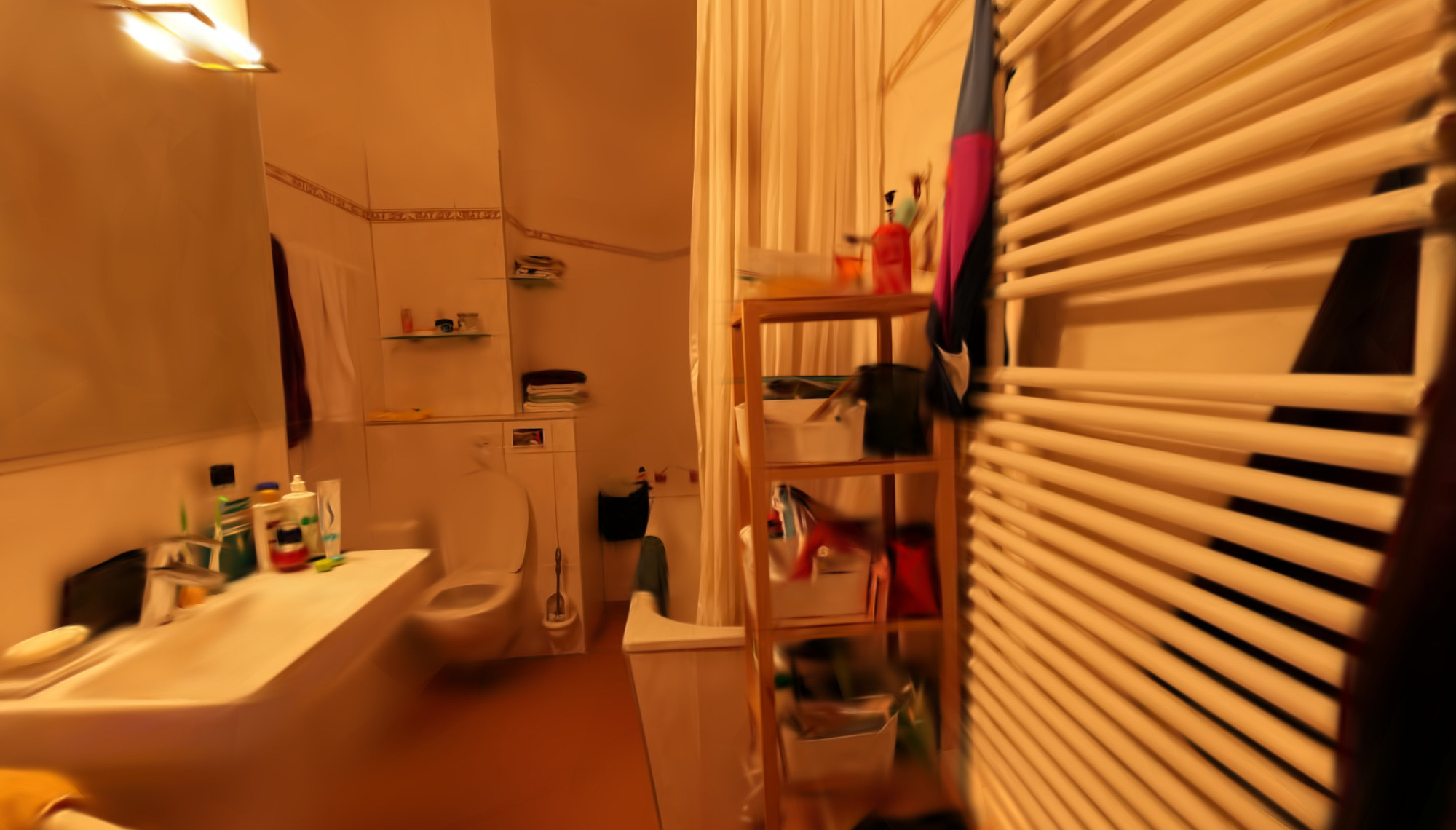} &
        \includegraphics[width=0.142\linewidth,trim=12 12 1510 650,clip]{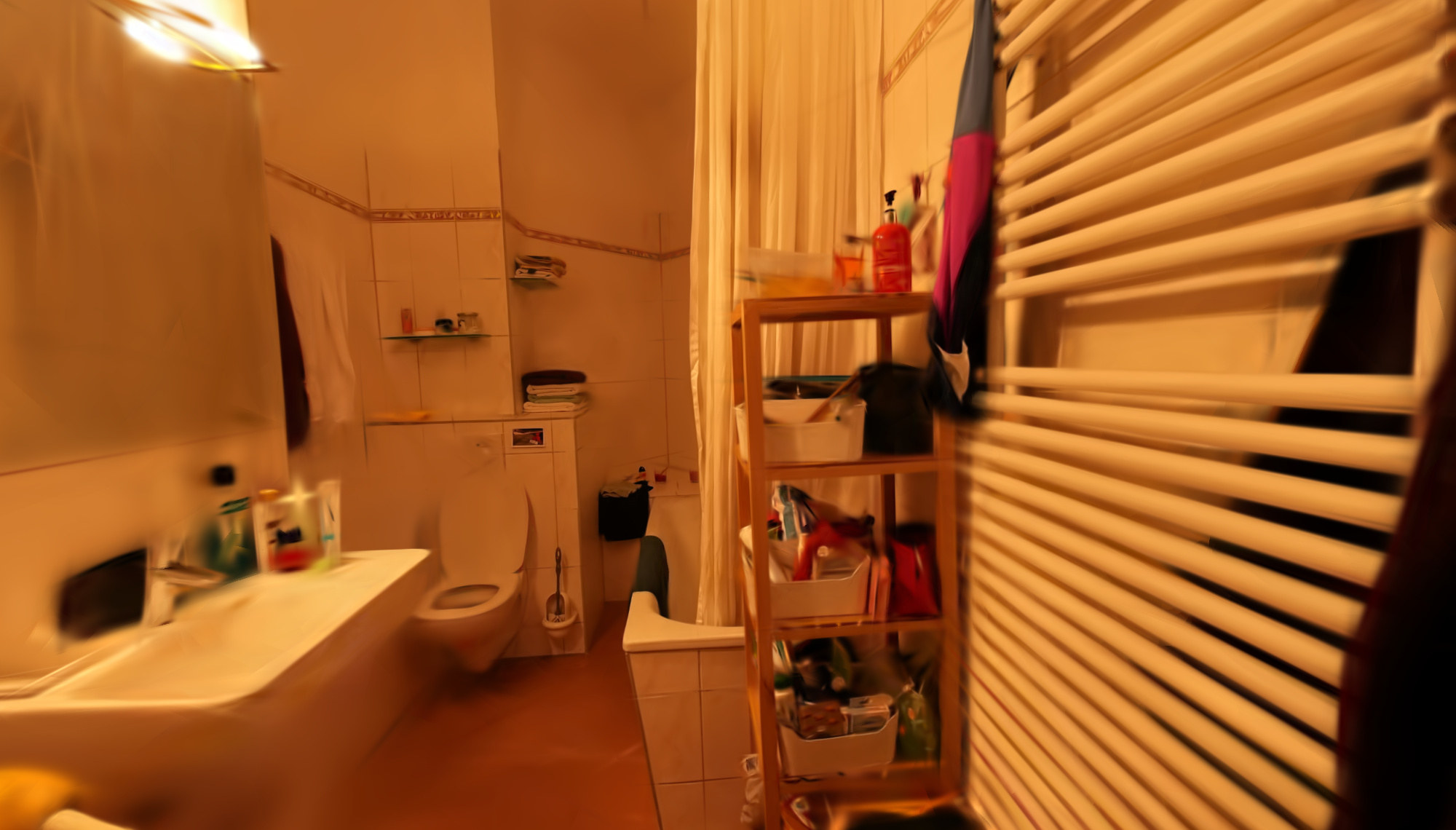} &
        \includegraphics[width=0.142\linewidth,trim=12 12 1510 650,clip]{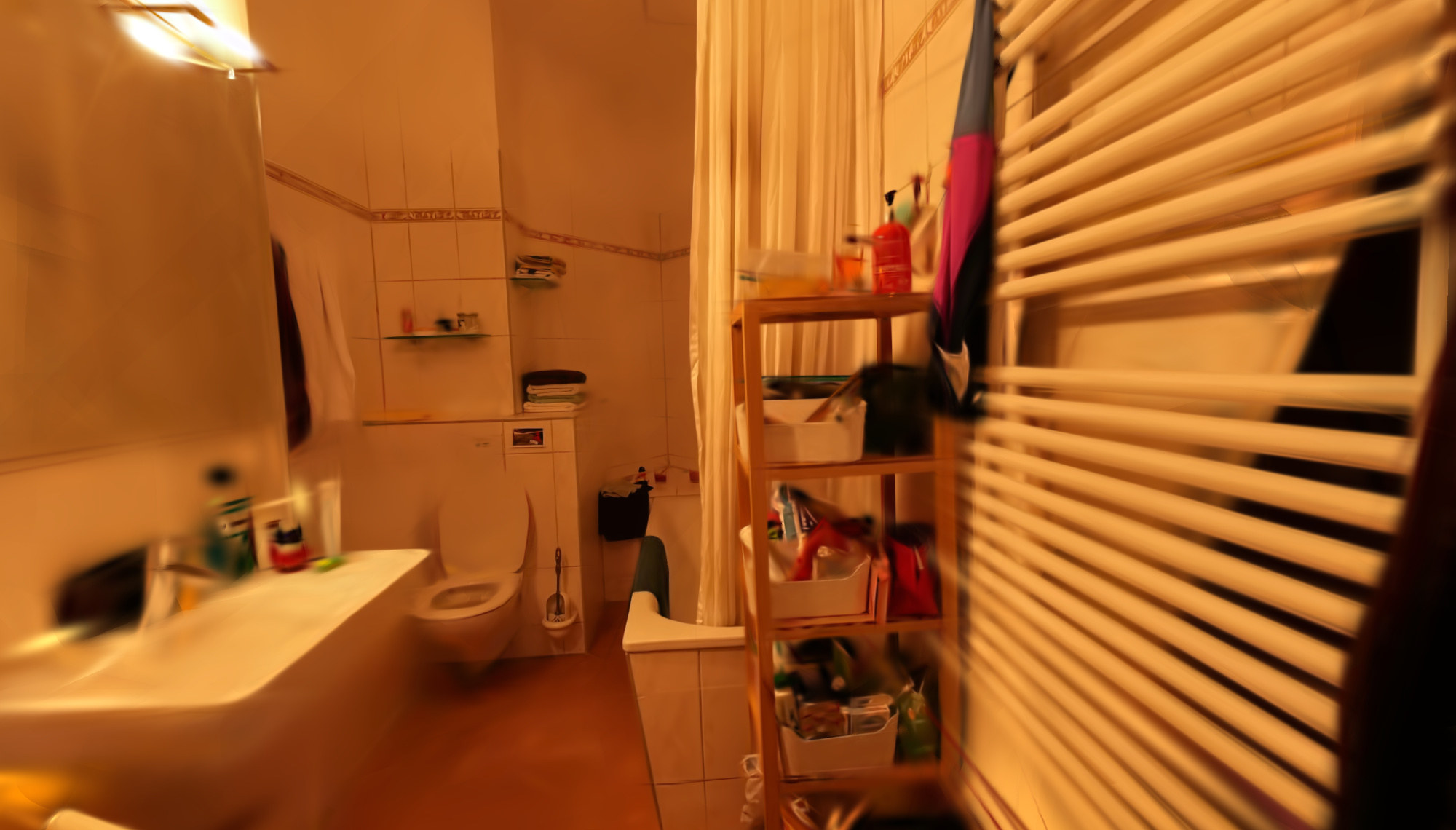} &
        \includegraphics[width=0.142\linewidth,trim=12 12 1510 650,clip]{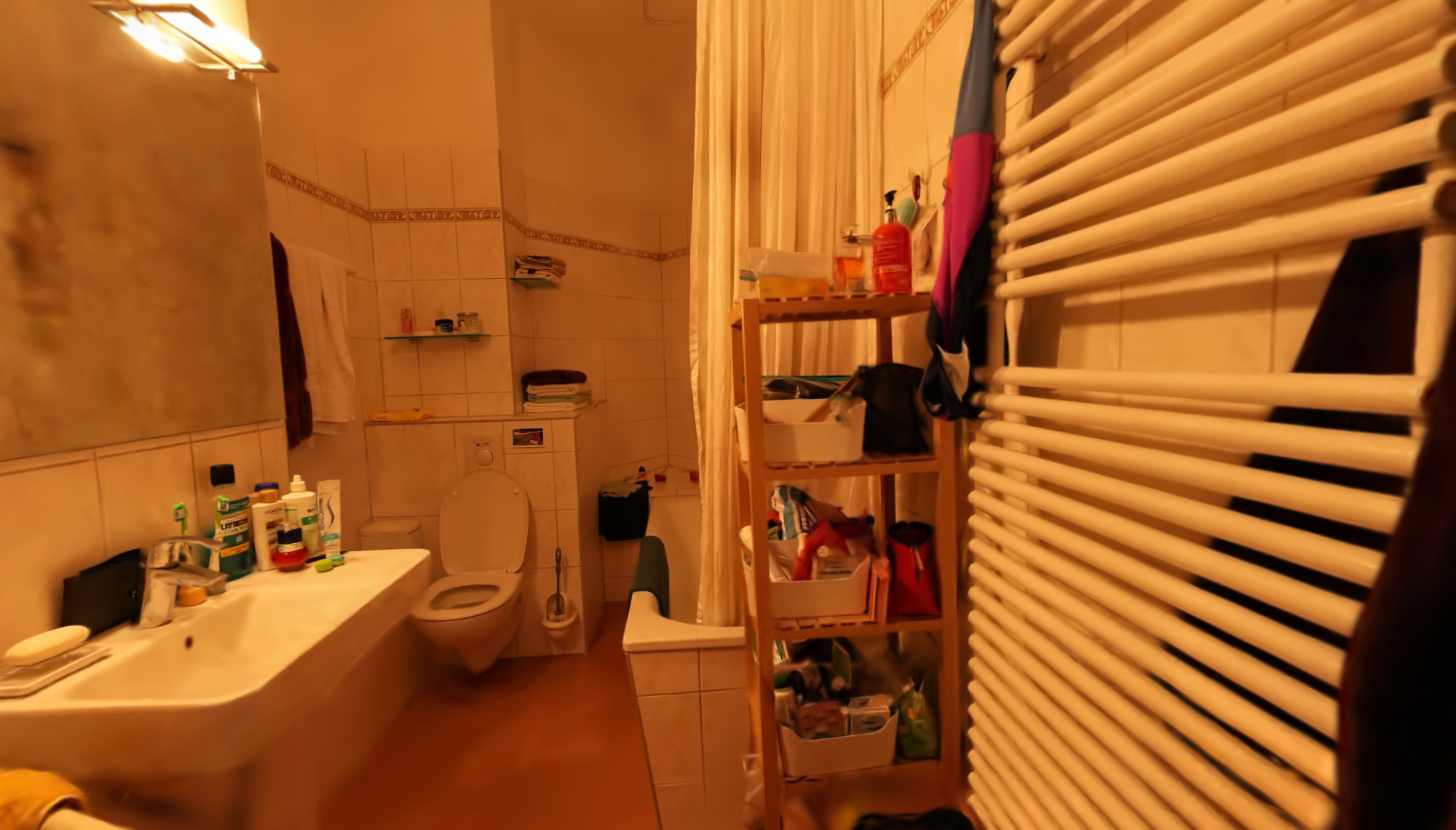} \\[1pt]
        \includegraphics[width=0.245\linewidth]{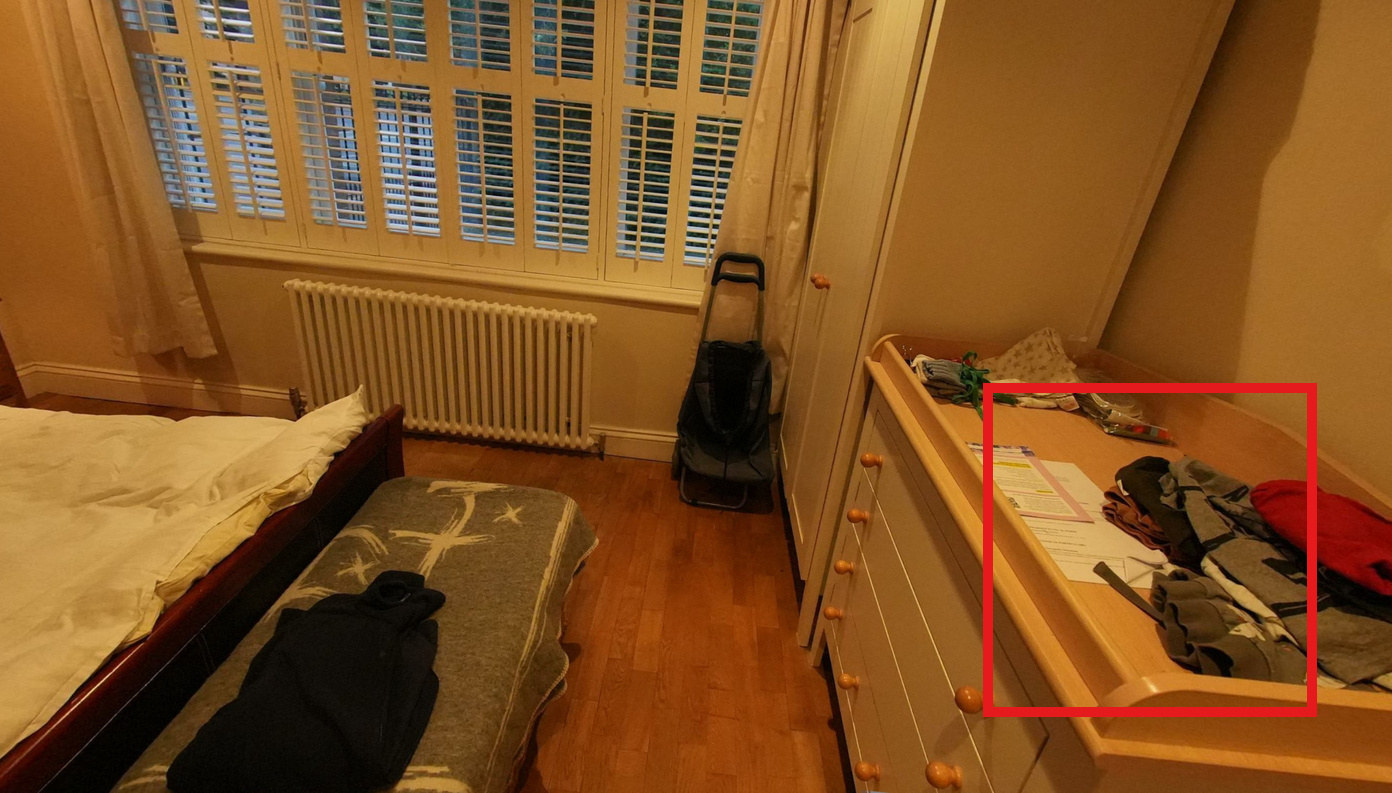} &
        \includegraphics[width=0.142\linewidth,trim=983 77 76 383,clip]{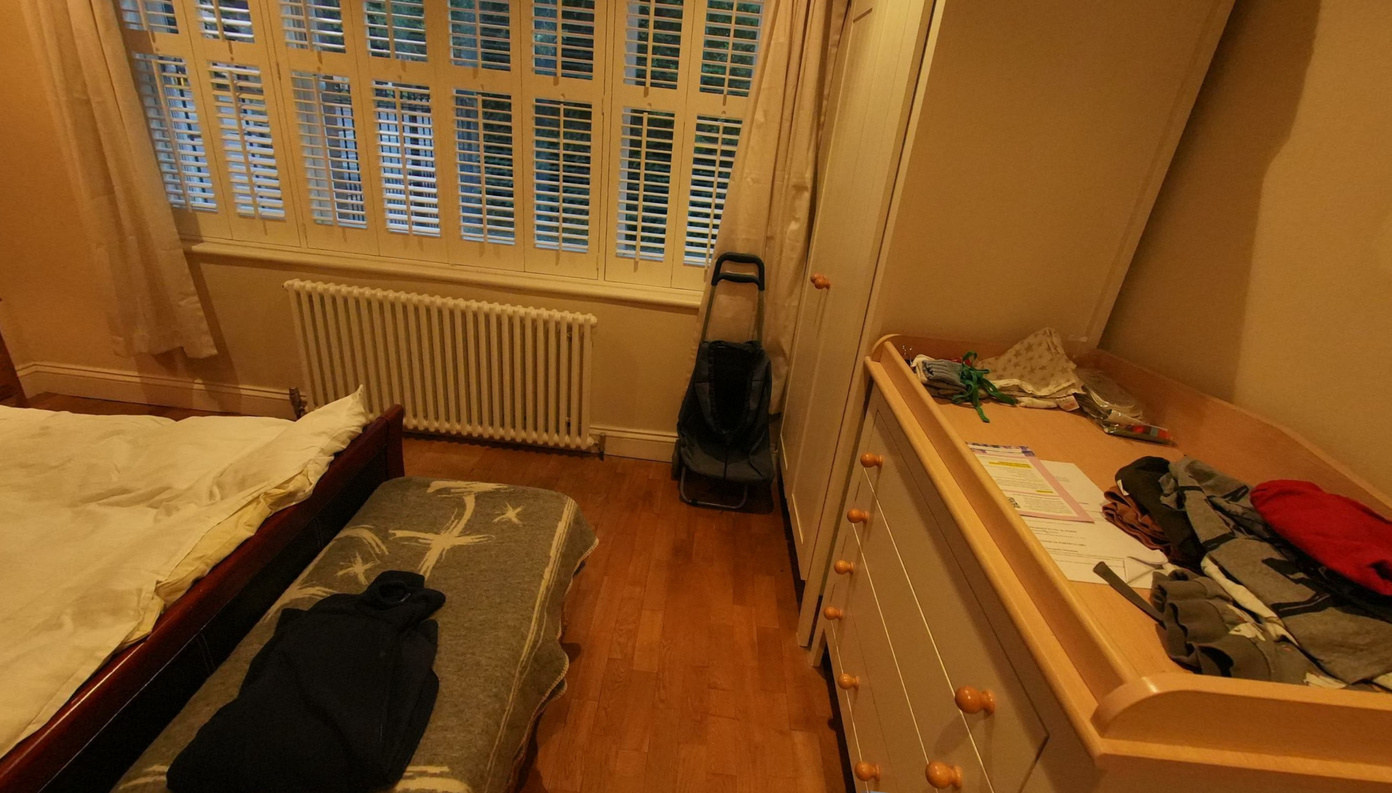} &
        \includegraphics[width=0.142\linewidth,trim=983 77 76 383,clip]{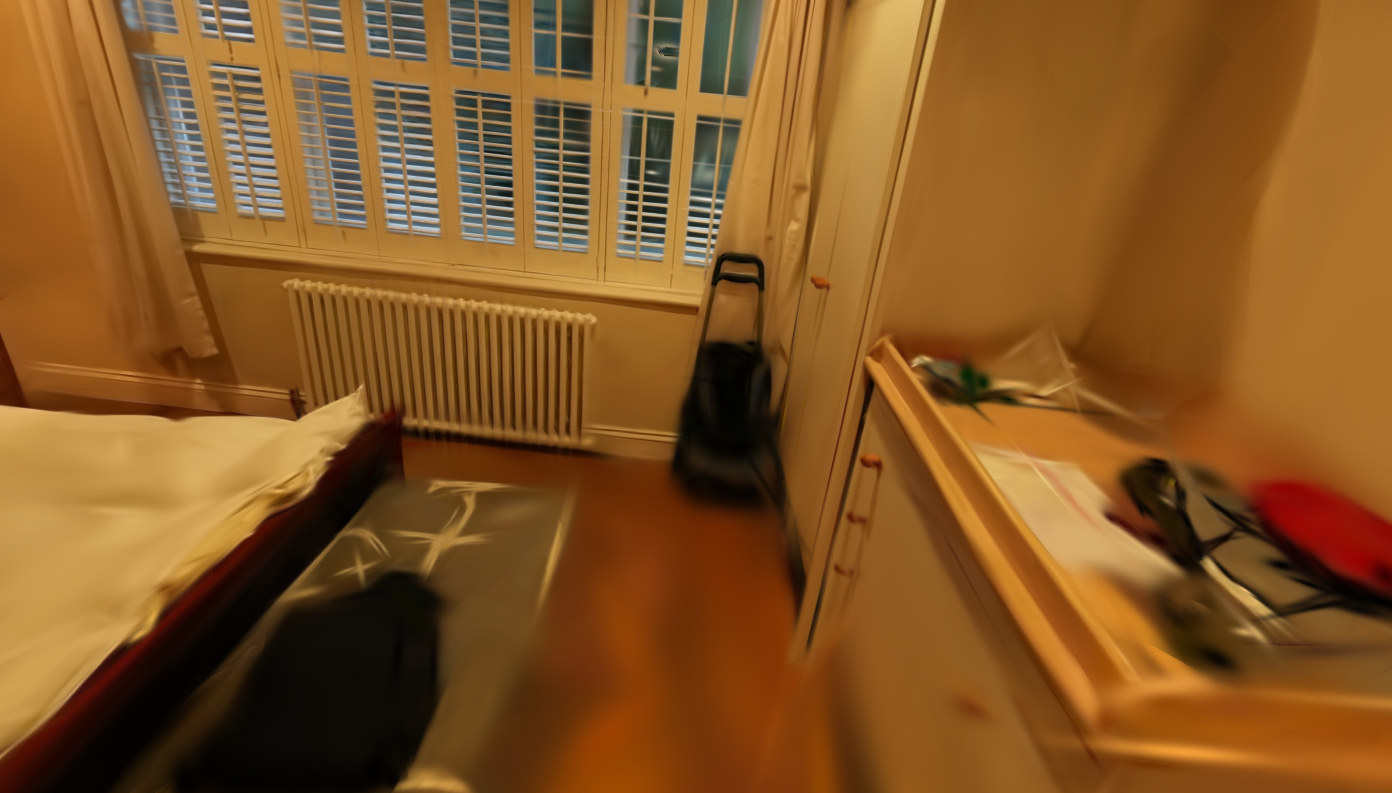} &
        \includegraphics[width=0.142\linewidth,trim=983 77 76 383,clip]{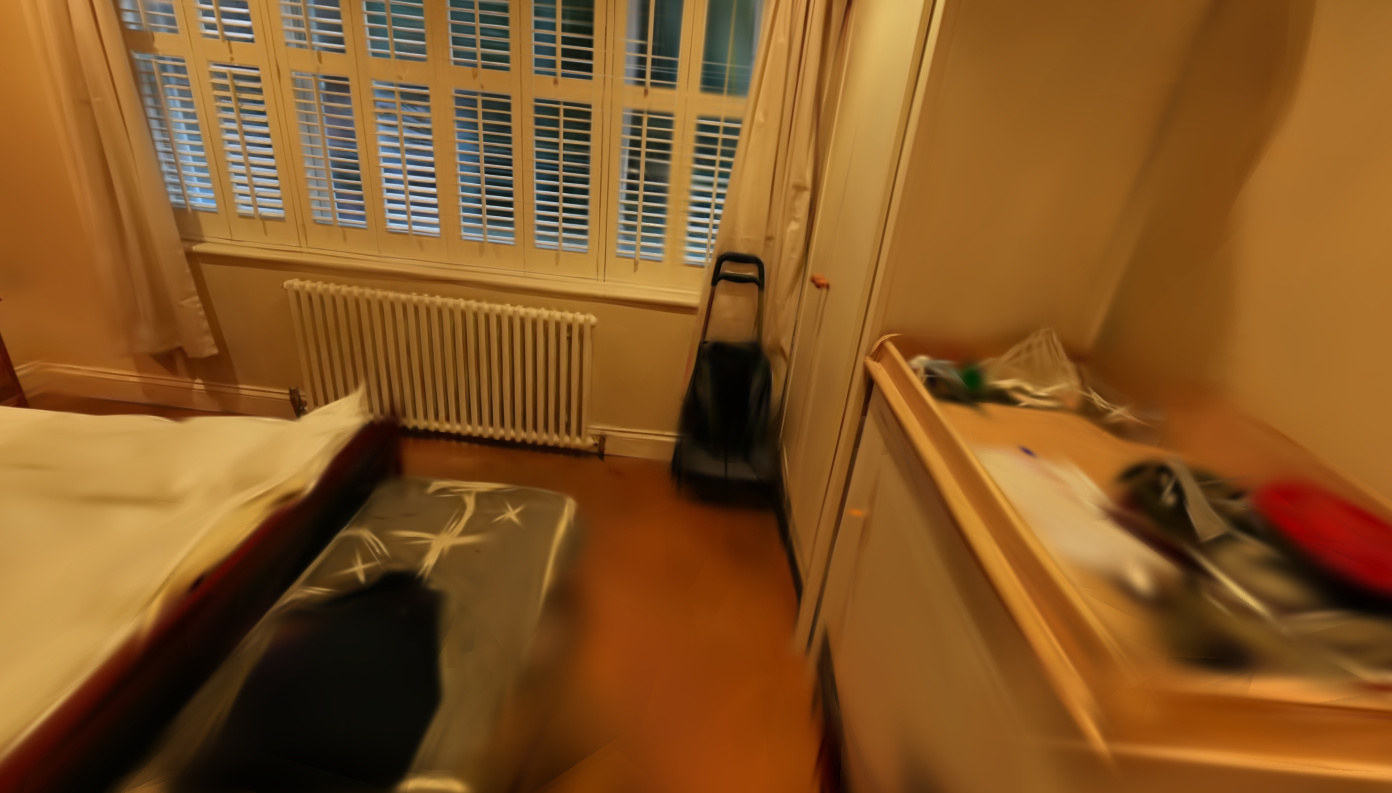} &
        \includegraphics[width=0.142\linewidth,trim=983 77 76 383,clip]{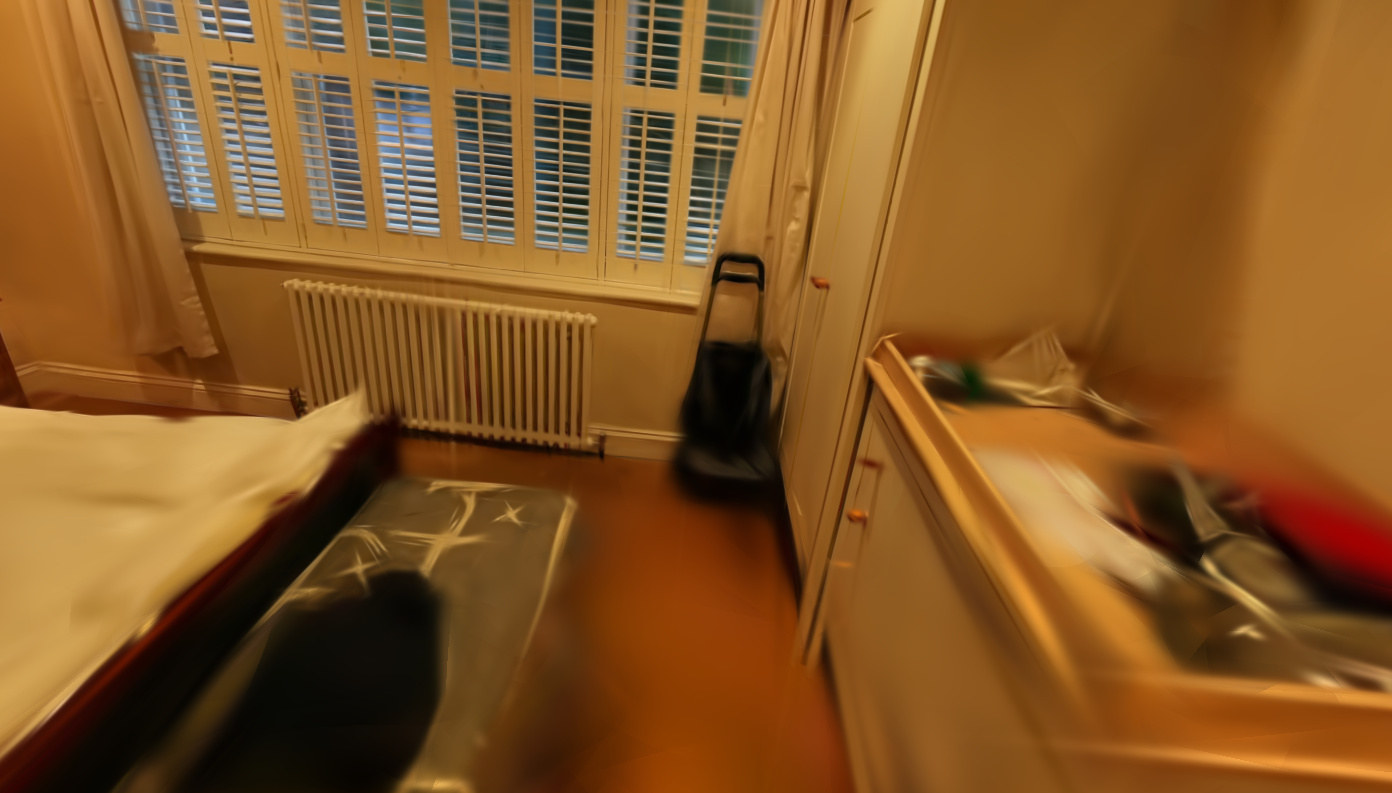} &
        \includegraphics[width=0.142\linewidth,trim=983 77 76 383,clip]{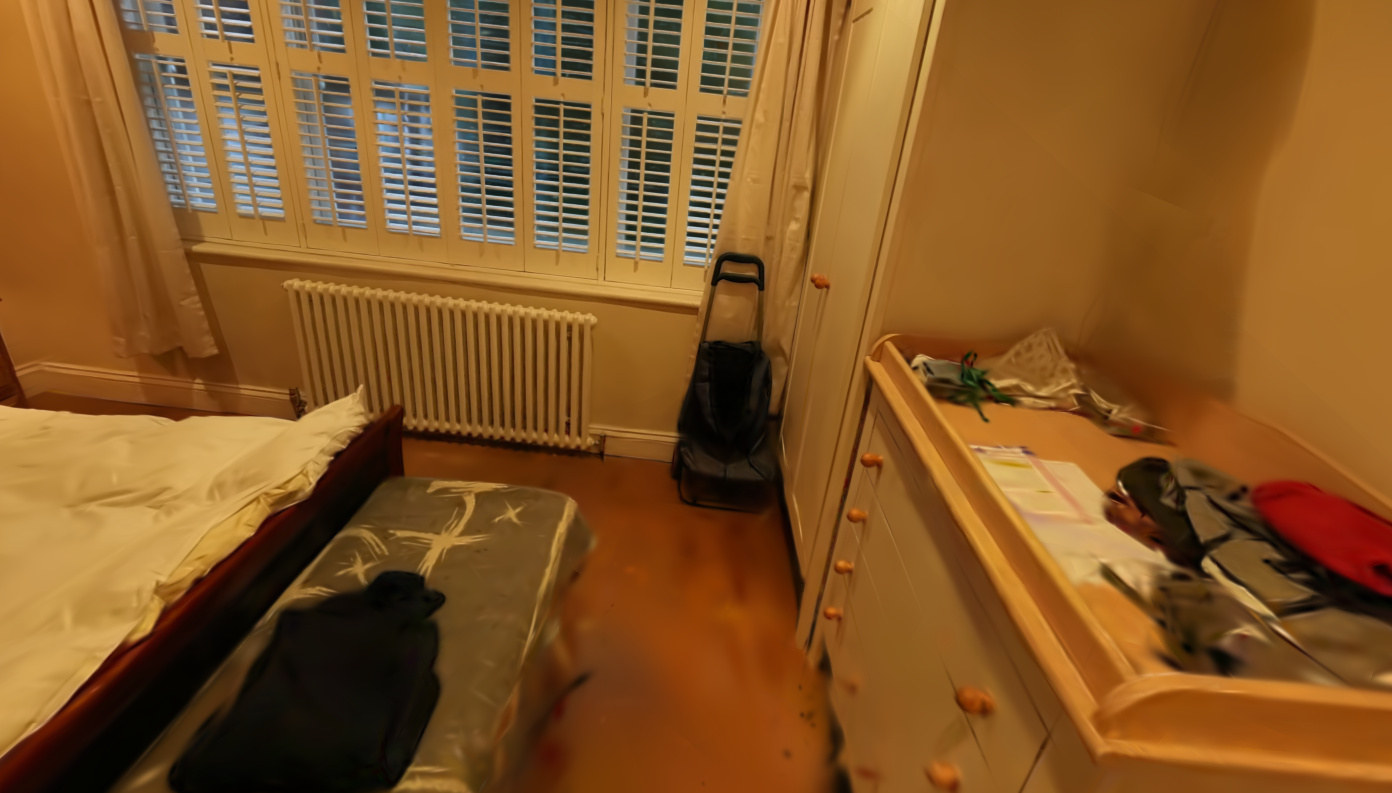} \\[1pt]
        \includegraphics[width=0.245\linewidth]{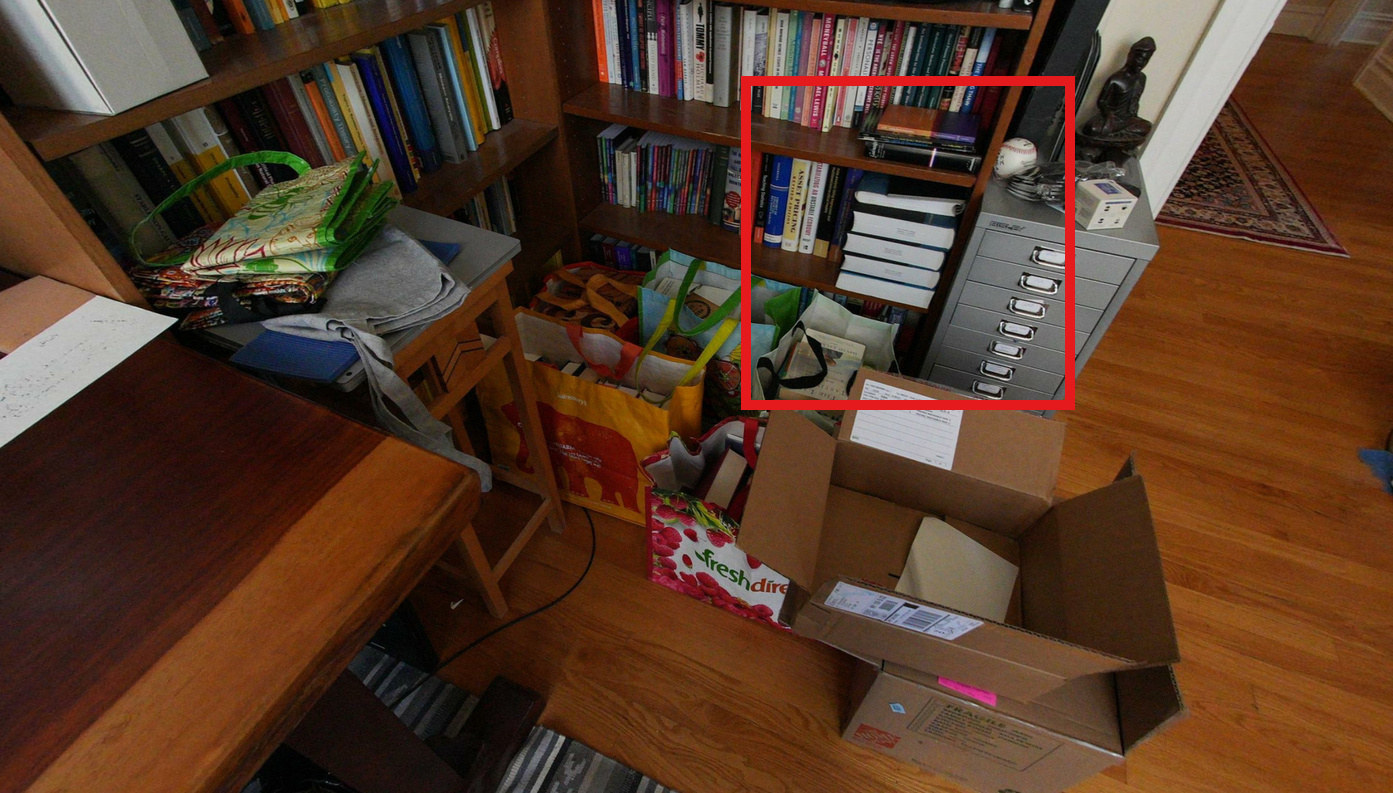} &
        \includegraphics[width=0.142\linewidth,trim=741 384 319 76,clip]{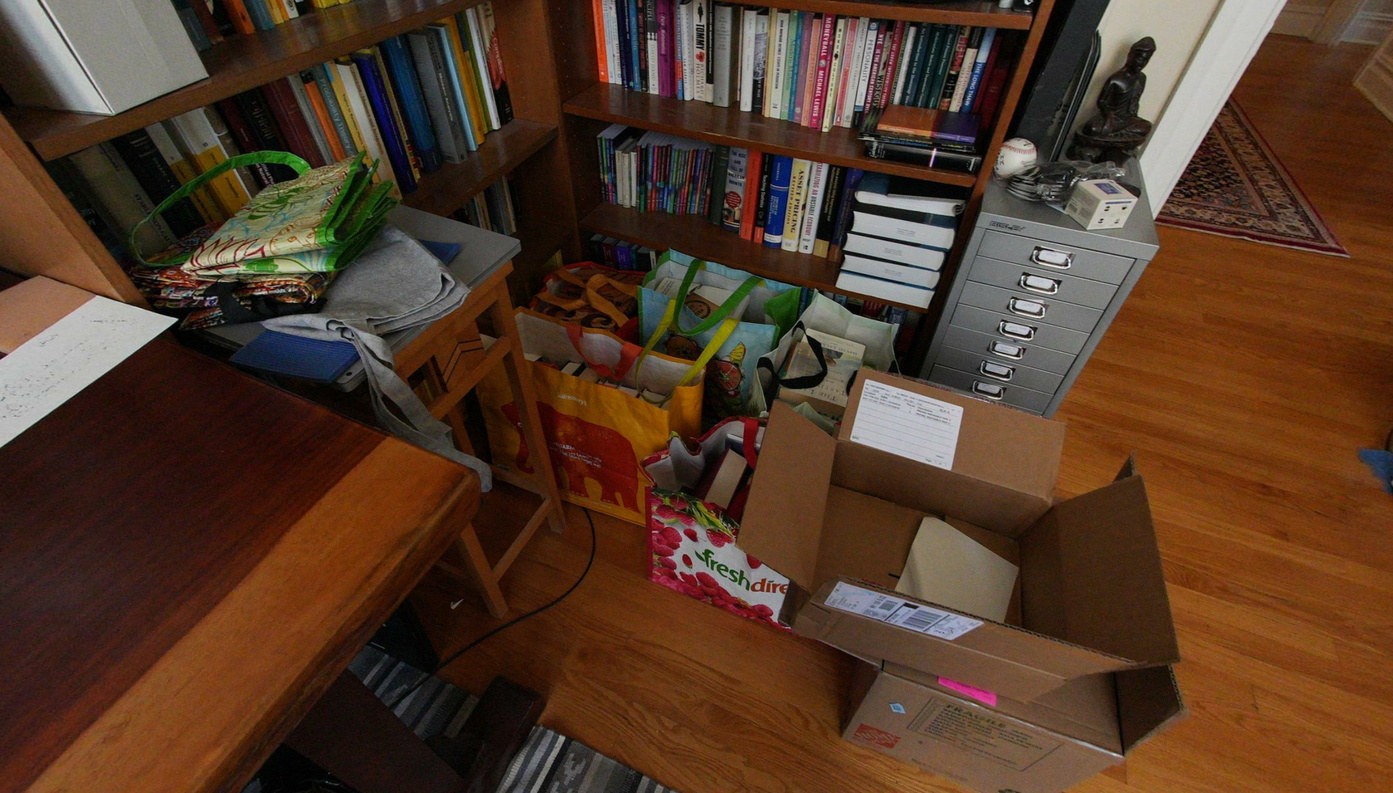} &
        \includegraphics[width=0.142\linewidth,trim=741 384 319 76,clip]{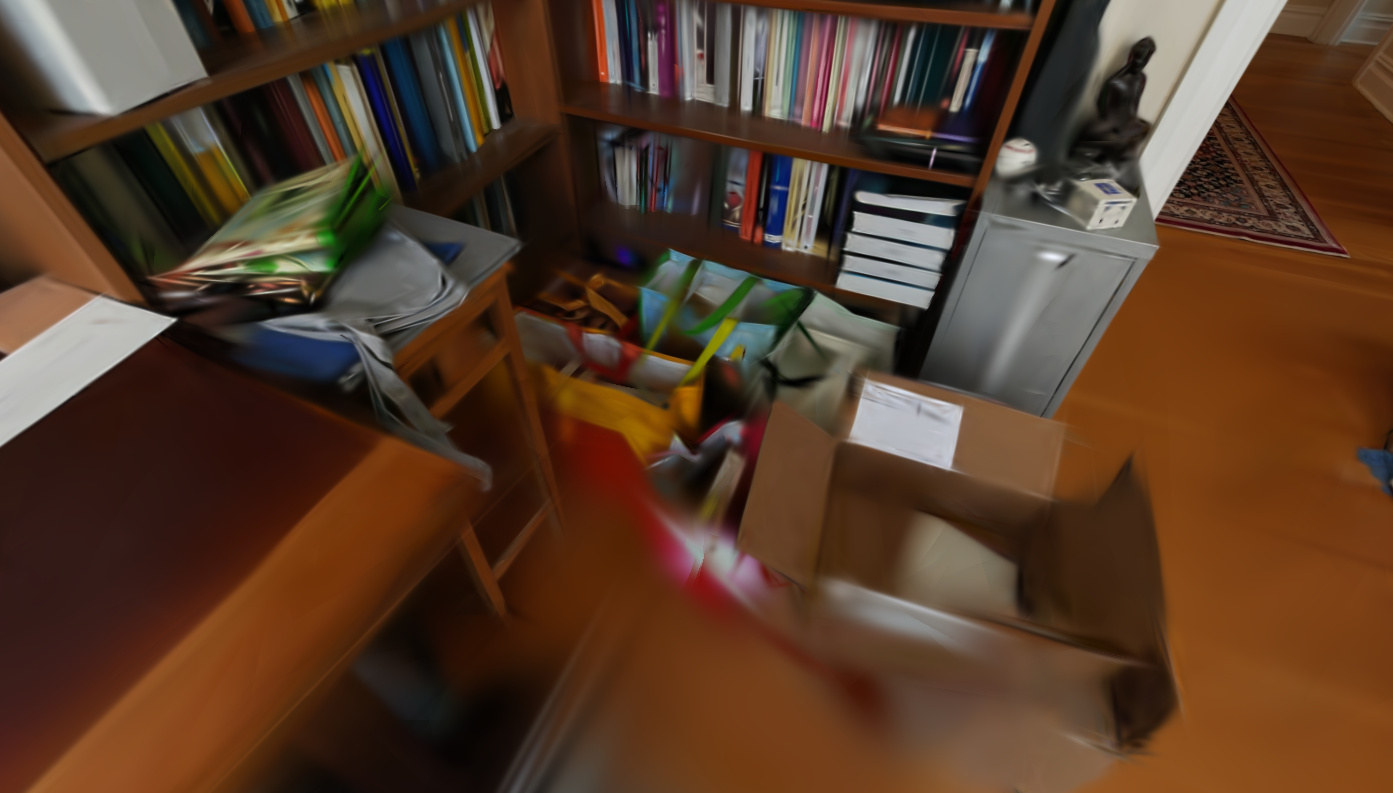} &
        \includegraphics[width=0.142\linewidth,trim=741 384 319 76,clip]{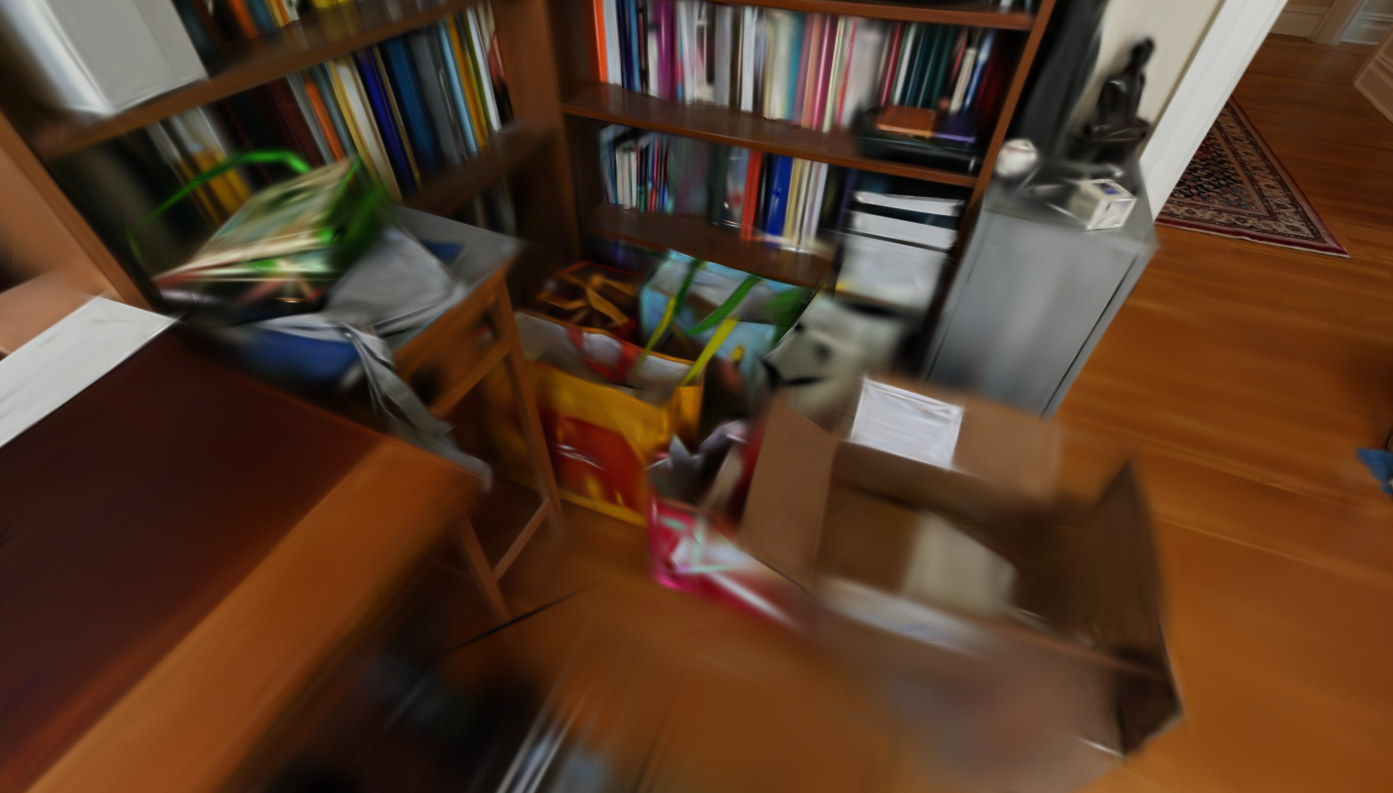} &
        \includegraphics[width=0.142\linewidth,trim=741 384 319 76,clip]{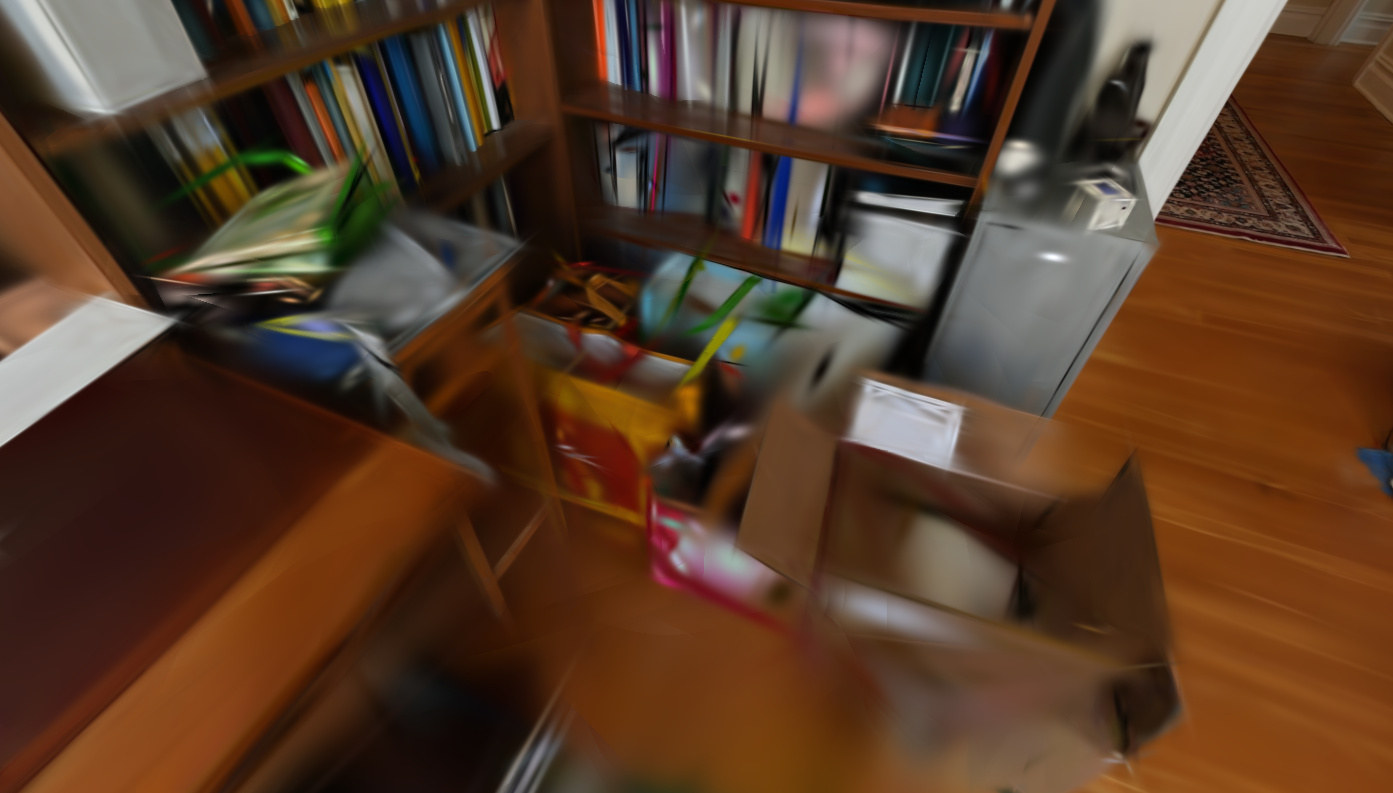} &
        \includegraphics[width=0.142\linewidth,trim=741 384 319 76,clip]{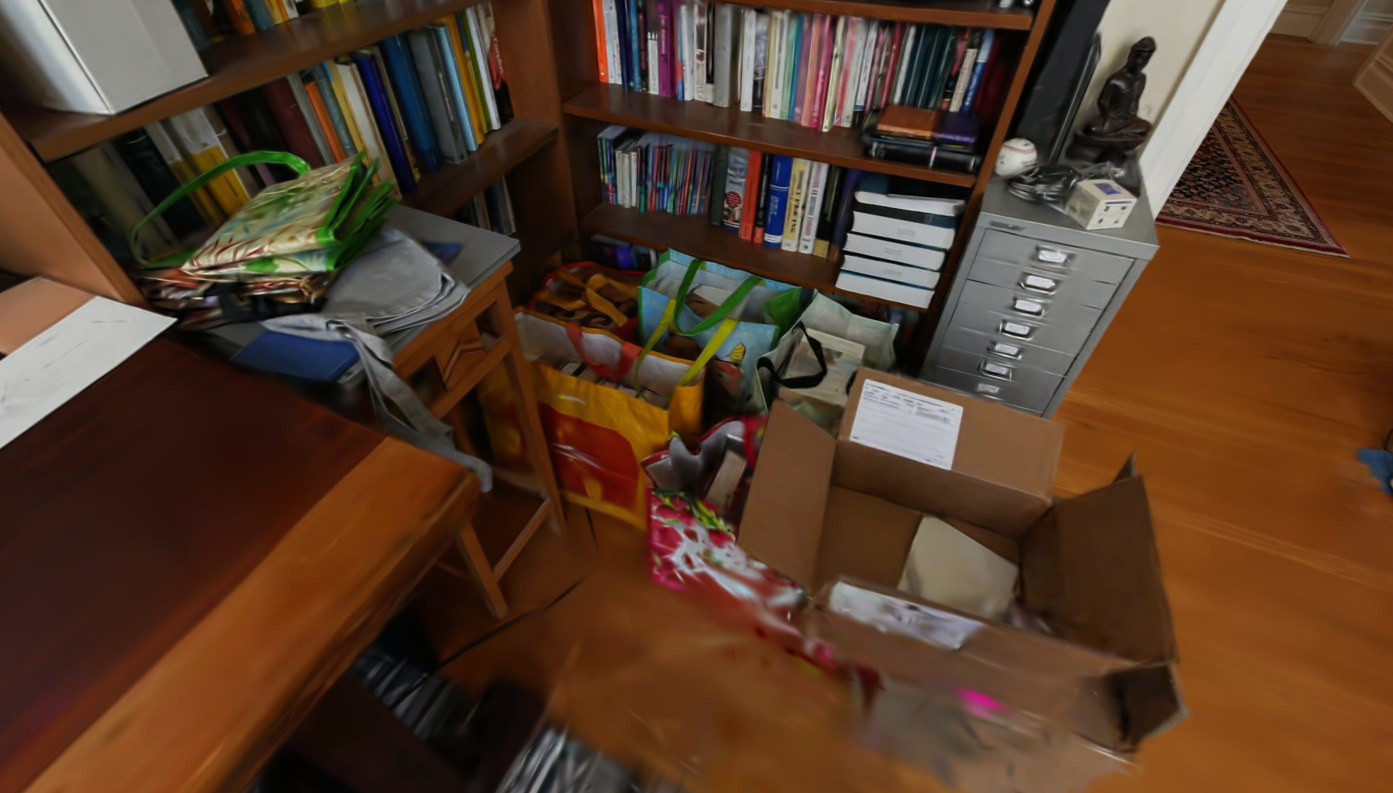} \\[1pt]
    \end{tabular}
    \caption{Qualitative comparison on Zip-NeRF scenes (berlin, london, nyc; $30$k schedule): StopThePop~\cite{radl2024stopthepop}, our re-implementations of PixelGS~\cite{zhang2024pixelgs} and Grubert \etal~\cite{Grubert2025improving} on the StopThePop base, and our full model. The crops show the areas marked with red boxes in the test view. The state-of-the-art methods remain blurry in these occlusion-heavy scenes, while ours reconstructs fine geometry.}
    \label{fig:exp:qualitative-zip}
\end{figure}

\subsection{Scalability under a Matched Training Budget} \label{sec:exp:largescale}
The large Zip-NeRF scenes are unevenly observed: 
training cameras cluster in interior hotspots and cover peripheral regions far more sparsely (camera map for the alameda example in Fig.~S3). 
They also hold about seven times as many training images as Mip-NeRF360, so the default $30$k schedule grants each image only $22$ optimization steps against Mip-NeRF360's $159$, and lets it contribute to only $11$ densification events against $77$. 
Sparsely covered regions therefore receive few densification opportunities, and blur there can signal missing opportunity rather than poor densification decisions. 
Longer schedules restore this opportunity and reveal whether a criterion resolves the remaining detail or leaves these regions under-densified, and whether its growth stays bounded while doing so. 
We therefore additionally train the large Zip-NeRF scenes under a matched schedule of $210$k iterations, which restores both quantities ($157$ steps and $76$ densification events per image), and sample around it at $100$k and $300$k. 
Every other result keeps the standard $30$k protocol for comparability. 
\Cref{tab:exp:largescale-zip} shows the results up to the matched schedule; the $300$k table and schedule details are in Sec.~G.2.
\begin{table*}[t]
    \centering
    \caption{Comparison on Deep Blending, Mip-NeRF360, Tanks \& Temples and Zip-NeRF. We list results as reported by Grubert \etal~\cite{Grubert2025improving} ($\dagger$) and RadSplat~\cite{niemeyer2025radsplat} ($\ddagger$). All other results are from our own experiments on the shared StopThePop (STP) base. $G$ is Gaussians in millions; rows are grouped by base, with RadSplat and Zip-NeRF as upper-bound references (excluded from the color scale).}
    \label{tab:exp:sota}
    \setlength{\tabcolsep}{3pt}
    \small
    \resizebox{\textwidth}{!}{
        \begin{tabular}{cl *{16}{c}}
            \toprule
            & Dataset & \multicolumn{4}{c}{Deep Blending}
                    & \multicolumn{4}{c}{Mip-NeRF360}
                    & \multicolumn{4}{c}{Tanks \& Temples}
                    & \multicolumn{4}{c}{Zip-NeRF} \\
            \cmidrule(lr){3-6}\cmidrule(lr){7-10}\cmidrule(lr){11-14}\cmidrule(lr){15-18}
            & Metric
             & PSNR$\uparrow$ & SSIM$\uparrow$ & LPIPS$\downarrow$ & G(M)$\downarrow$
             & PSNR$\uparrow$ & SSIM$\uparrow$ & LPIPS$\downarrow$ & G(M)$\downarrow$
             & PSNR$\uparrow$ & SSIM$\uparrow$ & LPIPS$\downarrow$ & G(M)$\downarrow$
             & PSNR$\uparrow$ & SSIM$\uparrow$ & LPIPS$\downarrow$ & G(M)$\downarrow$ \\
            \midrule
            \multirow{2}{*}{\rotatebox[origin=c]{90}{3DGS}} & PixelGS$^\dagger$
             & \hl{12}{28.95} & \hl{12}{0.892} & \hl{12}{0.250} & \hl{12}{$\sim$4.6}
             & \hl{48}{27.53} & \hl{54}{0.822} & \hl{52}{0.191} & \hl{12}{$\sim$5.5}
             & \hl{45}{23.83} & \hl{46}{0.854} & \hl{46}{0.151} & \hl{12}{$\sim$4.5}
             & \hl{0}{N/A} & \hl{0}{N/A} & \hl{0}{N/A} & \hl{0}{N/A} \\
             & Grubert \etal$^\dagger$
             & \hl{42}{29.63} & \hl{49}{0.902} & \hl{37}{0.240} & \hl{42}{$\sim$2.3}
             & \hl{60}{27.71} & \hl{60}{0.824} & \hl{50}{0.193} & \hl{40}{$\sim$3.3}
             & \hl{60}{24.18} & \hl{60}{0.860} & \hl{48}{0.149} & \hl{40}{$\sim$2.4}
             & \hl{0}{N/A} & \hl{0}{N/A} & \hl{0}{N/A} & \hl{0}{N/A} \\
            \midrule
            \multirow{6}{*}{\rotatebox[origin=c]{90}{StopThePop}} & STP (Reset)
             & \hl{54}{29.91} & \hl{56}{0.904} & \hl{52}{0.234} & \hl{35}{2.819}
             & \hl{32}{27.30} & \hl{28}{0.814} & \hl{27}{0.214} & \hl{40}{3.289}
             & \hl{15}{23.15} & \hl{21}{0.843} & \hl{24}{0.173} & \hl{47}{1.800}
             & \hl{20}{25.11} & \hl{21}{0.808} & \hl{21}{0.366} & \hl{45}{0.951} \\
             & PixelGS
             & \hl{38}{29.54} & \hl{45}{0.901} & \hl{52}{0.234} & \hl{15}{4.392}
             & \hl{37}{27.37} & \hl{54}{0.822} & \hl{53}{0.190} & \hl{14}{5.328}
             & \hl{20}{23.27} & \hl{42}{0.852} & \hl{50}{0.147} & \hl{16}{4.175}
             & \hl{23}{25.15} & \hl{26}{0.811} & \hl{26}{0.360} & \hl{12}{1.980} \\
             & Grubert \etal
             & \hl{60}{30.05} & \hl{60}{0.905} & \hl{57}{0.232} & \hl{43}{2.209}
             & \hl{52}{27.59} & \hl{60}{0.824} & \hl{51}{0.192} & \hl{40}{3.263}
             & \hl{42}{23.78} & \hl{55}{0.858} & \hl{51}{0.146} & \hl{42}{2.231}
             & \hl{29}{25.27} & \hl{27}{0.812} & \hl{26}{0.360} & \hl{39}{1.131} \\
             & Ours (Reset)
             & \hl{53}{29.90} & \hl{49}{0.902} & \hl{60}{0.231} & \hl{50}{1.656}
             & \hl{40}{27.42} & \hl{38}{0.817} & \hl{60}{0.184} & \hl{32}{3.913}
             & \hl{15}{23.16} & \hl{33}{0.848} & \hl{57}{0.140} & \hl{20}{3.871}
             & \hl{60}{25.86} & \hl{60}{0.834} & \hl{60}{0.320} & \hl{12}{1.991} \\
            \cmidrule(lr){2-18}
             & STP (Decay)
             & \hl{49}{29.80} & \hl{56}{0.904} & \hl{32}{0.242} & \hl{56}{1.219}
             & \hl{12}{27.01} & \hl{12}{0.809} & \hl{12}{0.227} & \hl{60}{1.711}
             & \hl{12}{23.09} & \hl{12}{0.839} & \hl{12}{0.185} & \hl{60}{0.848}
             & \hl{12}{24.95} & \hl{12}{0.802} & \hl{12}{0.376} & \hl{60}{0.486} \\
             & Ours (Decay)
             & \hl{59}{30.03} & \hl{60}{0.905} & \hl{52}{0.234} & \hl{60}{0.894}
             & \hl{32}{27.30} & \hl{47}{0.820} & \hl{58}{0.186} & \hl{57}{1.954}
             & \hl{18}{23.22} & \hl{44}{0.853} & \hl{60}{0.137} & \hl{45}{1.965}
             & \hl{57}{25.81} & \hl{54}{0.830} & \hl{51}{0.330} & \hl{39}{1.133} \\
            \midrule
            \multirow{2}{*}{\rotatebox[origin=c]{90}{NeRF}} & RadSplat$^\ddagger$
             & \hl{99}{N/A} & \hl{99}{N/A} & \hl{99}{N/A} & \hl{99}{N/A}
             & \hl{99}{28.14} & \hl{99}{0.843} & \hl{99}{0.171} & \hl{99}{1.924}
             & \hl{99}{N/A} & \hl{99}{N/A} & \hl{99}{N/A} & \hl{99}{N/A}
             & \hl{99}{26.17} & \hl{99}{0.839} & \hl{99}{0.364} & \hl{99}{N/A} \\
             & Zip-NeRF$^\ddagger$
             & \hl{99}{N/A} & \hl{99}{N/A} & \hl{99}{N/A} & \hl{99}{-}
             & \hl{99}{28.54} & \hl{99}{0.836} & \hl{99}{0.177} & \hl{99}{-}
             & \hl{99}{N/A} & \hl{99}{N/A} & \hl{99}{N/A} & \hl{99}{-}
             & \hl{99}{27.37} & \hl{99}{0.836} & \hl{99}{0.305} & \hl{99}{-} \\
            \bottomrule
        \end{tabular}
    }
\end{table*}

Under these schedules the naive pixel-count criterion (PixelGS without distance scaling) is not merely inefficient but unstable: 
on every Zip-NeRF scene it over-densifies past the $\sim$$16.8$M trainable-count limit until training crashes (count-versus-iteration plot in Fig.~S4). 
That limit lies nearly an order of magnitude beyond ours. 
The full PixelGS baseline, which includes the distance-scaling guard rail, does not diverge at its default threshold; the instability is a property of the pixel-count criterion in isolation. 
Our average-contribution criterion stays bounded throughout ($2.75$M even at $300$k) and continues to improve in LPIPS and GMSD, whereas the guarded PixelGS remains behind on every image-quality metric while using $62\%$ more Gaussians.
Section~G.4 traces this remaining gap to the guard itself: 
the distance term systematically suppresses near-object detail. 
At $210$k, even the occlusion-aware count over-densifies past the same limit on the two occlusion-heaviest scenes. 
These divergences are mechanistic, not single-run artifacts: 
the Gaussian count grows steadily into the trainable limit. 
We therefore also report the $100$k schedule, the longest that the remaining count-based criteria complete, to quantify their high-count behavior. 
There, the extra Gaussians buy no quality advantage, repeating the Mip-NeRF360 pattern: 
ours leads PSNR, SSIM, and GMSD outright and stays within about $1\%$ LPIPS of the occlusion-aware count, at $39$--$67\%$ of the competing counts. 
Two supplementary analyses corroborate this. 
A threshold sweep (Sec.~F.5) confirms that the self-limiting behavior is not specific to the chosen threshold: 
at half the default, the count-based criteria reach roughly three times our count. 
A per-view analysis (Sec.~G.5) shows that our per-view quality stays as tight as the guarded PixelGS's, while the unguarded occlusion-aware count develops a growing low-quality tail. 

\subsection{Comparison to State-of-the-Art}
\Cref{tab:exp:sota} compares our method against StopThePop~\cite{radl2024stopthepop}, our shared-base re-implementations of PixelGS and Grubert \etal, and published reference results. 
Across all four datasets our method reaches competitive or better image quality while using far fewer Gaussians than the strongest competitors, improving reconstruction efficiency rather than trading count for quality. 
On the occlusion-heavy Zip-NeRF scenes, both our variants lead every image-quality metric among the methods trained on the shared base, with up to $12\%$ lower LPIPS than StopThePop and $8\%$ (decay) to $11\%$ (reset) better LPIPS than the best competitor. 
The decay variant achieves this at the same Gaussian count as Grubert \etal, the strongest competitor. 
\Cref{fig:exp:qualitative-zip} shows this qualitatively: 
in heavily occluded interior regions, the occlusion-blind methods remain visibly blurry, while our method resolves the fine geometry. 
Even the RadSplat upper-bound reference trails both our variants in LPIPS, by $9$--$12\%$. 
On Deep Blending it matches the best quality (within $0.02$\,dB PSNR, tied SSIM with Grubert \etal) at less than half the Gaussians. 
On Mip-NeRF360 it achieves better LPIPS than every StopThePop, PixelGS, and Grubert variant with its other metrics within about $2\%$ of the best. 
The decay variant uses $35$--$60\%$ of the PixelGS and Grubert counts. 
On Tanks \& Temples the competitors' paper-reported PSNR advantage shrinks substantially on the shared base (Grubert \etal remain $0.56$\,dB ahead). 
Ours still holds the best LPIPS and stays within roughly $1\%$ on SSIM, the pattern expected when detail is reconstructed but not perfectly pixel-aligned.
Across scenes, the occlusion rate predicts where our contributions help: it correlates with our LPIPS gain over the occlusion-blind variant (Spearman $+0.83$). 
Ours also allocates its Gaussians more efficiently:
even without the contribution prune, our models contain the smallest share of fully occluded Gaussians ($7.2\%$ on Zip-NeRF, against $10.5$--$22.7\%$ for the occlusion-blind methods; occluded-Gaussian counts per method in Tab.~S8). 
This is no artifact of lower counts: 
at a nearly matched count, our unpruned variant contains $43\%$ fewer occluded Gaussians than Grubert \etal's.

\subsection{Ablation Study}
We ablate our full model on the Zip-NeRF scenes by removing or reverting one component at a time (\cref{tab:exp:ablation-zip}). 
The isolation of our two contributions on vanilla StopThePop, and the clone failure it exposes, are analyzed in Sec.~C.2 (Tab.~S5) of the supplementary material. 
By far the most consequential component is split-only densification: 
restoring the original clone operation (\emph{+ clone}) collapses reconstruction quality (nearly $5$\,dB PSNR and $31\%$ LPIPS worse). 
Cloning depends on the unreliable scene extent that these large scenes expose. 
The GOF absolute-gradient signal (\emph{- GOF Grad}) is the largest quality contributor among the remaining components ($6\%$ better LPIPS at a moderately higher count). 
Long-axis split reduces the count by $31\%$ at a marginal $3\%$ LPIPS trade, contribution pruning by $9\%$ at unchanged quality, while the statistics warm-up and size prune are negligible. 
Reverting opacity decay to reset gives marginally better quality at $76\%$ more Gaussians (both variants in \cref{tab:exp:sota}), confirming decay as the more efficient default.
\begin{table}[t]
    \centering
    \caption{Ablation on the Zip-NeRF scenes: removing ($-$) or adding ($+$) one component relative to our full model. The catastrophic failure row (gray) is excluded from the color scale. Counts are in millions of Gaussians; time in minutes.}
    \label{tab:exp:ablation-zip}
    \begin{tabular}{l*{6}{c}}
        \toprule
        Components & PSNR$\uparrow$ & SSIM$\uparrow$ & LPIPS$\downarrow$ & GMSD$\downarrow$ & Count$\downarrow$ & Time$\downarrow$ \\
        \midrule
        Full model      & \hl{60}{25.81} & \hl{50}{0.830} & \hl{46}{0.330} & \hl{51}{0.106} & \hl{40}{1.133} & \hl{12}{29.9} \\
        + clone             & \hl{99}{20.86} & \hl{99}{0.735} & \hl{99}{0.432} & \hl{99}{0.175} & \hl{99}{0.760} & \hl{99}{28.6} \\
        - GOF Grad          & \hl{12}{25.56} & \hl{12}{0.818} & \hl{12}{0.351} & \hl{12}{0.119} & \hl{60}{0.781} & \hl{60}{23.8} \\
        - long-axis split   & \hl{58}{25.80} & \hl{60}{0.833} & \hl{60}{0.321} & \hl{60}{0.103} & \hl{12}{1.635} & \hl{14}{29.6} \\
        - contrib prune     & \hl{56}{25.79} & \hl{50}{0.830} & \hl{46}{0.330} & \hl{51}{0.106} & \hl{34}{1.245} & \hl{28}{27.9} \\
        + warm-up stats     & \hl{60}{25.81} & \hl{50}{0.830} & \hl{46}{0.330} & \hl{51}{0.106} & \hl{40}{1.134} & \hl{37}{26.7} \\
        + size prune        & \hl{58}{25.80} & \hl{50}{0.830} & \hl{46}{0.330} & \hl{51}{0.106} & \hl{40}{1.133} & \hl{38}{26.6} \\
        \bottomrule
    \end{tabular}
\end{table}

\section{Conclusion \& Limitations} \label{sec:conclusion} \label{sec:limitations}
We expose the occlusion-blindness of the adaptive density control in standard 3DGS and propose a modular solution, transferable to existing Gaussian Splatting methods, that ensures occlusion-aware densification statistics. 
We further present a novel densification criterion that normalizes a Gaussian's accumulated screen-space gradients by the accumulated average contributions of its rendered views. 
This avoids the view- and pixel-count biases of previous criteria under occlusion, large footprints, and varying distance. 
This leads to perceptually better reconstructions at fewer Gaussians on large, occlusion-heavy scenes. 
We additionally integrate established densification improvements into a framework that reduces the dependence on unreliable heuristics such as the scene extent.

Large-scale reconstruction nevertheless does not yet reach small-scene quality. 
Beyond occlusion, the imbalanced camera coverage of the Zip-NeRF~\cite{barron2023zipnerf} scenes limits what densification can recover (\cref{sec:exp:largescale}). 
Longer schedules restore the missing densification opportunity, but a fairer distribution of observations would address the imbalance at its source. 
Several works group cameras by scene sections to pursue divide-and-conquer strategies~\cite{hierarchicalgaussians24,liu2025occlugaussian}. 
Combining occlusion-aware densification with a balanced training strategy should enable higher-quality large-scale reconstructions via single joint optimizations. 
Our code will be made available at \url{https://github.com/MarcelRogge/occlusion-aware-3dgs}.

\subsubsection*{\ackname}
This work was partially funded by the European Union under the grant number 101298864, project Victor-XR.

\bibliographystyle{splncs04}
\bibliography{main}

\let\begin\arxivbegin
\let\documentclass\arxivdocumentclass

\usepackage{docmute}

\begin{document}


\makeatletter
\let\maketitle\relax
\let\bibliographystyle\@gobble
\let\bibliography\@gobble
\makeatother

\clearpage
\setcounter{table}{0}
\setcounter{figure}{0}
\setcounter{equation}{0}
\renewcommand{\theHtable}{appx.\arabic{table}}
\renewcommand{\theHfigure}{appx.\arabic{figure}}
\renewcommand{\theHequation}{appx.\arabic{equation}}
\renewcommand{\theHsection}{appx.\arabic{section}}
\renewcommand{\theHsubsection}{appx.\arabic{section}.\arabic{subsection}}
\renewcommand{\theHsubsubsection}{appx.\arabic{section}.\arabic{subsection}.\arabic{subsubsection}}

\begin{center}
  {\Large\bfseries Supplementary Material\par}
\end{center}
\vspace{0.5em}

\title{Counting Contributions, Not Views:\texorpdfstring{\\}{ }Occlusion-Aware Densification\texorpdfstring{\\}{ }for 3D Gaussian Splatting\texorpdfstring{\\}{ }--- Supplementary Material ---} 

\titlerunning{Counting Contributions, Not Views}

\author{Marcel Rogge\inst{1,2} \and
René Schuster\inst{1,2} \and
Didier Stricker\inst{1,2}}

\authorrunning{M.~Rogge et al.}

\institute{University Kaiserslautern-Landau -- RPTU \and
German Research Center for Artificial Intelligence -- DFKI
}

\maketitle

\renewcommand{\thetable}{S\arabic{table}}
\renewcommand{\thefigure}{S\arabic{figure}}
\renewcommand{\theequation}{S\arabic{equation}}

\appendix

This document supplements the main paper. 
References to the main paper's figures, tables, and equations are written out explicitly, while all other references resolve within this document. 
As in the main paper, all runs use the (Decay) opacity schedule unless stated otherwise, and evaluation follows the protocol of the main paper's experiments section. 

\paragraph{Overview.}
\Cref{sec:appendix:method} derives the densification gradient behind our average-contribution criterion and transfers both contributions to 2DGS~\cite{Huang2DGS2024} and 3DGS~\cite{kerbl3Dgaussians}. 
\Cref{sec:appendix:baseline} quantifies the integrated baseline against vanilla StopThePop. 
\Cref{sec:appendix:cloneanalysis} analyzes the clone failure mode behind our split-only decision, including the scene-extent override and an ablation study. 
\Cref{sec:appendix:occlusion} quantifies how occluded each benchmark is and shows that the per-scene occlusion rate predicts where our contributions help. 
\Cref{sec:appendix:pruning} details the contribution-prune schedule and takes a census of fully occluded Gaussians across methods. 
\Cref{sec:appendix:criterion} compares densification criteria on Mip-NeRF360, maps where each allocates Gaussians, and examines the denominator floor, robustness to the threshold and the evaluation resolution, a hybrid of PixelGS and our criterion, and two alternative normalizers. 
\Cref{sec:appendix:scalability} covers extended training schedules, count growth over iterations, the distance-scaling guard rail, and per-view statistics. 
\Cref{sec:appendix:qualitative} closes with full-frame qualitative results and error maps.

\section{Method Details} \label{sec:appendix:method}
In this section, we derive our average-contribution criterion from the per-pixel blending weights (\cref{sec:appendix:derivation}) and test how far our two contributions generalize beyond StopThePop~\cite{radl2024stopthepop} by porting them to 2DGS~\cite{Huang2DGS2024} and 3DGS~\cite{kerbl3Dgaussians} (\cref{sec:appendix:2dgs}).

\subsection{Criterion Derivation} \label{sec:appendix:derivation}
We recall the projected footprint of Gaussian $i$ at pixel $p$,
\begin{equation}
    G_{i,p} = \exp\!\big(-\tfrac{1}{2}\,(p-\mu_{i})^{T}\,{\Sigma_{i}'}^{-1}\,(p-\mu_{i})\big),
\end{equation}
with the projected screen-space mean $\mu_{i}$ and the projected 2D covariance $\Sigma_{i}'$. 
As in the main text, the densification gradient differentiates with respect to this projected mean, so all position derivatives below are two-dimensional.

The 2D screen-space gradient $g_{i,v}=(g_{i,v,x},g_{i,v,y})$ is used to determine how much a Gaussian $i$ wants to move based on the loss computed in view $v$. 
The same holds independently for every Gaussian. 
Kerbl \etal~\cite{kerbl3Dgaussians} argue that a Gaussian that tries to fill many places at once requires densification. 
Therefore, the magnitude $\|g_{i,v}\|$ is used for the criterion to determine densification. 
For a single fixed view $v$, we drop the view index and write $g_{i} \equiv g_{i,v}$. 
Additionally, we consider only the derivation of $g_{i,x}$, since $g_{i,y}$ follows analogously.
\begin{equation} \label{eq:gradient:main}
    g_{i,x} = \frac{\partial L}{\partial\mu_{i,x}} = \sum_p \frac{\partial L_{p}}{\partial\mu_{i,x}} = \sum_p \frac{\partial L_{p}}{\partial \alpha_{i,p}}\cdot \frac{\partial\alpha_{i,p}}{\partial \mu_{i,x}} = \sum_p \frac{\partial L_{p}}{\partial C_p} \cdot \frac{\partial C_p}{\partial \alpha_{i, p}} \cdot \frac{\partial \alpha_{i,p}}{\partial \mu_{i,x}},
\end{equation}
where $\sum_p$ runs over the $N_{i}$ pixels to which Gaussian $i$ contributes in view $v$. 
Using the chain rule, we express the gradient's per-pixel term as a product of three derivatives: 
$\partial L_p / \partial C_p$ represents the photometric error, $\partial C_p / \partial \alpha_{i,p}$ represents the blending sensitivity, and $\partial \alpha_{i,p}/\partial \mu_{i,x}$ represents the positional sensitivity.

\begin{equation} \label{eq:gradient:contrast}
    \frac{\partial C_p}{\partial \alpha_{i, p}} = (c_i\cdot T_{i,p} - A_{i,p}^{\text{gs}}\cdot T_{i,p} - c_{bg}\cdot T_{i,p}^{\text{tail}}\cdot T_{i,p}) = (c_i - A_{i,p}) \cdot T_{i,p},
\end{equation}
where $A_{i,p}^{\text{gs}}$ represents the color composited from all Gaussians behind Gaussian $i$, $c_{bg}$ is the background fill color, $A_{i,p}$ expresses the entire color composite behind Gaussian $i$, $T_{i,p}$ is the transmittance reaching Gaussian $i$, and $T_{i,p}^{\text{tail}}$ represents the transmittance behind Gaussian $i$. 
Note that $T_{i,p} \cdot (1-\alpha_{i,p}) \cdot T_{i,p}^{\text{tail}} = T_{final,p}$, where $T_{final,p}$ represents the leftover transmittance after blending all Gaussians overlapping pixel $p$. 
We can see that the blending sensitivity consists of the transmittance $T_{i,p}$ at Gaussian $i$ and a contrast term between the color $c_i$ of Gaussian $i$ and the composite color $A_{i,p}$ behind it.

\begin{equation} \label{eq:gradient:steering}
    \frac{\partial \alpha_{i,p}}{\partial \mu_{i,x}} = o_i \cdot \frac{\partial G_{i,p}}{\partial \mu_{i,x}} = o_i \cdot G_{i,p} \cdot \frac{\partial \ln(G_{i,p})}{\partial \mu_{i,x}} = \alpha_{i,p} \cdot \frac{\partial \ln(G_{i,p})}{\partial \mu_{i,x}},
\end{equation}
where $o_i$ is the peak opacity of Gaussian $i$ (at its center) and $\alpha_{i,p}=o_iG_{i,p}$ is its opacity at pixel $p$, attenuated by the Gaussian's footprint $G_{i,p}\in(0,1]$ on pixel $p$. 
The positional sensitivity term consists of a geometric steering term $\partial\ln(G_{i,p})/\partial \mu_{i,x}$ modulated by the Gaussian's opacity at pixel $p$.

Inserting \cref{eq:gradient:contrast} and \cref{eq:gradient:steering} into \cref{eq:gradient:main} yields:
\begin{equation} \label{eq:gradient:weight}
    \begin{split}
    g_{i,x} &= \sum_p \frac{\partial L_p}{\partial C_p} \cdot (c_i - A_{i,p}) \cdot T_{i,p} \cdot \alpha_{i,p} \cdot \frac{\partial \ln(G_{i,p})}{\partial \mu_{i,x}} \\
    &= \sum_p \frac{\partial L_p}{\partial C_p} \cdot (c_i - A_{i,p}) \cdot w_{i,p} \cdot \frac{\partial \ln(G_{i,p})}{\partial \mu_{i,x}},
    \end{split}
\end{equation}
where $w_{i,p}=\alpha_{i,p}T_{i,p}$ is the Gaussian's weight with which its color is blended into pixel $p$, \ie its contribution. 
\Cref{eq:gradient:weight} shows that $g_{i,x}$ consists of a photometric error, a geometric steering, and a contrast term, which are all scaled according to its contribution $w_{i,p}$. 
These three weight-independent factors form the per-pixel vector $s_{i,p}$ of Eq.~(3) in the main paper. 
This shows that the gradient is intrinsically contribution-weighted, which motivates our normalization using the mean contribution $\bar{w}_{i}=(1/N_{i})\sum_p w_{i,p}$ rather than a raw view count. 
The full criterion accumulates these per-view means over all views and adds the $+1$ floor (Eq.~(5) of the main paper), which bounds the score of a Gaussian that has accumulated little contribution (more details on the floor in \cref{sec:criterion:floor}). 
A contribution-based denominator is therefore the matched choice, though not every contribution-based denominator will do: 
the per-view \emph{total} $\sum_p w_{i,p}$ grows with the on-screen footprint and would reintroduce the very pixel-count dependence we argue against (Sec.~3.2 of the main paper), whereas the per-view mean $\bar{w}_{i}$ is footprint-free. 

\subsection{Generalizability (2DGS \& 3DGS)} \label{sec:appendix:2dgs}
Our contributions rely only on the per-pixel blending weights, so they should transfer to most other Gaussian Splatting renderers. 
To test this, we port both contributions to 2DGS~\cite{Huang2DGS2024}, whose surfel-based rasterizer and regularization differ substantially from StopThePop~\cite{radl2024stopthepop}. 
We additionally port them to 3DGS~\cite{kerbl3Dgaussians}, which shares StopThePop's splatting formulation but none of its per-pixel sorting. 
We do not include any of the other adopted components to isolate the generality of our proposed contributions. 
We report results on each vanilla base, the occlusion-aware filter alone (\emph{+ filter}), and the filter with criterion (\emph{+ filter + criterion}).

\paragraph{Mip-NeRF360.} 
We calibrate each port with a small threshold sweep on the subset of the Mip-NeRF360~\cite{barron2022mipnerf360} dataset used for our own calibration (\cref{sec:criterion:calibration}), reported in \cref{tab:appendix:ports-mip}.
Both ports behave the same way. 
At our original threshold of $0.002$, the criterion improves every quality metric over both the baseline and the filter-only variant, and on both ports it does so at a markedly increased Gaussian count ($1.8\times$ on 2DGS, $2.4\times$ on 3DGS). 
Raising the threshold to $0.003$ returns the count closer to the baseline's level on both ports while retaining much of the gain. 
Raising it further under-densifies, with every metric falling below the baseline on both 2DGS and 3DGS. 
The port's missing contribution prune does not explain the inflated count at our default threshold of $0.002$: 
disabling it in our StopThePop framework raises the Mip-NeRF360 count by only $4\%$ (\cref{tab:exp:occluded-census}). 
Later in this section, we will discuss a possible explanation in the clone operation.
\begin{table}[t]
    \centering
    \caption{Both ports on the Mip-NeRF360 subset of the threshold sweep (single runs, averaged over the four scenes): the vanilla base, the occlusion-aware filter alone, and the full port across the densification-threshold sweep. Shading compares within each base. Counts are in millions of Gaussians.}
    \label{tab:appendix:ports-mip}
    \setlength{\tabcolsep}{3pt}
    \begin{tabular}{cl*{4}{c}}
        \toprule
        & Method & PSNR$\uparrow$ & SSIM$\uparrow$ & LPIPS$\downarrow$ & Count$\downarrow$ \\
        \midrule
        \multirow{6}{*}{\rotatebox[origin=c]{90}{2DGS}} & 2DGS & \hl{15}{27.69} & \hl{24}{0.850} & \hl{20}{0.218} & \hl{40}{2.059} \\
         & + filter & \hl{38}{27.87} & \hl{41}{0.857} & \hl{35}{0.204} & \hl{32}{2.512} \\
         & + filter + criterion ($0.001$) & \multicolumn{4}{c}{over-densifies; out of memory} \\
         & + filter + criterion ($0.002$) & \hl{60}{28.04} & \hl{60}{0.865} & \hl{60}{0.182} & \hl{12}{3.682} \\
         & + filter + criterion ($0.003$) & \hl{41}{27.89} & \hl{36}{0.855} & \hl{35}{0.204} & \hl{38}{2.184} \\
         & + filter + criterion ($0.004$) & \hl{12}{27.67} & \hl{12}{0.845} & \hl{12}{0.225} & \hl{60}{0.910} \\
        \midrule
        \multirow{7}{*}{\rotatebox[origin=c]{90}{3DGS}} & 3DGS & \hl{34}{28.46} & \hl{42}{0.868} & \hl{37}{0.181} & \hl{46}{2.770} \\
         & + filter & \hl{36}{28.48} & \hl{47}{0.871} & \hl{42}{0.176} & \hl{43}{3.126} \\
         & + filter + criterion ($0.002$) & \hl{60}{28.71} & \hl{60}{0.877} & \hl{60}{0.158} & \hl{12}{6.519} \\
         & + filter + criterion ($0.003$) & \hl{54}{28.65} & \hl{50}{0.873} & \hl{47}{0.171} & \hl{41}{3.347} \\
         & + filter + criterion ($0.004$) & \hl{34}{28.46} & \hl{34}{0.865} & \hl{31}{0.187} & \hl{54}{1.990} \\
         & + filter + criterion ($0.005$) & \hl{12}{28.25} & \hl{12}{0.854} & \hl{12}{0.207} & \hl{60}{1.302} \\
        \bottomrule
    \end{tabular}
\end{table}

\paragraph{Zip-NeRF.} 
\Cref{tab:appendix:ports-zip} reports both ports on the occlusion-heavy Zip-NeRF~\cite{barron2023zipnerf} scenes.
For 2DGS we find that $0.003$ does not allow much more densification than the base, so we report the results for a threshold of $0.002$. 
For 3DGS we report with the recalibrated threshold of $0.003$. 
In both cases, we avoid the clone's catastrophic failure through a manual override of the scene extent (\cref{sec:appendix:extent}) so that the original densification strategy can be retained. 
Both baselines are starved here, at $0.57$M and $1.10$M Gaussians against the $1.8$--$2.8$M they use on the smaller Mip-NeRF360 scenes, which is the same under-densification our contributions correct on StopThePop. 
The filter alone lifts the count sharply on both, to $3.03$M and $2.90$M, and the criterion then spends that budget more selectively, settling at $1.48$M and $1.68$M. 
On 2DGS the full port improves every metric over the baseline on three of the four scenes ($+0.38$ to $+0.58$\,dB, $5$--$8\%$ better LPIPS). 
On 3DGS it attains the best PSNR on all four scenes, while conceding a little SSIM and LPIPS to the filter-only variant. 
For 2DGS there is an exception on alameda, where it degrades: 
the vanilla 2DGS already exhibits large floater artifacts that we attribute to this scene's strong exposure and lighting changes, which the surfel geometry consolidates into large floater blobs. 
Due to densification no longer getting suppressed, these floater clusters are amplified. 
Our criterion reduces the number of floaters and recovers most of the lost quality, while the 3DGS port does not show this degradation as it does not form the floaters into solid clusters.
\begin{table}[t]
    \centering
    \caption{Both ports on the Zip-NeRF scenes ($30$k, single runs), at each port's own densification threshold ($0.002$ for 2DGS, $0.003$ for 3DGS). Shading compares within each scene and base. Counts are in millions of Gaussians.}
    \label{tab:appendix:ports-zip}
    \setlength{\tabcolsep}{1pt}
    \small
    \begin{tabular}{cl*{8}{c}}
        \toprule
        & & \multicolumn{4}{c}{2DGS} & \multicolumn{4}{c}{3DGS} \\
        \cmidrule(lr){3-6}\cmidrule(lr){7-10}
        & Method & PSNR$\uparrow$ & SSIM$\uparrow$ & LPIPS$\downarrow$ & Count$\downarrow$ & PSNR$\uparrow$ & SSIM$\uparrow$ & LPIPS$\downarrow$ & Count$\downarrow$ \\
        \midrule
        \multirow{3}{*}{\rotatebox[origin=c]{90}{alameda}} & Vanilla & \hl{60}{20.90} & \hl{60}{0.689} & \hl{60}{0.456} & \hl{60}{0.630} & \hl{12}{22.28} & \hl{12}{0.728} & \hl{12}{0.400} & \hl{60}{1.216} \\
         & + filter & \hl{12}{18.95} & \hl{12}{0.661} & \hl{12}{0.480} & \hl{12}{3.222} & \hl{43}{22.40} & \hl{60}{0.740} & \hl{60}{0.380} & \hl{12}{2.654} \\
         & + filter + crit. & \hl{41}{20.13} & \hl{51}{0.684} & \hl{58}{0.457} & \hl{45}{1.443} & \hl{60}{22.46} & \hl{45}{0.736} & \hl{41}{0.388} & \hl{40}{1.806} \\
        \midrule
        \multirow{3}{*}{\rotatebox[origin=c]{90}{berlin}} & Vanilla & \hl{34}{25.79} & \hl{12}{0.865} & \hl{12}{0.359} & \hl{60}{0.556} & \hl{12}{27.10} & \hl{12}{0.886} & \hl{12}{0.326} & \hl{60}{1.024} \\
         & + filter & \hl{12}{25.48} & \hl{36}{0.870} & \hl{55}{0.343} & \hl{12}{3.039} & \hl{53}{27.39} & \hl{60}{0.895} & \hl{60}{0.308} & \hl{12}{3.142} \\
         & + filter + crit. & \hl{60}{26.17} & \hl{60}{0.875} & \hl{60}{0.341} & \hl{40}{1.587} & \hl{60}{27.44} & \hl{39}{0.891} & \hl{37}{0.316} & \hl{44}{1.710} \\
        \midrule
        \multirow{3}{*}{\rotatebox[origin=c]{90}{london}} & Vanilla & \hl{24}{24.33} & \hl{12}{0.768} & \hl{12}{0.428} & \hl{60}{0.414} & \hl{12}{25.52} & \hl{12}{0.804} & \hl{12}{0.380} & \hl{60}{0.835} \\
         & + filter & \hl{12}{24.20} & \hl{50}{0.783} & \hl{60}{0.396} & \hl{12}{2.253} & \hl{59}{25.83} & \hl{60}{0.816} & \hl{60}{0.358} & \hl{12}{1.870} \\
         & + filter + crit. & \hl{60}{24.73} & \hl{60}{0.787} & \hl{58}{0.397} & \hl{41}{1.136} & \hl{60}{25.83} & \hl{49}{0.813} & \hl{44}{0.365} & \hl{41}{1.253} \\
        \midrule
        \multirow{3}{*}{\rotatebox[origin=c]{90}{nyc}} & Vanilla & \hl{47}{25.70} & \hl{24}{0.815} & \hl{12}{0.368} & \hl{60}{0.674} & \hl{12}{26.76} & \hl{12}{0.840} & \hl{12}{0.328} & \hl{60}{1.335} \\
         & + filter & \hl{12}{24.11} & \hl{12}{0.809} & \hl{32}{0.356} & \hl{12}{3.612} & \hl{43}{26.97} & \hl{60}{0.851} & \hl{60}{0.304} & \hl{12}{3.948} \\
         & + filter + crit. & \hl{60}{26.28} & \hl{60}{0.833} & \hl{60}{0.339} & \hl{43}{1.734} & \hl{60}{27.08} & \hl{42}{0.847} & \hl{37}{0.315} & \hl{49}{1.937} \\
        \bottomrule
    \end{tabular}
\end{table}

\paragraph{The clone operation.} 
To investigate the inflated counts for our criterion on the Mip-NeRF360 scenes, we isolate the effect of the clone operation, which we already identify as being responsible for a catastrophic failure on large-scale scenes in \cref{sec:appendix:cloneanalysis}. 
We disable cloning by overriding the scene extent (\cref{sec:appendix:extent}) such that every densification becomes a split, and compare against an otherwise identical run with cloning enabled. 
Notably, the size prune is disabled in both cases since the scene-extent override would otherwise result in a different behavior between the clone and split-only runs. 
\Cref{tab:appendix:ports-clone} shows the results for both 2DGS and 3DGS. 
The result agrees across both ports: 
removing the clone discards between $18\%$ and $43\%$ of the final Gaussians while changing PSNR by at most $0.25$\,dB and LPIPS by at most $0.006$. 
On 2DGS at our default threshold of $0.002$, it cuts the count by $43\%$. 
This reduces the final count below the level of the baseline while producing better quality metrics, which mirrors the behavior of the criterion that we find on the StopThePop base. 
However, for 3DGS this alone does not restore the expected behavior on the default threshold of $0.002$. 
We see that cloning accounts for a significant amount of the final Gaussians without notable quality improvements. 
This is not the catastrophic failure it produces on the large scenes, but it shows the same mechanism: 
cloning causes over-densification if not suppressed.
\begin{table}[t]
    \centering
    \caption{Isolating the clone operation on both ports, on the Mip-NeRF360 subset (single runs, averaged over the four scenes). Shading compares within each base. Counts are in millions of Gaussians.}
    \label{tab:appendix:ports-clone}
    \setlength{\tabcolsep}{3pt}
    \begin{tabular}{cll*{4}{c}}
        \toprule
        & Threshold & Densification & PSNR$\uparrow$ & SSIM$\uparrow$ & LPIPS$\downarrow$ & Count$\downarrow$ \\
        \midrule
        \multirow{4}{*}{\rotatebox[origin=c]{90}{2DGS}} & $0.002$ & with clone & \hl{50}{28.14} & \hl{60}{0.867} & \hl{60}{0.181} & \hl{12}{2.947} \\
         &  & split-only & \hl{60}{28.19} & \hl{54}{0.865} & \hl{53}{0.185} & \hl{43}{1.696} \\
         & $0.003$ & with clone & \hl{23}{28.02} & \hl{26}{0.859} & \hl{21}{0.202} & \hl{49}{1.441} \\
         &  & split-only & \hl{12}{27.97} & \hl{12}{0.855} & \hl{12}{0.206} & \hl{60}{0.980} \\
        \midrule
        \multirow{4}{*}{\rotatebox[origin=c]{90}{3DGS}} & $0.002$ & with clone & \hl{60}{28.85} & \hl{60}{0.878} & \hl{60}{0.157} & \hl{12}{6.455} \\
         &  & split-only & \hl{12}{28.60} & \hl{19}{0.870} & \hl{42}{0.164} & \hl{26}{5.267} \\
         & $0.003$ & with clone & \hl{37}{28.73} & \hl{36}{0.874} & \hl{23}{0.170} & \hl{48}{3.304} \\
         &  & split-only & \hl{19}{28.64} & \hl{12}{0.869} & \hl{12}{0.174} & \hl{60}{2.263} \\
        \bottomrule
    \end{tabular}
\end{table}

\section{The Integrated Baseline} \label{sec:appendix:baseline}
We present results of our integrated baseline in \cref{tab:method:baseline}. 
Adopting only the components from related work described in Sec.~4.2 of the main paper, without our proposed contributions, already results in an improvement across all quality metrics. 
The Gaussian count increases in all cases, which could be interpreted as worse. 
However, we believe the baseline still under-densifies on Zip-NeRF: 
the count is lower than on the Mip-NeRF360 scenes even though the apartments are larger and more detailed. 
For Mip-NeRF360, it is more difficult to evaluate which count is appropriate. 
Even so, we consider the $14\%$ LPIPS improvement a meaningful gain that justifies the $24\%$ increase in count.
\begin{table}[t]
    \centering
    \caption{StopThePop~\cite{radl2024stopthepop} versus our integrated baseline (StopThePop with the adopted densification components, still occlusion-blind) on the Zip-NeRF (top) and Mip-NeRF360 (bottom) scenes (3-run averages). Counts are in millions of Gaussians; time in minutes.}
    \label{tab:method:baseline}
    \setlength{\tabcolsep}{3pt}
    \begin{tabular}{cl*{6}{c}}
        \toprule
        & Method & PSNR$\uparrow$ & SSIM$\uparrow$ & LPIPS$\downarrow$ & GMSD$\downarrow$ & Count$\downarrow$ & Time$\downarrow$ \\
        \midrule
        \multirow{2}{*}{\rotatebox[origin=c]{90}{Zip}} & StopThePop & \hl{12}{24.95} & \hl{12}{0.802} & \hl{12}{0.376} & \hl{12}{0.133} & \hl{60}{0.486} & \hl{12}{24.3} \\
         & Baseline & \hl{60}{25.52} & \hl{60}{0.818} & \hl{60}{0.352} & \hl{60}{0.117} & \hl{12}{0.557} & \hl{60}{22.3} \\
        \midrule
        \multirow{2}{*}{\rotatebox[origin=c]{90}{Mip}} & StopThePop & \hl{12}{27.01} & \hl{12}{0.809} & \hl{12}{0.227} & \hl{12}{0.102} & \hl{60}{1.711} & \hl{60}{25.6} \\
         & Baseline & \hl{60}{27.29} & \hl{60}{0.821} & \hl{60}{0.196} & \hl{60}{0.087} & \hl{12}{2.117} & \hl{12}{31.4} \\
        \bottomrule
    \end{tabular}
\end{table}

\section{Clone Analysis} \label{sec:appendix:cloneanalysis}
In this section, we give a small analysis of the default clone operation from 3DGS (Sec.~2.1 of the main paper). 
Specifically, we notice the creation of large floater clusters when the scene extent $E$ is too large relative to the detail in the scene. 
The significant impact of this is also visible in the main paper's ablation, where adding clone to our proposed method causes catastrophic failure (Sec.~4.7 of the main paper). 

\subsection{Manual Scene-Extent Override} \label{sec:appendix:extent}
The scene extent $E$ is automatically computed based on the camera positions and does not actually identify what size is appropriate for fine detail. 
We see this in the Zip-NeRF~\cite{barron2023zipnerf} apartments: 
cameras span multiple rooms (\eg alameda in \cref{fig:appendix:cameras}) while detail represents small areas within individual rooms. 
To isolate the effect of the clone operation based on the scene extent, we determine appropriate scene extent values for the Zip-NeRF scenes and set them manually for an ablation experiment. 
Specifically, we subdivide each scene from Zip-NeRF into smaller occlusion-light subsets such that we would expect an optimization on each subset to produce a high-quality reconstruction, as on the Mip-NeRF360 dataset. 
Per scene, we then compute the scene extent for each of these subsets and pick the smallest as the scene extent for the full scene. %
The original scene extents according to 3DGS' automatic estimation are as follows: 
$7.80$ for alameda, $6.64$ for berlin, $7.27$ for london, and $7.52$ for nyc. 
We override these with the following: 
$2.47$ for alameda, $2.81$ for berlin, $2.16$ for london, and $1.25$ for nyc. 
Accordingly, the extent-dependent ADC mechanisms (the split-versus-clone decision and the world-space size prune) operate on this manually corrected scale. 
\Cref{tab:appendix:clone-ablation} shows this manual override of the scene extent as \emph{+ extent}, which will be discussed in more detail in \cref{sec:appendix:clone}.

Grubert \etal~\cite{Grubert2025improving} propose an alternative scene extent, computed as the mean distance from the average camera position to the SfM points. 
This can result in a smaller scene extent than the camera-based one if the cameras are well distributed throughout the scene, and a larger one if the cameras are restricted to a small area of the scene. 
On the Zip-NeRF apartments, it evaluates to $3.8$--$4.9$, below the camera-based $6.6$--$7.8$ but still well above the detail-appropriate values we derived above. 
Ultimately, the goal of such an estimate is to match the perceived size of the scene better than the camera-based extent. 
However, any such extent only reasons about the size of the full scene and not what size of a Gaussian is appropriate for detail within that scene. 
That is, a fixed detail size (\eg a pen) is a larger fraction of a small scene's extent than of a large one; tying the Gaussian scale to scene extent therefore assumes a proportionality between detail and scene size that does not hold.

\subsection{Clone Ablation} \label{sec:appendix:clone}
\begin{table}[t]
    \centering
    \caption{Complete clone-fix ablation on the Zip-NeRF scenes (3-run averages): StopThePop plus our two contributions and/or one clone fix---split-only densification or the manual scene-extent override of \cref{sec:appendix:extent}. The catastrophic failure row (gray) is excluded from the color scale. Counts are in millions of Gaussians; time in minutes.}
    \label{tab:appendix:clone-ablation}
    \setlength{\tabcolsep}{3pt}
    \begin{tabular}{l*{6}{c}}
        \toprule
        Components & PSNR$\uparrow$ & SSIM$\uparrow$ & LPIPS$\downarrow$ & GMSD$\downarrow$ & Count$\downarrow$ & Time$\downarrow$ \\
        \midrule
        StopThePop & \hl{12}{24.95} & \hl{12}{0.802} & \hl{12}{0.376} & \hl{12}{0.133} & \hl{60}{0.486} & \hl{38}{24.3} \\
        + contributions & \hl{99}{23.03} & \hl{99}{0.776} & \hl{99}{0.393} & \hl{99}{0.148} & \hl{99}{1.099} & \hl{99}{26.6} \\
        \addlinespace
        + extent & \hl{18}{25.06} & \hl{24}{0.809} & \hl{23}{0.365} & \hl{23}{0.127} & \hl{55}{0.564} & \hl{58}{20.1} \\
        + contributions + extent & \hl{45}{25.55} & \hl{50}{0.824} & \hl{49}{0.341} & \hl{46}{0.114} & \hl{12}{1.303} & \hl{26}{27.0} \\
        \addlinespace
        + split-only & \hl{25}{25.19} & \hl{22}{0.808} & \hl{22}{0.366} & \hl{19}{0.129} & \hl{55}{0.570} & \hl{60}{19.7} \\
        + contributions + split-only & \hl{50}{25.63} & \hl{46}{0.822} & \hl{46}{0.343} & \hl{44}{0.115} & \hl{17}{1.212} & \hl{34}{25.2} \\
        \midrule
        Full model & \hl{60}{25.81} & \hl{60}{0.830} & \hl{60}{0.330} & \hl{60}{0.106} & \hl{22}{1.133} & \hl{12}{29.9} \\
        \bottomrule
    \end{tabular}
\end{table}
\Cref{tab:appendix:clone-ablation} shows a more detailed ablation of the clone operation. 
We evaluate two approaches to prevent the catastrophic failure that the clone operation causes on large scenes: 
a manual override of the scene \emph{extent} as discussed in \cref{sec:appendix:extent} and \emph{split-only} densification (as in the main ablation and our framework). 
Our scene extent override ablation does not change any components of the densification. 
Changing the extent only shifts the scale at which a Gaussian is split rather than cloned. 
Meanwhile, the split-only alternative modifies the densification by disabling cloning. 
However, it does not require determining an alternative scene extent value. 
Both split-only and the reduced scene extent on their own improve the StopThePop baseline by a small amount. 
The overall impact is minor, however, confirming that neither adjustment interferes with the default StopThePop. 
Adding our contributions on top of StopThePop enables densification in the occlusion-heavy Zip-NeRF scenes, where it was previously suppressed. 
The effect is that without any handling of the clone operation, large floater clusters appear and cause quality metrics to deteriorate significantly. 
This is also shown qualitatively in the first row of \cref{fig:appendix:clone}. 
Adding our contributions on top of StopThePop with either of the two clone corrections (scene extent override or split-only) prevents the catastrophic failure caused by the clone operation. 
In both cases, the no-longer-suppressed densification allocates more Gaussians to starved regions, which improves all quality metrics over the StopThePop baseline ($0.6$ to $0.7$\,dB PSNR and roughly $9\%$ LPIPS). 
\Cref{fig:appendix:clone} gives a qualitative comparison for the split-only variant in the second row, with zoomed crops in the third row. 
Our contributions clearly sharpen fine detail that split-only densification alone leaves blurred.

\begin{figure}[t]
    \centering
    \includegraphics[width=0.45\linewidth]{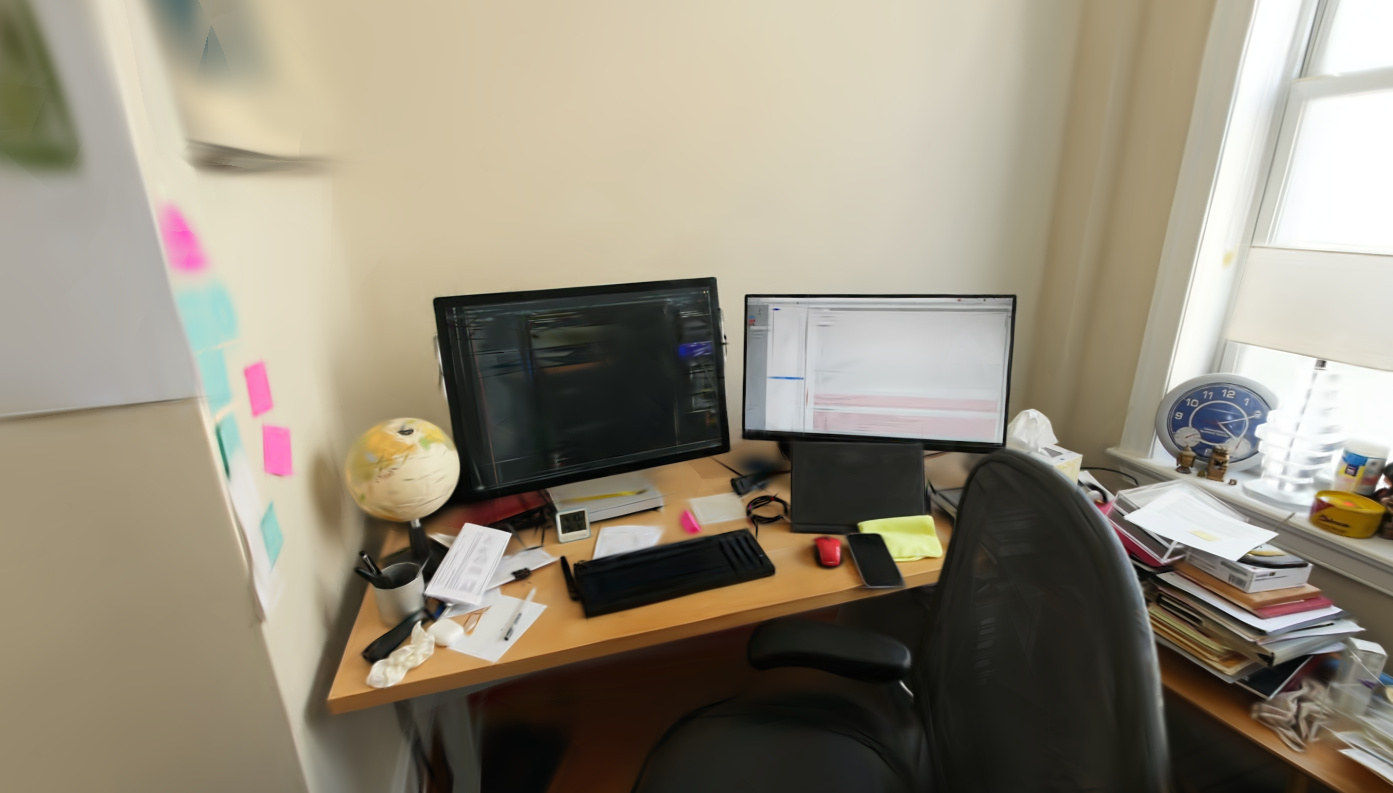}\hfill
    \includegraphics[width=0.45\linewidth]{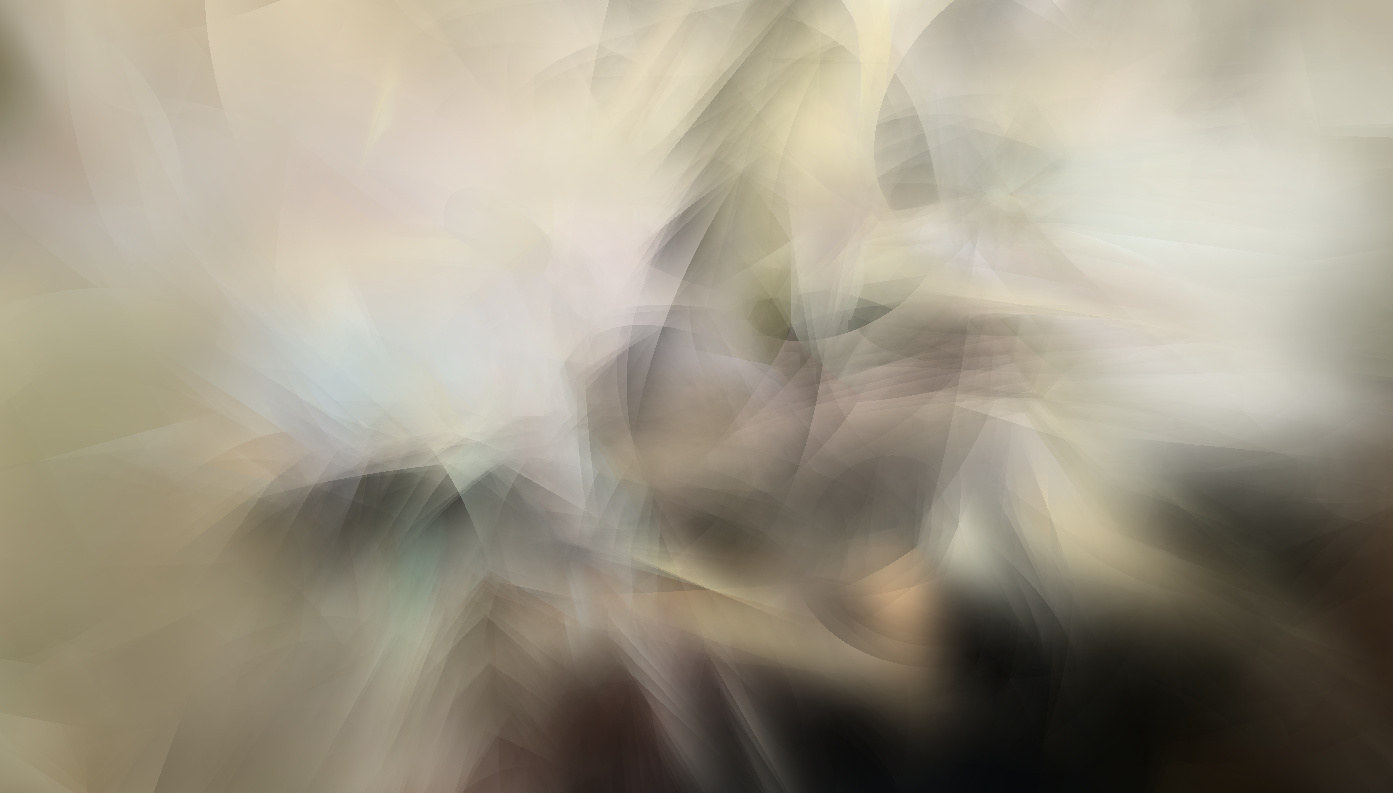}\\[1pt]
    \makebox[0.45\linewidth]{\scriptsize StopThePop}\hfill
    \makebox[0.45\linewidth]{\scriptsize \emph{+ contributions}}\\[3pt]
    \includegraphics[width=0.45\linewidth]{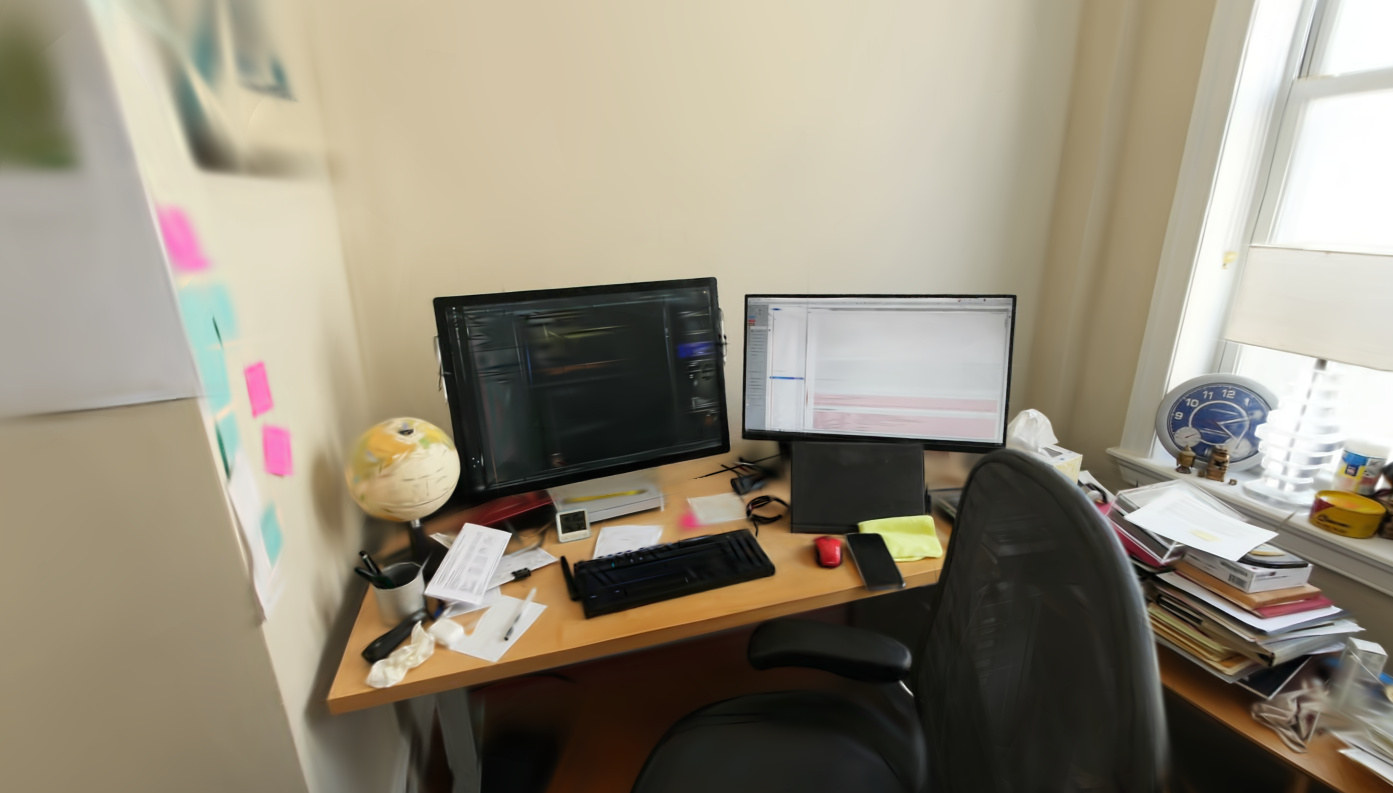}\hfill
    \includegraphics[width=0.45\linewidth]{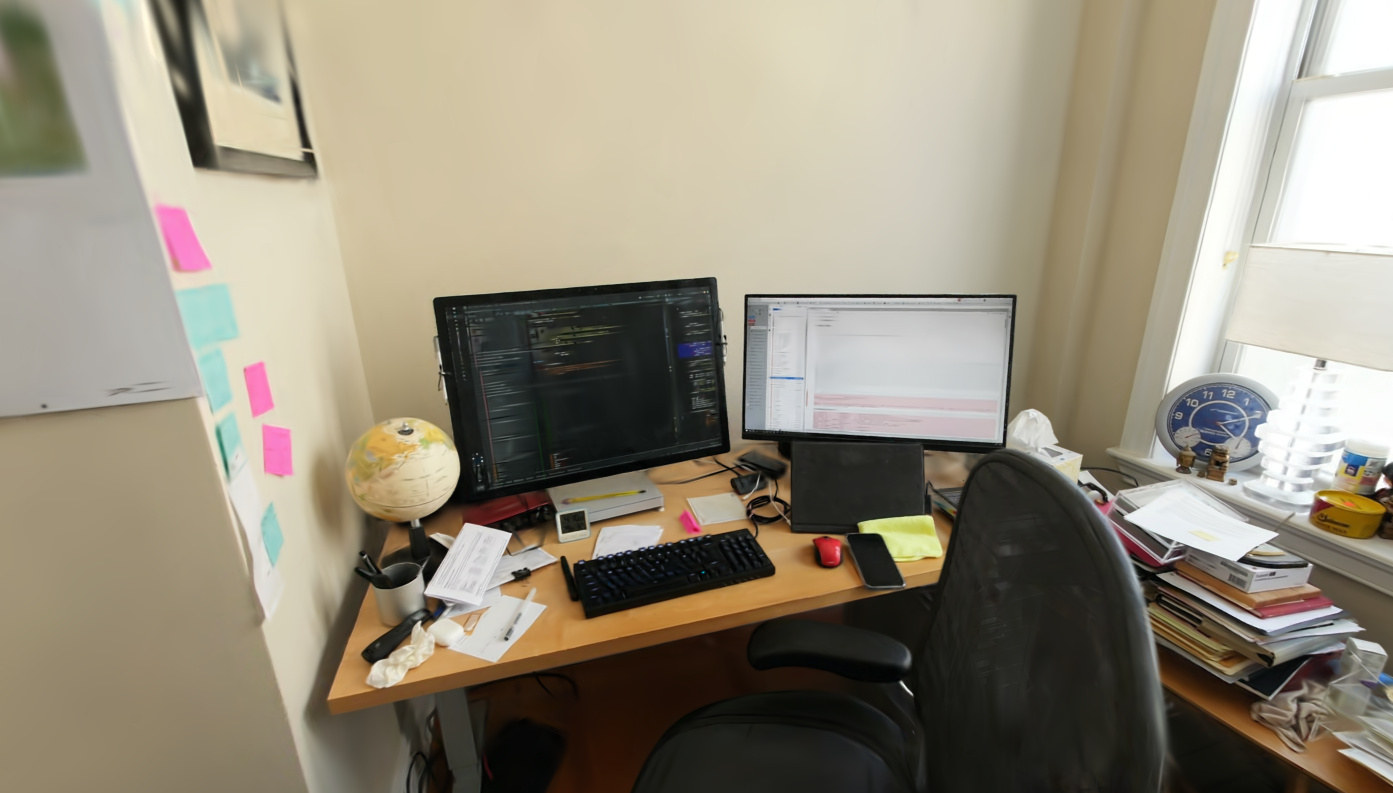}\\[1pt]
    \makebox[0.45\linewidth]{\scriptsize \emph{+ split-only}}\hfill
    \makebox[0.45\linewidth]{\scriptsize \emph{+ split-only + contributions}}\\[3pt]
    \setlength{\tabcolsep}{1.5pt}%
    \begin{tabular}{cccc}
        \multicolumn{2}{c}{\scriptsize Keyboard} & \multicolumn{2}{c}{\scriptsize Left monitor} \\
        \includegraphics[height=1.80cm,trim=557 180 666 511,clip]{figures/clonefail/nyc00007_so.jpg} &
        \includegraphics[height=1.80cm,trim=557 180 666 511,clip]{figures/clonefail/nyc00007_contrib_so.jpg} &
        \includegraphics[height=1.80cm,trim=441 325 684 294,clip]{figures/clonefail/nyc00007_so.jpg} &
        \includegraphics[height=1.80cm,trim=441 325 684 294,clip]{figures/clonefail/nyc00007_contrib_so.jpg} \\
        \scriptsize\shortstack{\emph{+ split-only}\\\strut} & \scriptsize\shortstack{\emph{+ split-only}\\ \emph{+ contributions}} & \scriptsize\shortstack{\emph{+ split-only}\\\strut} & \scriptsize\shortstack{\emph{+ split-only}\\ \emph{+ contributions}} \\
    \end{tabular}
    \caption{The clone failure mode on a nyc test view ($30$k schedule; one of the three ablation runs). Top row: StopThePop ($25.5$\,dB) reconstructs the view, then \emph{+ contributions} collapses it into scene-covering floater clusters ($12.3$\,dB). Bottom row: \emph{+ split-only} behaves like StopThePop ($25.7$\,dB), then \emph{+ contributions + split-only} resolves the view best ($28.5$\,dB). Insets (bottom): our contributions recover the individual keyboard keys and near-legible on-screen text that split-only densification alone leaves blurred.}
    \label{fig:appendix:clone}
\end{figure}

\section{Occlusion Rate of the Datasets} \label{sec:appendix:occlusion}
Our contributions build on occlusion-aware visibility, so their impact should depend on the rate of occlusions in a scene. 
We determine the rate of occlusions in the evaluated datasets in \cref{sec:appendix:occlusiondef} and correlate the impact of our proposed contributions with this occlusion rate in \cref{sec:appendix:occlusiongain}.

\subsection{Definition and Measurement of Occlusions} \label{sec:appendix:occlusiondef}
A Gaussian is occluded if it passes the frustum-based visibility filter while failing our occlusion-aware visibility filter. 
Let $f_{i,v}=1$ if Gaussian $i$ passes the frustum filter in view $v$ and $0$ otherwise, and let $r_{i,v}=1$ if $i$ also passes the occlusion-aware filter in $v$, that is, if it is rendered to at least one pixel of the output (otherwise $0$). 
We define the per-scene occlusion rate as
\begin{equation} \label{eq:appendix:occlusion}
    o = 1 - \frac{\sum_{v}\sum_{i} r_{i,v}}{\sum_{v}\sum_{i} f_{i,v}}.
\end{equation}
It is the share of frustum-visible Gaussian observations that do not actually contribute, computed across all training views of a scene. 
This is precisely the set of observations that our occlusion-aware filter excludes from the densification statistics. 

This approach of measuring occlusions requires a Gaussian model to render from. 
We considered two different types of models: 
initializations, which are method-independent, and fully optimized models ($30$k iterations), which have the most complete geometry but depend on the method used to produce them. 
We tried initializations with both the original initial opacity of $0.1$ and a raised opacity of $0.75$. 
Both initializations are unreliable and fail in opposite directions: 
$0.1$ returns an occlusion rate below $0.02$ for $13$ of the $17$ scenes. 
At $0.75$, four scenes (including Mip-NeRF360 scenes) exceed an occlusion rate of $0.90$. 
While a low opacity obviously reduces occlusions, we believe that \eg floaters in the sparse point cloud can exaggerate occlusions in the case of high opacity initialization. 
When evaluating occlusions on fully optimized models, there might be a bias depending on the method that was used to optimize the model. 
To avoid this bias, we compared occlusion rates between models produced by five different methods: 
StopThePop (Reset), StopThePop (Decay), PixelGS, Grubert \etal, and ours. 
We found that the occlusion rankings correlate strongly (Spearman $0.92$--$0.99$), which indicates that the occlusion rate is a property of the scenes rather than the methods used to optimize the model. 
Therefore, we choose to report occlusion rates measured on the fully optimized models, averaged over the five methods.

\Cref{tab:appendix:occlusion-dataset} reports the occlusion rates per dataset. 
The four Zip-NeRF apartments are the most occluded at $0.63$, followed by Tanks \& Temples at $0.48$, while Mip-NeRF360 and Deep Blending are far lower at $0.19$ and $0.15$. 
This ordering follows the geometry: 
the Zip-NeRF scenes are multi-room interiors in which most surfaces are hidden behind walls from any given camera, whereas the Mip-NeRF360 and Deep Blending scenes are largely centered on a single object or a single open space. 
Tanks \& Temples seems like an outlier: 
despite containing small scenes with a camera orbiting an object, it has a large occlusion rate. 
The reason is that the objects in this dataset (a truck and a train) are very large such that they occlude significant amounts of the scene from any given camera.
\begin{table}[t]
    \centering
    \caption{Occlusion rate (\cref{eq:appendix:occlusion}) per dataset, averaged over five trained models and all training views. The range gives the minimum and maximum over the dataset's scenes. Training views are averages.}
    \label{tab:appendix:occlusion-dataset}
    \begin{tabular}{lccc}
        \toprule
        Dataset & Occlusion & Scene range & Train views \\
        \midrule
        Zip-NeRF~\cite{barron2023zipnerf} & $0.630$ & $0.568$--$0.700$ & $1336$ \\
        Tanks \& Temples~\cite{Knapitsch2017tanks} & $0.482$ & $0.388$--$0.577$ & $241$ \\
        Mip-NeRF360~\cite{barron2022mipnerf360} & $0.194$ & $0.090$--$0.342$ & $189$ \\
        Deep Blending~\cite{DeepBlending2018} & $0.145$ & $0.083$--$0.208$ & $214$ \\
        \bottomrule
    \end{tabular}
\end{table}

\subsection{Impact of Contributions Correlates with Occlusion Rate} \label{sec:appendix:occlusiongain}
\Cref{tab:appendix:occlusion-scene} pairs each scene's occlusion rate with the LPIPS improvement our two contributions deliver over the occlusion-blind baseline. 
Both are measured inside the integrated framework with every other component held fixed, so the comparison isolates the contributions themselves. 
Across all $17$ scenes, the two quantities correlate at Spearman $+0.83$ (Pearson $+0.80$).
\begin{table}[t]
    \centering
    \caption{Per-scene occlusion rate against the LPIPS improvement of our two contributions over the occlusion-blind baseline. $\Delta$LPIPS is the improvement of ours. Sorted by occlusion rate.}
    \label{tab:appendix:occlusion-scene}
    \setlength{\tabcolsep}{4pt}
    \begin{tabular}{llccc}
        \toprule
        Dataset & Scene & Occlusion & LPIPS base & $\Delta$LPIPS \\
        \midrule
        Deep Blending & playroom  & $0.083$ & $0.2291$ & $-0.0037$ ($-1.6\%$) \\
        Mip-NeRF360   & room      & $0.089$ & $0.2062$ & $+0.0047$ ($+2.3\%$) \\
        Mip-NeRF360   & kitchen   & $0.144$ & $0.1253$ & $+0.0074$ ($+5.9\%$) \\
        Mip-NeRF360   & bicycle   & $0.146$ & $0.1797$ & $+0.0015$ ($+0.8\%$) \\
        Mip-NeRF360   & stump     & $0.178$ & $0.1910$ & $+0.0062$ ($+3.3\%$) \\
        Mip-NeRF360   & garden    & $0.179$ & $0.1029$ & $+0.0063$ ($+6.1\%$) \\
        Mip-NeRF360   & bonsai    & $0.201$ & $0.1974$ & $+0.0134$ ($+6.8\%$) \\
        Deep Blending & drjohnson & $0.208$ & $0.2355$ & $+0.0002$ ($+0.1\%$) \\
        Mip-NeRF360   & counter   & $0.225$ & $0.1972$ & $+0.0163$ ($+8.2\%$) \\
        Mip-NeRF360   & treehill  & $0.243$ & $0.2844$ & $+0.0109$ ($+3.8\%$) \\
        Mip-NeRF360   & flowers   & $0.342$ & $0.2801$ & $+0.0239$ ($+8.5\%$) \\
        Tanks \& Temples & truck  & $0.388$ & $0.1252$ & $+0.0186$ ($+14.9\%$) \\
        Zip-NeRF      & alameda   & $0.568$ & $0.4014$ & $+0.0253$ ($+6.3\%$) \\
        Tanks \& Temples & train  & $0.577$ & $0.1927$ & $+0.0250$ ($+13.0\%$) \\
        Zip-NeRF      & london    & $0.581$ & $0.3650$ & $+0.0271$ ($+7.4\%$) \\
        Zip-NeRF      & nyc       & $0.673$ & $0.3188$ & $+0.0186$ ($+5.8\%$) \\
        Zip-NeRF      & berlin    & $0.700$ & $0.3237$ & $+0.0177$ ($+5.5\%$) \\
        \bottomrule
    \end{tabular}
\end{table}

However, we note that the Zip-NeRF scenes were only optimized for $30$k iterations, which we find to be insufficient (more details on the training schedule in \cref{sec:appendix:scalability}). 
The quality metrics might, therefore, also be affected by insufficient reconstruction on those four scenes. 
Considering only the other $13$ scenes still results in a Spearman correlation of $+0.81$ (Pearson $+0.84$). 
The largest relative gains are on the two scenes from Tanks \& Temples, which are the most occluded besides the Zip-NeRF dataset.

\section{Occluded-Gaussian Pruning} \label{sec:appendix:pruning}
In this section, we detail our contribution prune in \cref{sec:appendix:prunesched}. 
We then evaluate how many fully occluded Gaussians remain in the final reconstructions of different methods in \cref{sec:appendix:census}.

\subsection{Contribution-Prune Implementation} \label{sec:appendix:prunesched}
We employ contribution pruning, as used by RadSplat~\cite{niemeyer2025radsplat} and ReLaGS~\cite{xiearafa2026relags}. 
However, our pruning runs alongside training, similarly to OC2DGS~\cite{RoggeOC2DGS2025} and EAGLES~\cite{girish2024eagles}. 
We improve the scheduling to be automatically synchronized with the training-view queue: 
a pruning decision is made only after every view has been seen once since the last decision. 
This prevents Gaussians in rarely observed regions from getting pruned due to a lack of opportunity. 
We further integrate the pruning with the existing opacity prune, which avoids additional pruning calls due to a misalignment of the training queue and densification intervals. 
This is done by recording a pruning mask when the pruning decision can be made, and postponing the actual pruning to the next densification event. 
After densification finishes, contribution pruning continues to remove redundant Gaussians on its own. 
Lastly, a post-training step iterates once through all training views for a final contribution prune, ensuring no low-contribution Gaussians remain from the final training iterations. 
Overall, our contribution prune contains only a single hyperparameter: 
the contribution threshold. 
We conservatively set it equal to the opacity prune's $0.005$ threshold. 
\Ie for fully visible foreground Gaussians, it does not prune anything that the opacity prune would not already prune. 
However, it extends this to partially and fully occluded Gaussians, while restricting pruning to those that had a chance to be seen.

\subsection{Occluded-Gaussian Prevalence} \label{sec:appendix:census}
\Cref{tab:exp:occluded-census} shows the share of occluded Gaussians relative to the total Gaussian count in the final reconstructions of state-of-the-art methods and ours. 
Occluded Gaussians are those that are fully occluded, \ie those with a maximum contribution of zero across all training views. 
Ours shows no occluded Gaussians in the final reconstructions, as expected given the design of our contribution prune. 
Meanwhile, other methods carry $10.5$--$15.7\%$ occluded Gaussians on the occlusion-heavy Zip-NeRF scenes and $6.9$--$9.2\%$ on the occlusion-light Mip-NeRF360 scenes. 
This represents a substantial amount of waste and highlights the importance of contribution-based pruning. 
The ablation study of the main paper (Sec.~4.7) also shows that this does not negatively affect the quality of the final model at our conservative threshold of $0.005$. 
We additionally evaluate ours without the contribution prune to isolate its impact on our approach. 
We note that even without the contribution prune, ours contains the smallest share of fully occluded Gaussians in the final reconstructions ($7.2\%$ on Zip-NeRF and $1.6\%$ on Mip-NeRF360). 
This indicates that our proposed average-contribution criterion makes better and more efficient densification decisions with less waste. 
We emphasize that this is not merely an artifact of our method being more compact than the others: 
on Zip-NeRF, the method by Grubert \etal produces final reconstructions with $9\%$ fewer total Gaussians but $75\%$ more occluded Gaussians. 
An example of the same can be seen on the Mip-NeRF360 results: 
StopThePop (Decay) contains $15\%$ fewer total Gaussians while carrying more than four times as many occluded Gaussians as ours without the contribution prune.
\begin{table}[t]
    \centering
    \caption{Fully occluded Gaussians (those that never contribute to any training view) in each method's reported model, on the Zip-NeRF (top) and Mip-NeRF360 (bottom) scenes (3-run averages). The occlusion-blind methods carry substantial dead weight, up to $23\%$ on the occlusion-heavy scenes. Our contribution prune removes them during training, so the final models contain none.}
    \label{tab:exp:occluded-census}
    \begin{tabular}{clccc}
        \toprule
        & Method & G(M) & Occluded & Share \\
        \midrule
        \multirow{7}{*}{\rotatebox[origin=c]{90}{Zip}} & StopThePop (Reset) & \hl{48}{0.951} & \hl{36}{149k} & \hl{24}{15.7\%} \\
        & StopThePop (Decay) & \hl{60}{0.486} & \hl{60}{50.9k} & \hl{48}{10.5\%} \\
        & PixelGS~\cite{zhang2024pixelgs} & \hl{12}{1.980} & \hl{12}{449k} & \hl{12}{22.7\%} \\
        & Grubert \etal~\cite{Grubert2025improving} & \hl{36}{1.131} & \hl{24}{156k} & \hl{36}{13.8\%} \\
        & Ours (no contribution prune) & \hl{24}{1.245} & \hl{48}{89.1k} & \hl{60}{7.2\%} \\
        \addlinespace
        & Ours (Reset)       & \hl{99}{1.991} & \hl{99}{0} & \hl{99}{0\%} \\
        & Ours (Decay)       & \hl{99}{1.133} & \hl{99}{0} & \hl{99}{0\%} \\
        \midrule
        \multirow{7}{*}{\rotatebox[origin=c]{90}{Mip}} & StopThePop (Reset) & \hl{24}{3.289} & \hl{24}{296k} & \hl{24}{9.0\%} \\
        & StopThePop (Decay) & \hl{60}{1.711} & \hl{48}{146k} & \hl{36}{8.5\%} \\
        & PixelGS            & \hl{12}{5.328} & \hl{12}{492k} & \hl{12}{9.2\%} \\
        & Grubert \etal      & \hl{36}{3.263} & \hl{36}{225k} & \hl{48}{6.9\%} \\
        & Ours (no contribution prune) & \hl{48}{2.023} & \hl{60}{33.3k} & \hl{60}{1.6\%} \\
        \addlinespace
        & Ours (Reset)       & \hl{99}{3.913} & \hl{99}{0} & \hl{99}{0\%} \\
        & Ours (Decay)       & \hl{99}{1.954} & \hl{99}{0} & \hl{99}{0\%} \\
        \bottomrule
    \end{tabular}
\end{table}

\section{Average Contribution Criterion} \label{sec:appendix:criterion}
In this section, we present a deeper analysis of our proposed average-contribution criterion: 
we calibrate its densification threshold (\cref{sec:criterion:calibration}), examine the denominator floor (\cref{sec:criterion:floor}), report the full Mip-NeRF360 results (\cref{sec:criterion:results}), analyze where the contributions redistribute Gaussians (\cref{sec:criterion:allocation}), examine robustness to the threshold (\cref{sec:exp:robustness}), isolate a hybrid of PixelGS and our criterion (\cref{sec:appendix:sumw}), and test alternative normalizers (\cref{sec:appendix:altnorm}).

\subsection{Threshold Calibration} \label{sec:criterion:calibration}
We calibrate our densification threshold once, with a single-run sweep of our full configuration on the calibration subset (\cref{tab:appendix:calibration}). 
The subset consists of four representative Mip-NeRF360 scenes: 
bicycle, bonsai, counter, and garden. 
The two outdoor scenes feature a prominent central object, one of them with thin structures, and the two indoor scenes contain a single object and multiple objects, respectively. 
The threshold $0.002$ sits at an interior optimum, attaining the best PSNR, SSIM, and GMSD of the sweep at $1.79$M Gaussians. 
Quality degrades on both sides, as over-densification ($0.001$, $4.90$M) lowers PSNR and under-densification ($0.004$, $0.66$M) raises LPIPS. 
We adopt this threshold and leave it unchanged on every other scene and dataset.
\begin{table}[t]
    \centering
    \caption{Threshold calibration of our criterion on the Mip-NeRF360 calibration subset (bicycle, bonsai, counter, garden), single runs, using our full integrated configuration: the occlusion-aware filter, the average-contribution criterion, and the adopted densification components. Our operating point $0.002$ (bold) attains the best PSNR, SSIM, and GMSD of the sweep; over- and under-densification degrade quality on either side. Counts are in millions of Gaussians.}
    \label{tab:appendix:calibration}
    \setlength{\tabcolsep}{5pt}
    \begin{tabular}{lccccc}
        \toprule
        Threshold & PSNR$\uparrow$ & SSIM$\uparrow$ & LPIPS$\downarrow$ & GMSD$\downarrow$ & Count \\
        \midrule
        $0.001$ & \hl{12}{27.98} & \hl{12}{0.864} & \hl{53}{0.162} & \hl{12}{0.077} & 4.90 \\
        $0.0015$ & \hl{51}{28.25} & \hl{56}{0.874} & \hl{60}{0.157} & \hl{60}{0.073} & 2.75 \\
        $\mathbf{0.002}$ & \hl{60}{28.31} & \hl{60}{0.875} & \hl{56}{0.160} & \hl{60}{0.073} & 1.79 \\
        $0.0025$ & \hl{50}{28.24} & \hl{56}{0.874} & \hl{47}{0.166} & \hl{48}{0.074} & 1.30 \\
        $0.004$ & \hl{38}{28.16} & \hl{16}{0.865} & \hl{12}{0.191} & \hl{24}{0.076} & 0.66 \\
        \bottomrule
    \end{tabular}
\end{table}

\subsection{The Denominator Floor} \label{sec:criterion:floor}
Our criterion normalizes the accumulated gradient by $1+\sum_v \bar{w}_{i,v}$, where the constant is a fixed floor (\cref{sec:appendix:derivation}). 
A vanishing accumulated contribution is not the failure case: a Gaussian that was never observed accumulates no gradient either, and is never densified. 
The failure case lies strictly in cases where a Gaussian has only few observations with low contributions, causing the accumulated contribution to fall between $0$ and $1$. 
There the accumulated contribution divides the gradient sum by less than one, inflating the score instead of normalizing it, and this inflation grows without bound as the sum approaches zero. 
So, densification stops responding to the signal the criterion is meant to measure. 
We sweep the floor over $0$, $0.5$, $2$ and $4$ on the Zip-NeRF scenes to establish two things: 
whether the floor is necessary, and what effect its specific value has.

\paragraph{Without the floor, training diverges.} 
We evaluate the behavior of the criterion without a floor in \cref{tab:appendix:floor-zero}. 
At floor $0$ and our default threshold, all four scenes cross the $2^{24}\!\approx\!16.8$M trainable limit and crash within roughly $1{,}800$ of the $30{,}000$ iterations. 
Raising the threshold delays the divergence but does not remove it: 
the criterion still diverges on all four scenes at five times the default threshold, and on three of four at ten times the default. 
Only at fifteen times the default ($0.03$) does training complete on every scene. 
However, the resulting model is far off the quality our criterion reaches at a comparable count: 
at $1.63$M Gaussians it loses $0.72$\,dB and $0.050$ LPIPS against floor $1$ at $1.77$M. 
The unbounded criterion's unstable performance is therefore not a matter of calibrating the right threshold. 
This indicates the necessity of the denominator floor.
\begin{table}[t]
    \centering
    \caption{Floor $0$ across a threshold ladder on the Zip-NeRF scenes ($30$k, single runs). Diverging rows report how many scenes crossed the trainable limit, together with the peak count reached and the iteration by which training stopped. The last two rows give floor $1$ for reference: our operating point, and the setting whose count is closest to the only floor-$0$ threshold that completes. Counts are in millions of Gaussians.}
    \label{tab:appendix:floor-zero}
    \setlength{\tabcolsep}{4pt}
    \begin{tabular}{llccccc}
        \toprule
        Floor & Threshold & PSNR$\uparrow$ & SSIM$\uparrow$ & LPIPS$\downarrow$ & GMSD$\downarrow$ & Count \\
        \midrule
        $0$ & $0.002$ & \multicolumn{5}{c}{\small diverges on $4/4$ scenes; peak $17.7$--$21.0$M by iteration ${\sim}1{,}800$} \\
        $0$ & $0.004$ & \multicolumn{5}{c}{\small diverges on $4/4$ scenes; peak $17.4$--$20.1$M by iteration ${\sim}2{,}300$} \\
        $0$ & $0.01$ & \multicolumn{5}{c}{\small diverges on $4/4$ scenes; peak $17.2$--$17.9$M by iteration ${\sim}5{,}000$} \\
        $0$ & $0.02$ & \multicolumn{5}{c}{\small diverges on $3/4$ scenes; berlin completes at $2.00$M} \\
        $0$ & $0.03$ & $25.15$ & $0.803$ & $0.369$ & $0.123$ & $1.63$ \\
        $0$ & $0.05$ & $24.36$ & $0.775$ & $0.417$ & $0.150$ & $0.21$ \\
        $0$ & $0.25$ & $22.94$ & $0.746$ & $0.455$ & $0.175$ & $0.04$ \\
        \midrule
        $1$ & $0.002$ & $25.80$ & $0.830$ & $0.330$ & $0.106$ & $1.13$ \\
        $1$ & $0.0015$ & $25.87$ & $0.835$ & $0.319$ & $0.102$ & $1.77$ \\
        \bottomrule
    \end{tabular}
\end{table}

\paragraph{Above zero, the floor rescales the threshold.} 
For the remaining floors, training is stable and the floor acts as a reparameterization of the threshold. 
At the default threshold (upper block of \cref{tab:appendix:floor}), the final Gaussian count increases for lower floors and decreases with higher floors. 
The quality moves with the count exactly as it does when the threshold itself is varied.

\paragraph{At a matched count, the floor value stops mattering.} 
To separate this count effect from a genuine quality effect, we re-run each floor with a predicted threshold that should move the final count closer to the floor $1$ baseline. %
For simplicity, we then compare these to the closest matching count from a floor $1$ threshold sweep (lower block of \cref{tab:appendix:floor}). 
For all three tested floors, we find a floor $1$ baseline within $3.8\%$--$21.5\%$ of the Gaussian count. 
Quality between these differs by only $0.01$--$0.08$\,dB PSNR and $0.001$--$0.007$ LPIPS. 
All three trace the same count--quality curve as floor $1$. %
Any floor large enough to keep the denominator from inflating the score is thus a reparameterization of the densification threshold rather than a quality parameter. 
We keep the floor at $1$ for all experiments, since it is the minimal value that guarantees that the denominator will not fall below $1$. 
We treat the densification threshold as the single tunable parameter of the criterion.
\begin{table}[t]
    \centering
    \caption{The denominator floor on the Zip-NeRF scenes ($30$k, single runs). Upper block: at the default threshold, the floor acts as a count--quality trade off. Lower block: with each floor's threshold scaled by a predicted value, interleaved with a threshold sweep of floor $1$ and ordered by count. Counts are in millions of Gaussians.}
    \label{tab:appendix:floor}
    \setlength{\tabcolsep}{5pt}
    \begin{tabular}{llccccc}
        \toprule
        Floor & Threshold & PSNR$\uparrow$ & SSIM$\uparrow$ & LPIPS$\downarrow$ & GMSD$\downarrow$ & Count \\
        \midrule
        $0.5$ & $0.002$ & $25.83$ & $0.836$ & $0.314$ & $0.100$ & $2.39$ \\
        $1$ & $0.002$ & $25.80$ & $0.830$ & $0.330$ & $0.106$ & $1.13$ \\
        $2$ & $0.002$ & $25.58$ & $0.818$ & $0.353$ & $0.117$ & $0.52$ \\
        $4$ & $0.002$ & $25.11$ & $0.801$ & $0.381$ & $0.132$ & $0.24$ \\
        \midrule
        $1$ & $0.004$ & $25.44$ & $0.812$ & $0.363$ & $0.121$ & $0.40$ \\
        $0.5$ & $0.004$ & $25.71$ & $0.824$ & $0.342$ & $0.111$ & $0.78$ \\
        $1$ & $0.0025$ & $25.74$ & $0.825$ & $0.340$ & $0.110$ & $0.81$ \\
        $1$ & $0.002$ & $25.80$ & $0.830$ & $0.330$ & $0.106$ & $1.13$ \\
        $2$ & $0.001$ & $25.88$ & $0.833$ & $0.323$ & $0.103$ & $1.44$ \\
        $4$ & $0.0005$ & $25.86$ & $0.834$ & $0.320$ & $0.103$ & $1.65$ \\
        $1$ & $0.0015$ & $25.87$ & $0.835$ & $0.319$ & $0.102$ & $1.77$ \\
        $1$ & $0.001$ & $25.82$ & $0.837$ & $0.309$ & $0.099$ & $3.29$ \\
        \bottomrule
    \end{tabular}
\end{table}

\subsection{Results on Mip-NeRF360} \label{sec:criterion:results}
In this section, we present the result tables for the Mip-NeRF360 scenes whose numbers are discussed in the main paper. 
\Cref{tab:appendix:criterion-mip} shows the Mip-NeRF360 results corresponding to the Zip-NeRF results reported in Tab.~1 of the main paper.
\begin{table}[t]
    \centering
    \caption{Densification criteria on Mip-NeRF360 (same criteria as Tab.~1 of the main paper; 3-run averages) and the occlusion-blind count of the integrated framework for reference. Counts are in millions of Gaussians; time in minutes.}
    \label{tab:appendix:criterion-mip}
    \setlength{\tabcolsep}{3pt}
    \begin{tabular}{l*{6}{c}}
        \toprule
        Method & PSNR$\uparrow$ & SSIM$\uparrow$ & LPIPS$\downarrow$ & GMSD$\downarrow$ & Count$\downarrow$ & Time$\downarrow$ \\
        \midrule
        Occlusion-blind count & \hl{45}{27.29} & \hl{44}{0.821} & \hl{12}{0.196} & \hl{12}{0.087} & \hl{55}{2.117} & \hl{60}{31.4} \\
        Occlusion-aware count & \hl{60}{27.33} & \hl{60}{0.822} & \hl{42}{0.188} & \hl{60}{0.086} & \hl{39}{2.595} & \hl{45}{35.4} \\
        PixelGS & \hl{19}{27.22} & \hl{12}{0.819} & \hl{56}{0.184} & \hl{12}{0.087} & \hl{12}{3.433} & \hl{12}{43.8} \\
        PixelGS + DS & \hl{12}{27.20} & \hl{28}{0.820} & \hl{60}{0.183} & \hl{12}{0.087} & \hl{21}{3.157} & \hl{18}{42.2} \\
        Ours & \hl{49}{27.30} & \hl{28}{0.820} & \hl{49}{0.186} & \hl{12}{0.087} & \hl{60}{1.954} & \hl{45}{35.4} \\
        \bottomrule
    \end{tabular}
\end{table}

Ours is the only criterion that ends below the occlusion-blind count it is built on, at $1.954$M Gaussians against $2.117$M, while every count-based criterion grows the model instead: 
$2.595$M for the occlusion-aware count, and $3.157$--$3.433$M for the two PixelGS variants. 
Despite being more compact, quality does not suffer for ours: 
against the occlusion-blind count, ours improves LPIPS by $5\%$ at a PSNR difference of $0.01$\,dB. 
Against the best LPIPS in the table it gives up $0.003$ while using $38\%$ fewer Gaussians. 
SSIM and GMSD barely move at all, spanning $0.819$--$0.822$ and $0.086$--$0.087$ across the table, which indicates that there are no noteworthy occlusion-induced errors in these lightly occluded scenes (\cref{sec:appendix:occlusion}).

We additionally present an evaluation of the difference between indoor and outdoor scenes of Mip-NeRF360 in \cref{tab:appendix:indoor-outdoor}, as is done by related work~\cite{radl2024stopthepop, Huang2DGS2024}. 
We note that ours performs generally better on the indoor scenes where most of the scene consists of well-defined geometry. 
On outdoor scenes where a lot of distant objects exist which are not observed well and contain a lot of foliage, ours drops a little in PSNR, SSIM, and GMSD but stays roughly on par with the occlusion-aware count in LPIPS. 
In terms of Gaussian count, ours is more compact than the occlusion-aware count on every split and trails only the vanilla StopThePop baseline overall, which is also the weakest in quality. 
We provide an additional evaluation of the decreased outdoor quality metrics in \cref{sec:criterion:allocation}, where we show that these pixel-aligned metrics are less reliable than the perceptual LPIPS on such foliage-heavy scenes.
\begin{table}[t]
    \centering
    \caption{Mip-NeRF360 performance split by scene type (3-run averages): the full 9-scene average, the 4 indoor scenes (bonsai, counter, kitchen, room), and the 5 outdoor scenes (bicycle, flowers, garden, stump, treehill). Counts are in millions of Gaussians.}
    \label{tab:appendix:indoor-outdoor}
    \setlength{\tabcolsep}{3pt}
    \begin{tabular}{cl*{5}{c}}
        \toprule
        & Method & PSNR$\uparrow$ & SSIM$\uparrow$ & LPIPS$\downarrow$ & GMSD$\downarrow$ & Count$\downarrow$ \\
        \midrule
        \multirow{4}{*}{\rotatebox[origin=c]{90}{Full}} & StopThePop & \hl{99}{27.01} & \hl{99}{0.809} & \hl{99}{0.227} & \hl{99}{0.102} & \hl{99}{1.711} \\
         & Occlusion-blind count & \hl{12}{27.29} & \hl{36}{0.821} & \hl{12}{0.196} & \hl{12}{0.087} & \hl{48}{2.117} \\
         & Occlusion-aware count & \hl{60}{27.33} & \hl{60}{0.822} & \hl{50}{0.188} & \hl{60}{0.086} & \hl{12}{2.595} \\
         & Ours & \hl{24}{27.30} & \hl{12}{0.820} & \hl{60}{0.186} & \hl{12}{0.087} & \hl{60}{1.954} \\
        \midrule
        \multirow{4}{*}{\rotatebox[origin=c]{90}{Indoor}} & StopThePop & \hl{99}{30.16} & \hl{99}{0.917} & \hl{99}{0.194} & \hl{99}{0.064} & \hl{99}{0.756} \\
         & Occlusion-blind count & \hl{12}{30.60} & \hl{12}{0.922} & \hl{12}{0.182} & \hl{12}{0.060} & \hl{60}{0.741} \\
         & Occlusion-aware count & \hl{33}{30.72} & \hl{31}{0.924} & \hl{34}{0.177} & \hl{36}{0.058} & \hl{12}{0.852} \\
         & Ours & \hl{60}{30.87} & \hl{60}{0.927} & \hl{60}{0.171} & \hl{60}{0.056} & \hl{29}{0.812} \\
        \midrule
        \multirow{4}{*}{\rotatebox[origin=c]{90}{Outdoor}} & StopThePop & \hl{99}{24.49} & \hl{99}{0.722} & \hl{99}{0.254} & \hl{99}{0.133} & \hl{99}{2.475} \\
         & Occlusion-blind count & \hl{60}{24.65} & \hl{60}{0.741} & \hl{12}{0.208} & \hl{60}{0.109} & \hl{45}{3.218} \\
         & Occlusion-aware count & \hl{51}{24.61} & \hl{53}{0.740} & \hl{60}{0.197} & \hl{60}{0.109} & \hl{12}{3.989} \\
         & Ours & \hl{12}{24.44} & \hl{12}{0.734} & \hl{56}{0.198} & \hl{12}{0.112} & \hl{60}{2.868} \\
        \bottomrule
    \end{tabular}
\end{table}

\subsection{Spatial Gaussian Allocation} \label{sec:criterion:allocation}
\begin{figure}[t]
    \centering
    \includegraphics[width=\linewidth]{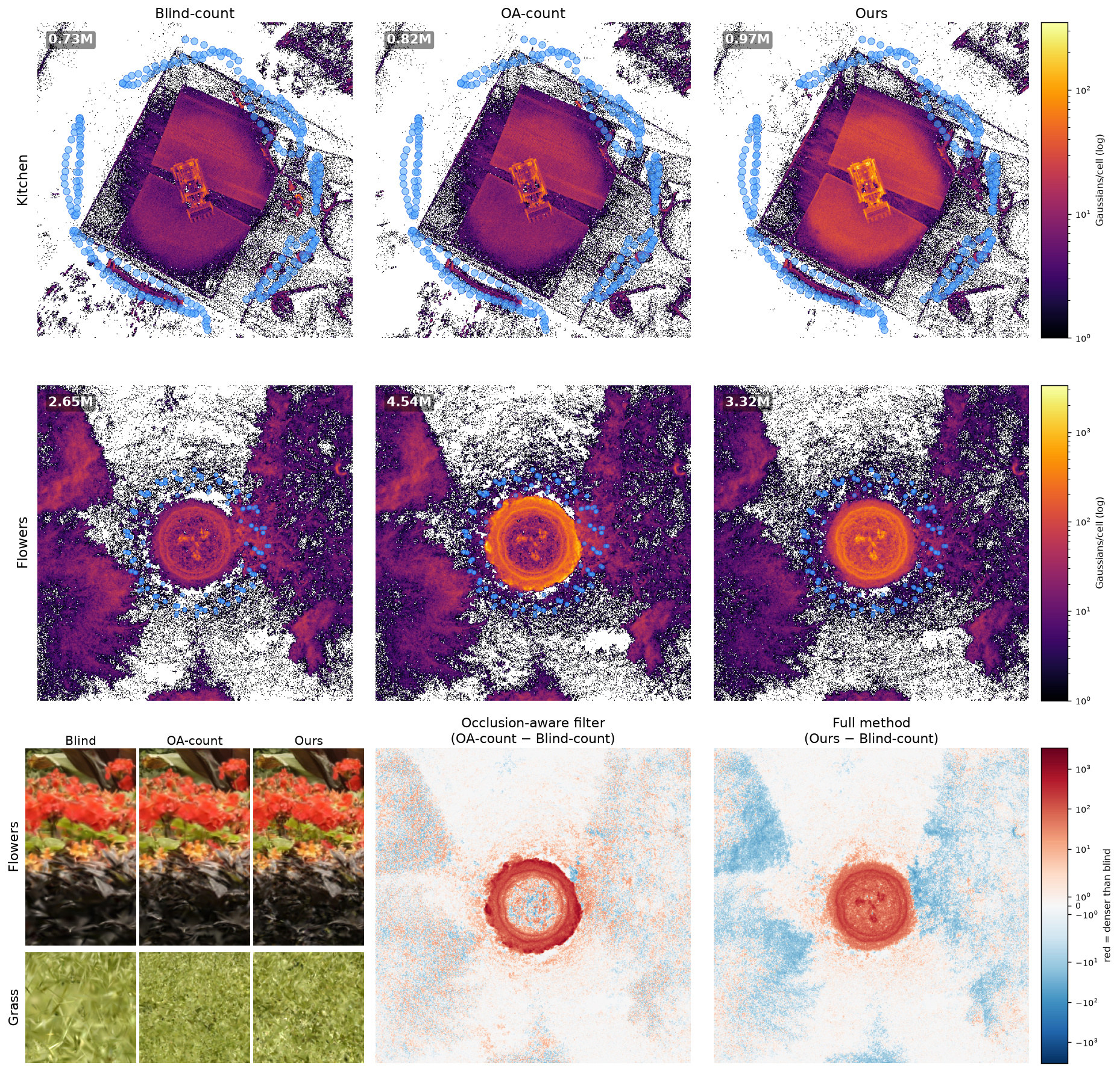}
    \caption{Top-down Gaussian density of two Mip-NeRF360 scenes (log Gaussians per ground cell; the color scale is shared across the three criteria within each scene), comparing the occlusion-blind count (Blind), the occlusion-aware count (OA-count), and our average-contribution criterion (Ours) on the same integrated baseline. Camera positions are blue dots, and each map is annotated with its total Gaussian count. \textbf{Kitchen (top):} Ours concentrates Gaussians on the well-observed central object. \textbf{Flowers (middle):} OA-count over-densifies the flat lawn around the flower bed (its bright disk balloons outward), whereas Ours keeps the bed dense and the lawn lean. \textbf{Bottom:} render crops of a flowers test view, the flower bed (top) and the surrounding grass (bottom), each as Blind $\vert$ OA-count $\vert$ Ours, followed by two difference maps relative to Blind (OA-count $-$ Blind and Ours $-$ Blind), red denser. The former isolates the occlusion-aware filter, which adds Gaussians across the bed and the lawn; the latter shows the full effect of Ours, which concentrates them on the observed bed. Blind renders visibly blurrier grass, while OA-count and Ours are similarly sharp (\cref{tab:appendix:flowers-region}).}
    \label{fig:appendix:allocation}
\end{figure}

We analyze \emph{where} the occlusion-aware filter and average-contribution criterion redistribute Gaussians. 
\Cref{fig:appendix:allocation} maps the top-down Gaussian density of two representative Mip-NeRF360 scenes, comparing the occlusion-blind count, the occlusion-aware count, and our average-contribution criterion on the same integrated baseline. 
We choose the Mip-NeRF360 dataset as it is one that all methods are able to reconstruct well enough to isolate the effect of our contributions absent other problems such as camera coverage (\cref{fig:appendix:cameras}) or exposure and lighting changes~\cite{niemeyer2025radsplat}.
On the well-observed indoor object (kitchen) the criterion concentrates Gaussians on the central region that the dense observations can reconstruct most reliably. 
Its higher count ($0.97$M against $0.73$M for the blind count) therefore places Gaussians where the observations are most informative. 
This is also reflected in its overall better quality metrics on the indoor scenes. 
On the foliage-heavy outdoor scene (flowers) the occlusion-aware count over-densifies the flat lawn around the bed, which our criterion keeps lean while allocating more Gaussians to the central flower bed. 
This spatial view is the counterpart of \cref{tab:appendix:indoor-outdoor}: 
on the indoor scene, ours allocates more Gaussians to the central object than the occlusion-blind count, which increases overall quality. 
Meanwhile, on the outdoor scene, ours needs roughly a quarter fewer Gaussians than the occlusion-aware count, which heavily over-densifies the grass around the flower bed.

\Cref{tab:appendix:flowers-region} makes the disproportionate allocation explicit. 
In the flower bed our criterion places the most Gaussians of the three ($2.3\times$ the occlusion-aware count and $3.7\times$ the occlusion-blind count) and reconstructs it best, with the highest SSIM and lowest GMSD. 
On the grass ring around the flower bed, the occlusion-aware count instead allocates ten times as many Gaussians as the occlusion-blind baseline. 
While it improves its GMSD slightly, both PSNR and SSIM deteriorate. 
The occlusion-blind count achieves the best PSNR for the grass ring although it is noticeably more blurry than the occlusion-aware count and ours (see the sample grass crop in \cref{fig:appendix:allocation}). 
This shows that PSNR and SSIM reward the smoothing that blur produces, so the blurry occlusion-blind lawn scores highest out of the three. 
Meanwhile, GMSD penalizes blur and ranks it worst. 
However, due to GMSD still being pixel-aligned, it marginally favors the occlusion-aware count's denser grass over ours.
\begin{table}[t]
    \centering
    \caption{Region-based allocation and image quality on the flowers scene (3-run averages). The two regions are the flower bed and the surrounding grass ring (radius $<1.9$ and $1.9$--$2.9$ from the centre of the flower bed in the top-down projection, respectively). For each region, Count is the number of Gaussians (in millions) positioned within it, and PSNR/SSIM/GMSD are measured on rough image crops of those regions across all test views.}
    \label{tab:appendix:flowers-region}
    \setlength{\tabcolsep}{3pt}
    \begin{tabular}{l*{8}{c}}
        \toprule
        & \multicolumn{4}{c}{Flower bed} & \multicolumn{4}{c}{Grass ring} \\
        \cmidrule(lr){2-5}\cmidrule(lr){6-9}
        Criterion & Count & PSNR$\uparrow$ & SSIM$\uparrow$ & GMSD$\downarrow$ & Count & PSNR$\uparrow$ & SSIM$\uparrow$ & GMSD$\downarrow$ \\
        \midrule
        Blind & 0.26 & \hl{12}{20.52} & \hl{12}{0.694} & \hl{12}{0.169} & 0.18 & \hl{60}{19.49} & \hl{60}{0.279} & \hl{12}{0.216} \\
        OA-count & 0.42 & \hl{60}{20.77} & \hl{47}{0.712} & \hl{41}{0.163} & 1.80 & \hl{54}{19.42} & \hl{41}{0.265} & \hl{60}{0.202} \\
        Ours & 0.97 & \hl{48}{20.71} & \hl{60}{0.719} & \hl{60}{0.159} & 0.62 & \hl{12}{18.95} & \hl{12}{0.243} & \hl{50}{0.205} \\
        \bottomrule
    \end{tabular}
\end{table}

\subsection{Robustness} \label{sec:exp:robustness}
We assess how sensitive our results are to two protocol choices: 
the densification threshold each criterion is given, and the resolution at which we evaluate.

\begin{figure}[t]
    \centering
    \includegraphics[width=\linewidth]{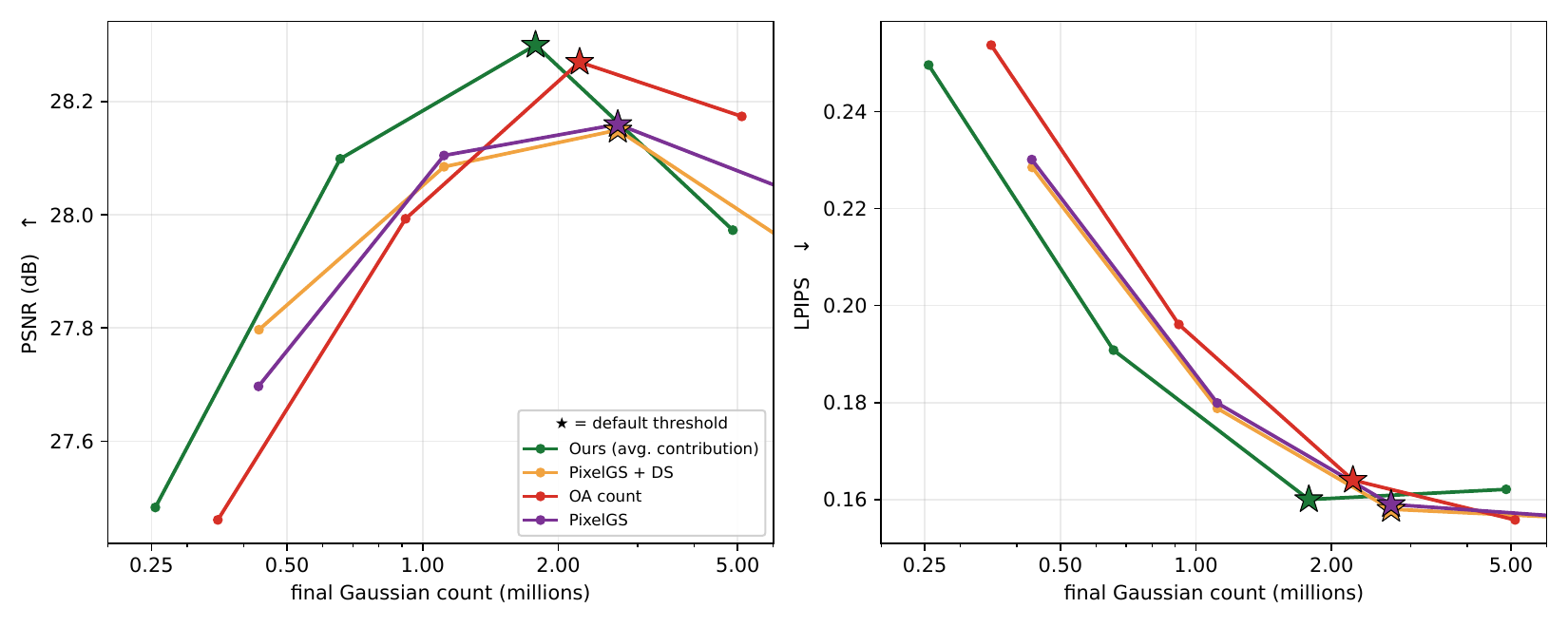}
    \caption{Robustness to the densification threshold on the Mip-NeRF360 subset: image quality versus final Gaussian count. Each data point represents a different threshold decreasing from left to right, as the count increases. All points are three-run averages.}
    \label{fig:robustness}
\end{figure}

\paragraph{Densification threshold sweep.} 
We sweep the densification threshold from values that starve the model to values that force heavy over-densification, for every criterion on a subset of the Mip-NeRF360~\cite{barron2022mipnerf360} dataset (garden, bicycle, bonsai and counter). 
We plot image quality against final count in \cref{fig:robustness}, sweeping each criterion over $\tfrac{1}{2}$, $1$, $2$ and $4$ times its own default threshold, so the threshold decreases from left to right as the count grows. 
Across the sweep every criterion follows the same arc: 
image quality rises with the Gaussian count up to a knee and then deteriorates as over-densification begins. 
On this subset, our average-contribution criterion reaches this knee at the \emph{lowest} count of any criterion: 
its default operating point already attains the best quality of the whole sweep at only $1.78$M Gaussians, so it offers the best quality per Gaussian. 
Ours even sits at an interior optimum: 
halving the threshold grows the count $2.7\times$ while degrading both PSNR and LPIPS, whereas the count-based criteria keep improving LPIPS only by adding Gaussians while their PSNR declines. 
At half the default threshold they already reach roughly three times our operating count ($5.1$--$6.3$M), the raw pixel-count criterion inflating the most while distance scaling barely changes it on this dataset. 
At a quarter of the default threshold every criterion over-densifies past the $2^{24}\!\approx\!16.8$M trainable limit and diverges on both outdoor scenes of the subset, which is why that setting is absent from the plot. 
So the meaningful comparison is at each criterion's operating point, where ours is simultaneously the most compact and the highest quality.

\paragraph{Native resolution.}
Following the common protocol, we evaluate the Zip-NeRF scenes at half resolution (Sec.~4.1 of the main paper). 
Since downsampling removes exactly the fine detail our contributions are meant to recover, we repeat the $30$k comparison against vanilla StopThePop~\cite{radl2024stopthepop} at the native, full resolution of the dataset (\cref{tab:appendix:nativeres}). 
The resolution does not change the outcome: 
ours improves every metric on all four scenes. 
The PSNR margin is in fact slightly larger than at half resolution ($+0.91$ against $+0.86$\,dB). 
The perceptual margins shrink but remain substantial, with LPIPS improving by $7.5\%$ instead of $12.2\%$ and GMSD by $16\%$ instead of $20\%$. 
Ours achieves this with $2.1\times$ the vanilla count, against $2.3\times$ at half resolution. 
The reported gains are therefore a property of the densification, not of the evaluation resolution.
\begin{table}[t]
    \centering
    \caption{Zip-NeRF trained with full-resolution images against the half-resolution protocol used throughout the paper ($30$k, single runs). Shading compares within each resolution block. Counts are in millions of Gaussians.}
    \label{tab:appendix:nativeres}
    \setlength{\tabcolsep}{4pt}
    \begin{tabular}{cl*{5}{c}}
        \toprule
        & Method & PSNR$\uparrow$ & SSIM$\uparrow$ & LPIPS$\downarrow$ & GMSD$\downarrow$ & Count$\downarrow$ \\
        \midrule
        \multirow{2}{*}{\rotatebox[origin=c]{90}{Full}} & StopThePop & \hl{12}{24.67} & \hl{12}{0.813} & \hl{12}{0.372} & \hl{12}{0.152} & \hl{60}{0.396} \\
         & Ours & \hl{60}{25.58} & \hl{60}{0.833} & \hl{60}{0.344} & \hl{60}{0.128} & \hl{12}{0.832} \\
        \midrule
        \multirow{2}{*}{\rotatebox[origin=c]{90}{Half}} & StopThePop & \hl{12}{24.95} & \hl{12}{0.802} & \hl{12}{0.376} & \hl{12}{0.133} & \hl{60}{0.486} \\
         & Ours & \hl{60}{25.81} & \hl{60}{0.830} & \hl{60}{0.330} & \hl{60}{0.106} & \hl{12}{1.133} \\
        \bottomrule
    \end{tabular}
\end{table}

\subsection{The PixelGS-ours Hybrid} \label{sec:appendix:sumw}
Our criterion and the PixelGS~\cite{zhang2024pixelgs} criterion differ in two aspects. 
The first is the per-view weight: our average contribution versus PixelGS's visible-pixel count. 
The second is how that weight enters the update: PixelGS forms a weighted mean of the per-view gradients, whereas we sum the gradients and normalize by the accumulated average contributions. 
To isolate the aggregation rule, we evaluate the PixelGS-ours hybrid: 
PixelGS's weighted-mean structure fed our contribution signal, using each view's total contribution $W_{i,v}=\sum_p w_{i,p}$ as the weight instead of the visible-pixel count, \ie $\gamma_i^{W} = (\sum_v W_{i,v}\, \|g_{i,v}\|)\,/ \sum_v W_{i,v}$. 
As this hybrid has no published default, we calibrate its threshold on the Mip-NeRF360 subset of the threshold sweep (\cref{sec:exp:robustness}), selecting $0.0003$. 
\Cref{tab:appendix:sumw} repeats the Zip-NeRF criterion comparison with this arm added (three runs). 
The transmittance weighting does not repair the weighted-mean aggregation: 
at its operating point the PixelGS-ours hybrid lands almost exactly on the raw PixelGS point ($6.4$M versus $6.6$M Gaussians at near-identical quality and training time), over-densifying heavily. 
The same contribution signal, entering instead through our normalizer, stays far more compact at $1.1$M. 
Under the extended $100$k schedule of the main paper, the PixelGS-ours hybrid diverges on three of the four scenes and reaches $7.4$M Gaussians on the fourth (single runs). 
This is the same long-schedule instability as the raw pixel-count criterion.
\begin{table}[t]
    \centering
    \caption{The PixelGS-ours hybrid on the occlusion-heavy Zip-NeRF scenes, added to the criterion comparison of the main paper (calibrated threshold $0.0003$, three runs), preceded by the occlusion-blind count of the integrated baseline; the extended-schedule row reports single runs. Counts are in millions of Gaussians; time in minutes.}
    \label{tab:appendix:sumw}
    \setlength{\tabcolsep}{3pt}
    \begin{tabular}{l*{6}{c}}
        \toprule
        Method & PSNR$\uparrow$ & SSIM$\uparrow$ & LPIPS$\downarrow$ & GMSD$\downarrow$ & Count$\downarrow$ & Time$\downarrow$ \\
        \midrule
        Occlusion-blind count & \hl{40}{25.52} & \hl{12}{0.818} & \hl{12}{0.352} & \hl{12}{0.117} & \hl{60}{0.557} & \hl{60}{22.3} \\
        Occlusion-aware count & \hl{56}{25.75} & \hl{60}{0.834} & \hl{52}{0.319} & \hl{60}{0.103} & \hl{44}{2.567} & \hl{45}{36.6} \\
        PixelGS & \hl{12}{25.11} & \hl{45}{0.829} & \hl{60}{0.312} & \hl{53}{0.105} & \hl{12}{6.607} & \hl{12}{67.0} \\
        PixelGS + DS & \hl{45}{25.59} & \hl{30}{0.824} & \hl{25}{0.341} & \hl{29}{0.112} & \hl{53}{1.443} & \hl{53}{28.8} \\
        PixelGS-ours hybrid & \hl{17}{25.18} & \hl{42}{0.828} & \hl{59}{0.313} & \hl{57}{0.104} & \hl{14}{6.366} & \hl{14}{65.2} \\
        Ours & \hl{60}{25.81} & \hl{48}{0.830} & \hl{38}{0.330} & \hl{50}{0.106} & \hl{55}{1.133} & \hl{52}{29.9} \\
        \midrule
        PixelGS-ours hybrid ($100$k) & \multicolumn{6}{c}{diverges on 3 of 4 scenes} \\
        \bottomrule
    \end{tabular}
\end{table}

\subsection{Alternative Normalizers} \label{sec:appendix:altnorm}
Our criterion is one of a family of possible contribution-based normalizers, so we test some plausible alternatives explicitly. 
The \emph{per-view} variant divides both sums by the number of rendered views $N_i$:
\begin{equation} \label{eq:appendix:perview}
    \gamma_{i}^{\text{per-view}}=\frac{\sum_v \|g_{i,v}\|/N_i}{1+\sum_v \bar{w}_{i,v}/N_i}.
\end{equation}
The \emph{sum-allocation} variant replaces the per-view mean by the per-view total $W_{i,v}=\sum_p w_{i,p}$: 
\begin{equation} \label{eq:appendix:sumalloc}
    \gamma_{i}^{\text{sum-alloc}}=\frac{\sum_v \|g_{i,v}\|}{1+\sum_v W_{i,v}}.
\end{equation}
Both place the denominator on a different scale than ours, so we run a threshold sweep for each such that it brings the final count into our range (\cref{tab:appendix:altnorm}).

Neither improves on our normalizer. 
The per-view variant is similar in quality to ours but requires $1.4\times$ as many Gaussians. 
Sum-allocation fails outright. 
A slight change in the threshold to push its count closer to ours causes it to diverge on three of the four scenes: 
each crossing the $2^{24}\!\approx\!16.8$M trainable limit, as the unfloored criterion shows at this schedule (\cref{sec:criterion:floor}). 
The pixel-count criterion from PixelGS reaches the same limit under the longer schedules of \cref{sec:appendix:countiter}. 
The single scene that completes overshoots our count by $69\%$ while remaining $2.77$\,dB behind. 
This is precisely the failure anticipated in \cref{sec:appendix:derivation}: 
the per-view total grows with the on-screen footprint, reintroducing the pixel-count dependence that the per-view mean removes.
\begin{table}[t]
    \centering
    \caption{Alternative normalizers on the Zip-NeRF scenes ($30$k, single runs), swept to bring each into our count range; our operating point is the $3$-run average of the main paper. Counts are in millions of Gaussians.}
    \label{tab:appendix:altnorm}
    \setlength{\tabcolsep}{4pt}
    \begin{tabular}{llcccc}
        \toprule
        Normalizer & Threshold & PSNR$\uparrow$ & LPIPS$\downarrow$ & GMSD$\downarrow$ & Count \\
        \midrule
        \multirow{3}{*}{per-view} & $0.002$ & $23.80$ & $0.431$ & $0.162$ & $0.06$ \\
         & $0.0005$ & $25.33$ & $0.366$ & $0.125$ & $0.45$ \\
         & $0.00025$ & $25.74$ & $0.329$ & $0.107$ & $1.60$ \\
        \midrule
        \multirow{3}{*}{sum-allocation} & $0.0003$ & $22.86$ & $0.455$ & $0.176$ & $0.12$ \\
         & $0.0002$ & $22.93$ & $0.449$ & $0.173$ & $0.89$ \\
         & $0.0001$ & \multicolumn{4}{c}{\small diverges on $3/4$ scenes; berlin completes at $1.79$M} \\
        \midrule
        Ours & $0.002$ & $25.81$ & $0.330$ & $0.106$ & $1.13$ \\
        \bottomrule
    \end{tabular}
\end{table}

\section{Scalability under a Matched Training Budget} \label{sec:appendix:scalability}
In this section, we analyze the effect of extended training schedules on the reconstruction performance. 
Specifically, the larger Zip-NeRF scenes contain an order of magnitude more images than the smaller Mip-NeRF360 scenes. 
This could reasonably result in an imbalance in the results between the small scenes and large scenes due to missing opportunity. 
Extending the training schedule isolates the opportunity and shows each method's performance ceiling.
We first map the uneven camera coverage of the large scenes (\cref{sec:appendix:cameras}), then detail the schedule scaling and report the $300$k results (\cref{sec:appendix:schedules}), trace the Gaussian count over iterations (\cref{sec:appendix:countiter}), examine the distance-scaling guard rail (\cref{sec:exp:distancescale}), and close with per-view statistics (\cref{sec:appendix:perview}).

\subsection{Camera Distribution} \label{sec:appendix:cameras}
\begin{figure}[t]
    \centering
    \includegraphics[width=0.92\linewidth]{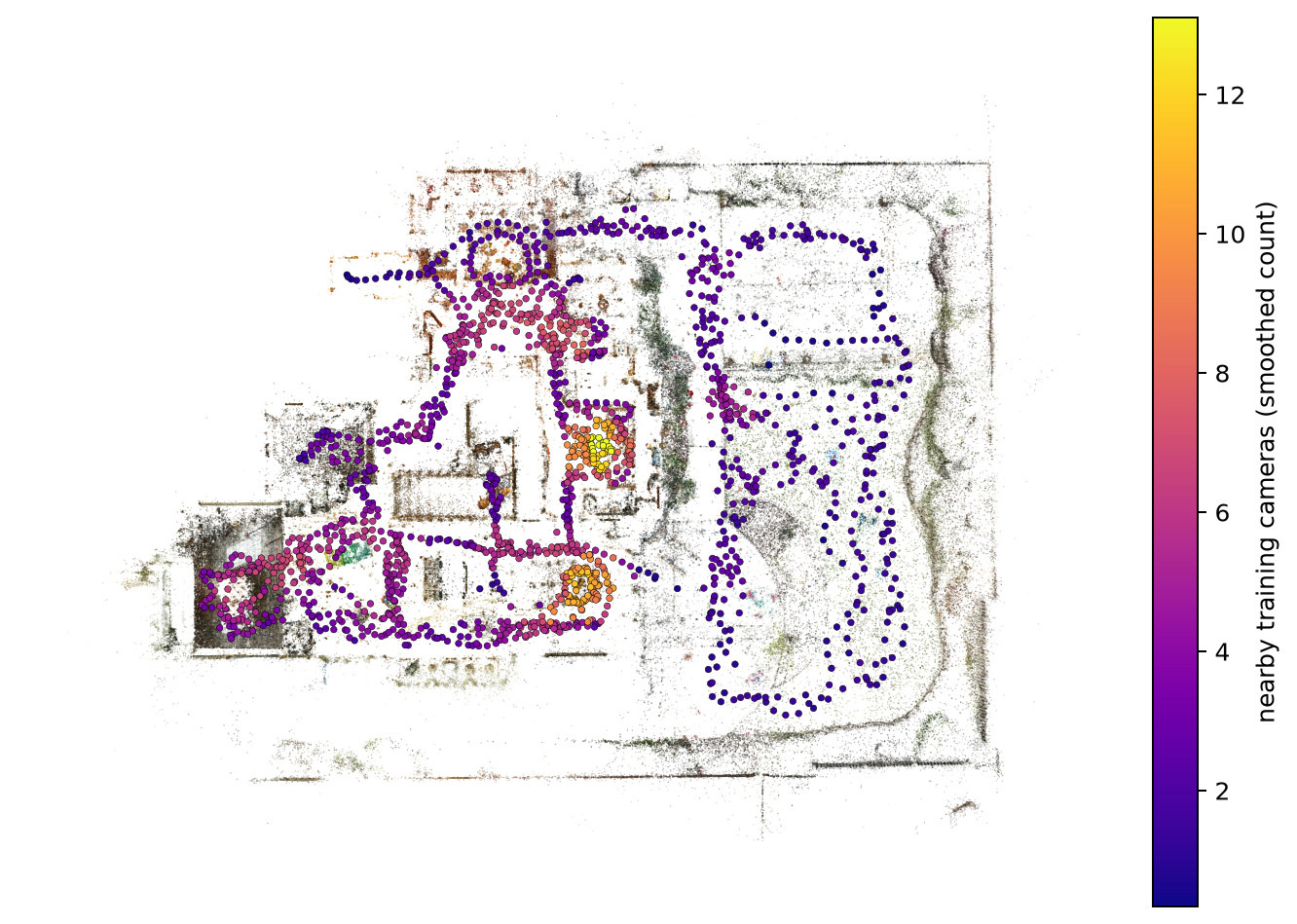}
    \caption{Training-camera distribution on the alameda scene (top-down). Dots mark the COLMAP camera positions, colored by the local density of training cameras; the sparse reconstruction is shown underneath for orientation. Interior hotspots are observed by an order of magnitude more cameras than the garden loop or the peripheral rooms.}
    \label{fig:appendix:cameras}
\end{figure}

\Cref{fig:appendix:cameras} maps the training-camera distribution of the alameda scene, extracted from its COLMAP reconstruction. 
Coverage is starkly imbalanced: 
some interior hotspots are covered by an order of magnitude more cameras than the garden or peripheral rooms. 
This is consistent with the per-view analysis (\cref{sec:appendix:perview}), whose weakest views lie in exactly such under-observed regions, and supports camera distribution as a remaining quality factor beyond occlusion.

\subsection{Extended Schedules and 300k Results} \label{sec:appendix:schedules}
For the matched $210$k schedule of the main paper and the $100$k and $300$k schedules sampled around it, we lengthen the training schedule and adjust its hyperparameters accordingly, and we report single runs owing to their cost. 
Relative to the $30$k base, we scale the schedule by a common factor $s = 10/3$, $7$, and $10$ for the $100$k, $210$k, and $300$k schedules. 
The iteration count becomes $s \times 30{,}000$. 
The positional learning-rate schedule, the densification window (iterations $500$ to $15{,}000$ at the $30$k base), the opacity-reset interval, and the opacity-decay interval all scale by the same factor of $s$. 
The densification interval is instead held fixed at $100$ iterations. 
This ensures that the densification opportunity per image also scales with $s$. 
\Cref{tab:appendix:300k} extends the main paper's extended-training comparison to a $300$k schedule: 
ours stays bounded (only $10\%$ more Gaussians than at $210$k) with LPIPS still improving marginally, while the guarded PixelGS~\cite{zhang2024pixelgs} also remains bounded but at $62\%$ more Gaussians and $40\%$ more training time than ours.

\begin{table}[t]
    \centering
    \caption{Extended $300$k schedule on the Zip-NeRF scenes (single runs), continuing the main paper's extended-training comparison. Counts are in millions of Gaussians; time in minutes.}
    \label{tab:appendix:300k}
    \setlength{\tabcolsep}{5pt}
    \begin{tabular}{l*{6}{c}}
        \toprule
        Method & PSNR$\uparrow$ & SSIM$\uparrow$ & LPIPS$\downarrow$ & GMSD$\downarrow$ & Count$\downarrow$ & Time$\downarrow$ \\
        \midrule
        PixelGS + DS & \hl{12}{26.46} & \hl{12}{0.848} & \hl{12}{0.285} & \hl{12}{0.090} & \hl{12}{4.472} & \hl{12}{553.5} \\
        Ours & \hl{60}{26.59} & \hl{60}{0.852} & \hl{60}{0.278} & \hl{60}{0.086} & \hl{60}{2.754} & \hl{60}{395.7} \\
        \bottomrule
    \end{tabular}
\end{table}

\subsection{Count vs.\ Iteration} \label{sec:appendix:countiter}
\begin{figure}[t]
    \centering
    \includegraphics[width=\linewidth]{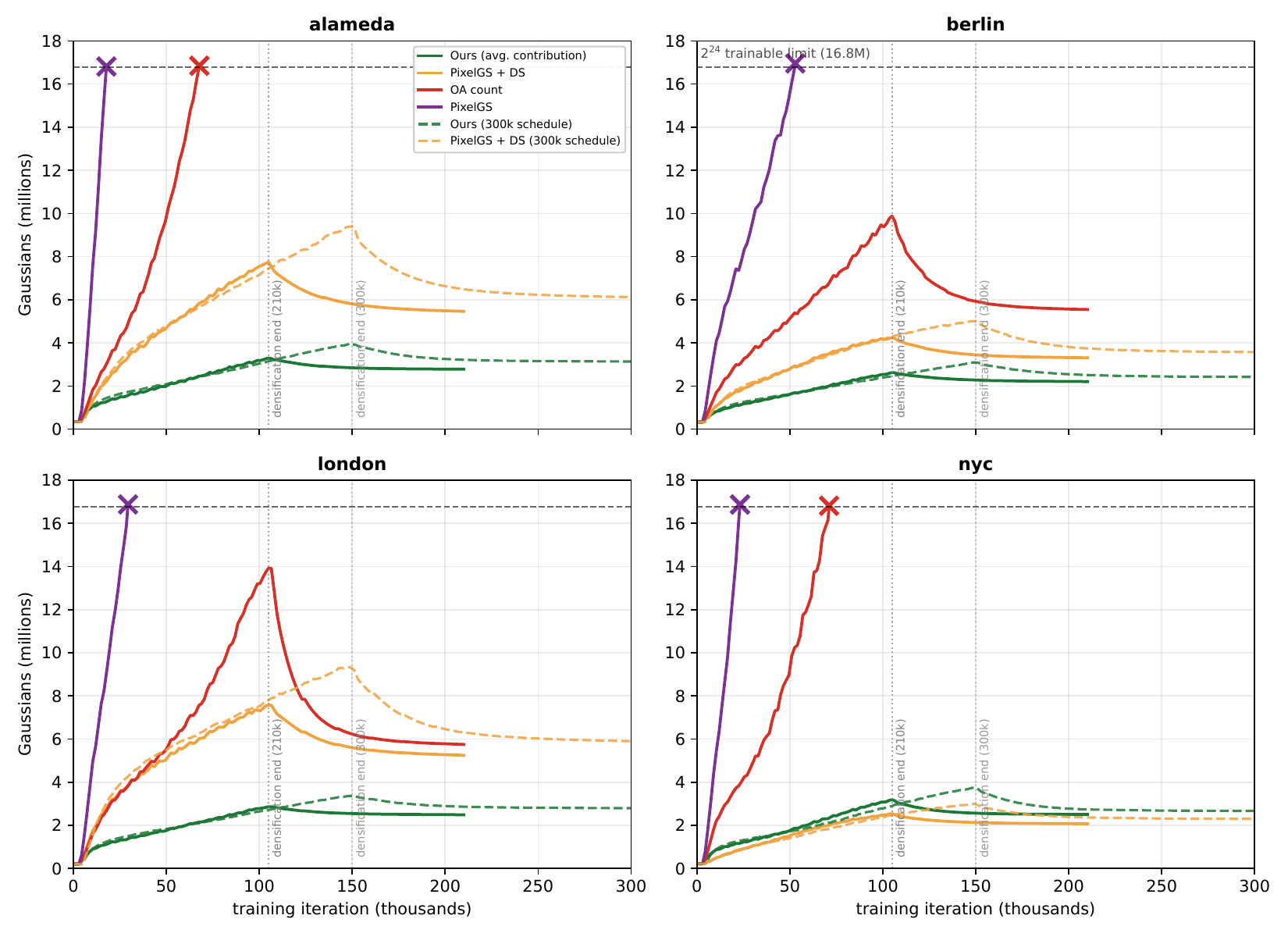}
    \caption{Gaussian count versus iteration on the four Zip-NeRF scenes (single run per scene, $210$k solid, $300$k dashed, dotted verticals mark each schedule's densification end, crosses mark crashes at the trainable limit). Ours (green) and PixelGS+DS (orange) stay bounded. The PixelGS criterion (purple) itself diverges on every scene, while the occlusion-aware count (red) diverges on two scenes.}
    \label{fig:exp:largescale-count}
\end{figure}

We evaluate the trajectory of the Gaussian counts over the training duration. 
\Cref{fig:exp:largescale-count} traces the count per iteration for all four criteria on the four Zip-NeRF scenes, at the $210$k schedule and, for the two criteria that complete it, also at $300$k. 
The occlusion-aware filter and all other components of the integrated framework are held fixed, isolating the criteria's influence alone.

Ours is the only criterion that stays bounded without a guard rail. 
It grows from $1.3$--$1.6$M Gaussians at iteration $30$k to a peak of $2.6$--$3.3$M at the end of the densification window. 
The count then declines monotonically to $2.2$--$2.8$M, with it flattening well before the training ends. 
This decline is the contribution prune, which keeps removing occluded primitives once densification no longer replaces them (\cref{sec:appendix:prunesched}).

The two count-based criteria diverge instead. 
The PixelGS criterion without the distance scaling crosses the $2^{24}\!\approx\!16.8$M trainable limit (\cref{sec:criterion:floor}) on all four scenes and completes none of them. 
The occlusion-aware count peaks later than the PixelGS criterion but does not avoid the divergence: 
it crashes on two of the four scenes. 
Notably, on the scenes where it does not crash it peaks at counts that are $1.8$--$2.5\times$ the final count. 
Since all criteria are using the same framework in these experiments, the occlusion-aware count also benefits from the contribution prune. 
The large amount of pruning indicates that a significant amount of Gaussians that are created are redundant and not beneficial to the overall geometry or appearance of the reconstruction. 
Conversely, this also indicates that ours is more efficient in choosing Gaussians to densify as the amount of pruning post densification is comparatively small. 
Despite heavy pruning, the occlusion-aware count still ends at $2.3$--$2.5\times$ our count.

The PixelGS~\cite{zhang2024pixelgs} criterion with distance scaling (PixelGS+DS) is bounded like ours, though by an additional guard rail than by the criterion itself. 
It still ends at $1.6\times$ our count on average and pays for it by suppressing the near-field densification whose absence shows up as blur (\cref{sec:exp:distancescale}).

\subsection{The Distance-Scaling Guard Rail} \label{sec:exp:distancescale}
\begin{figure}[t]
    \centering
    \setlength{\tabcolsep}{1pt}
    \begin{tabular}{cccc}
        \scriptsize Test view (GT) & \scriptsize GT crop & \scriptsize PixelGS + DS & \scriptsize Ours \\
        \includegraphics[width=0.36\linewidth]{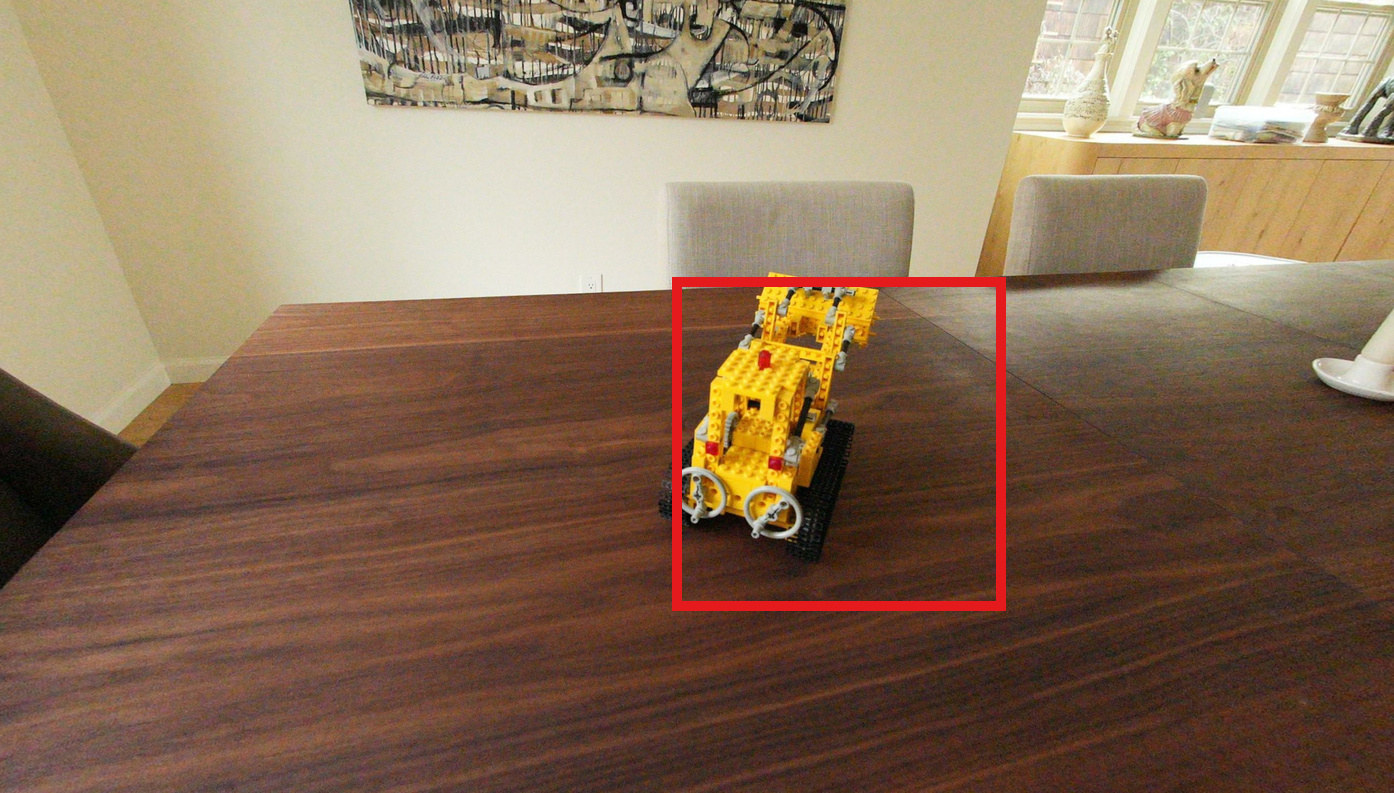} &
        \includegraphics[width=0.205\linewidth,trim=672 183 389 277,clip]{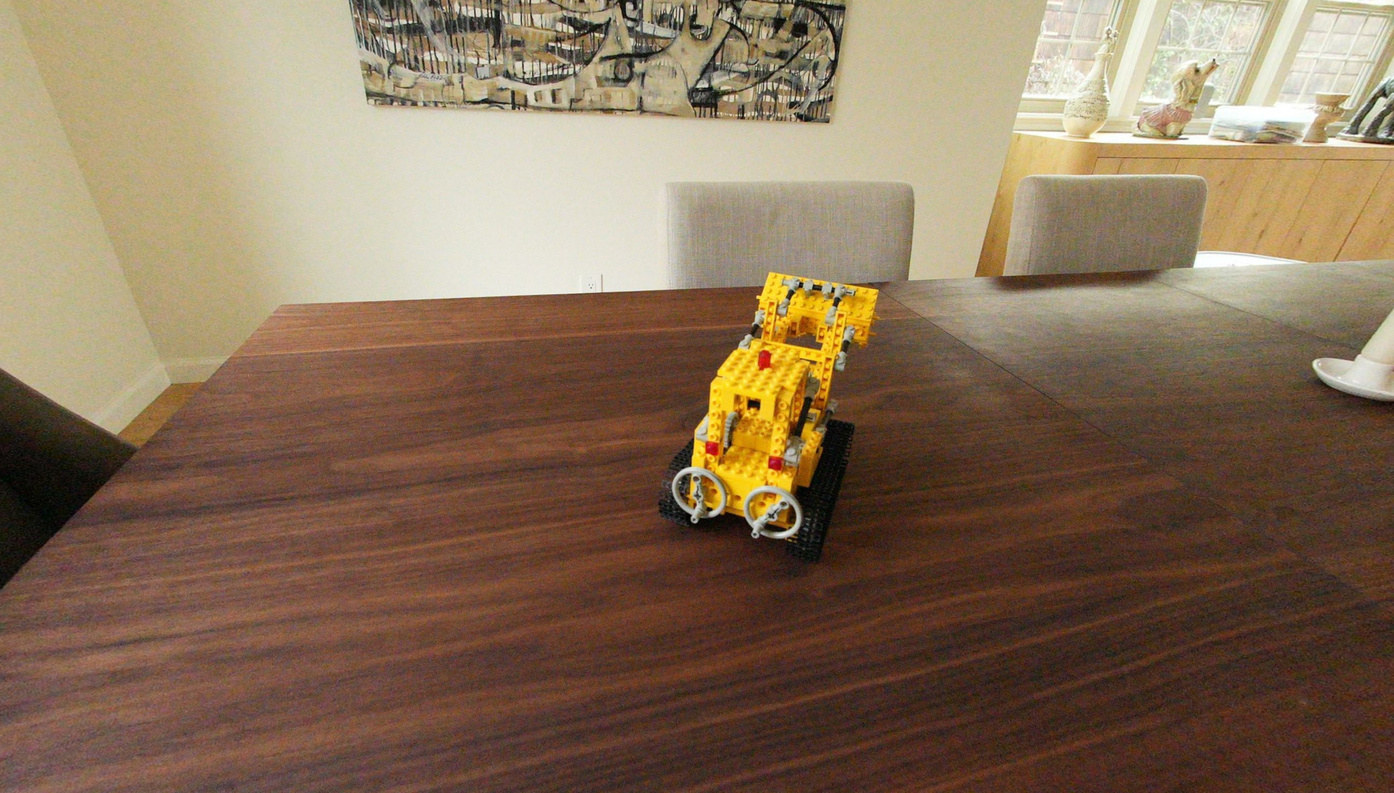} &
        \includegraphics[width=0.205\linewidth,trim=672 183 389 277,clip]{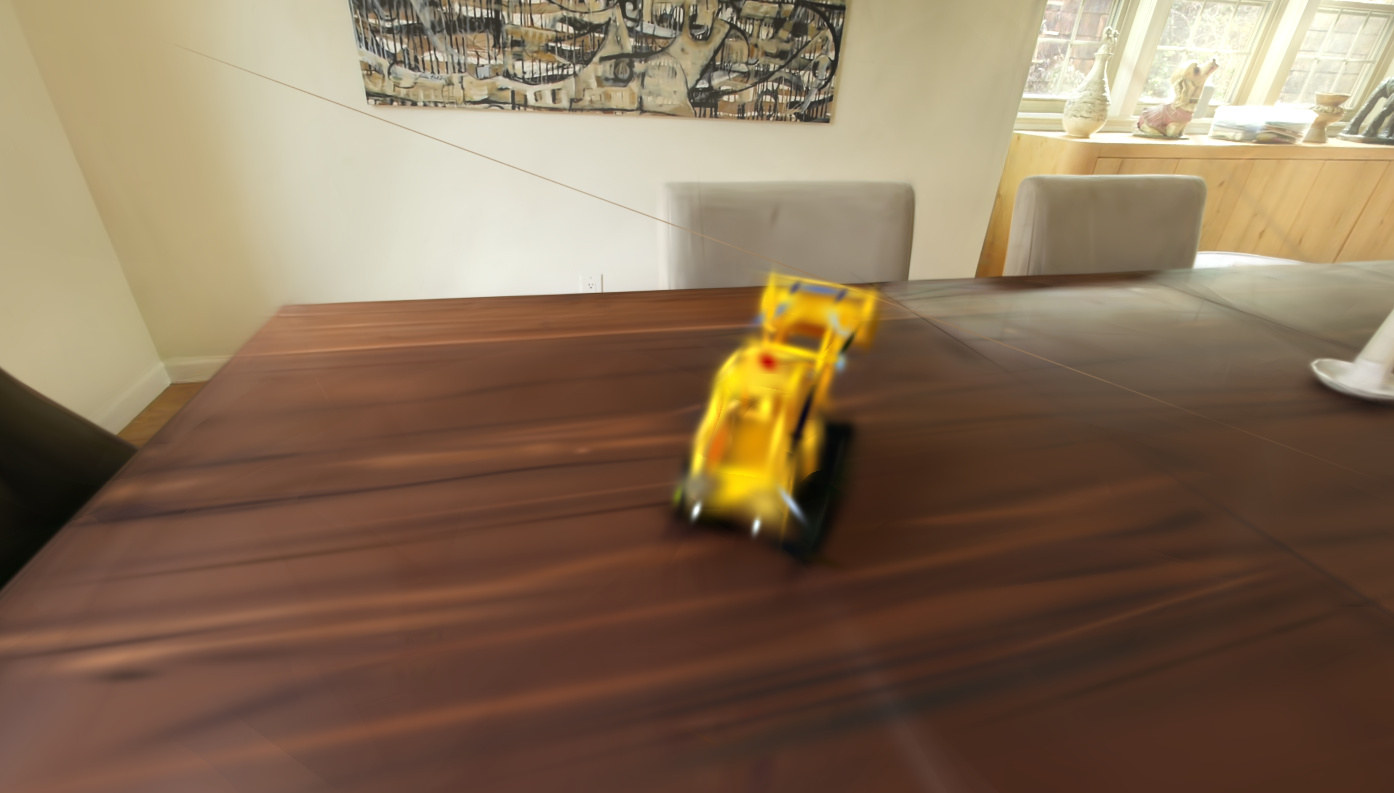} &
        \includegraphics[width=0.205\linewidth,trim=672 183 389 277,clip]{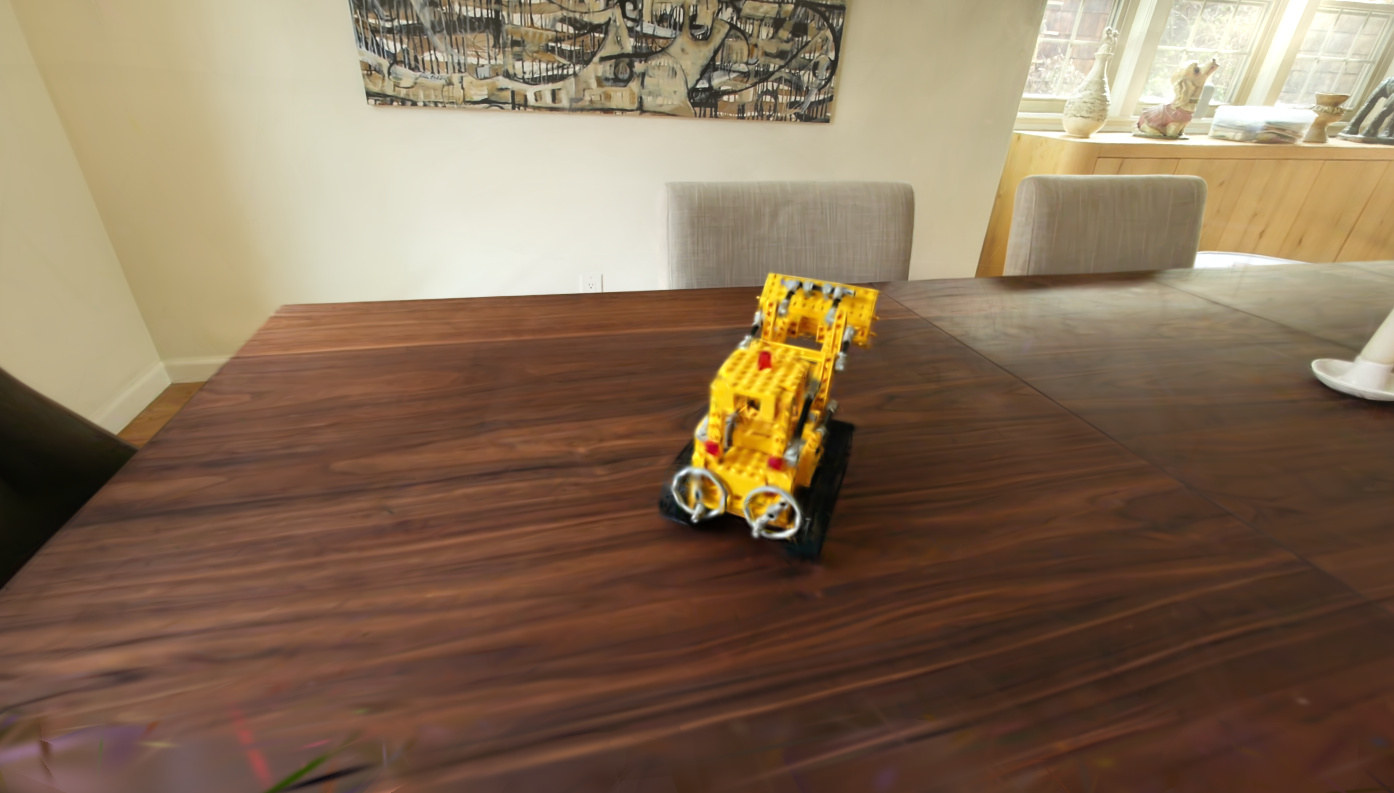} \\
    \end{tabular}
    \caption{Distance scaling (DS) suppresses near-object densification (alameda, $300$k schedule). On this close-up test view, PixelGS with DS ($6.1$M Gaussians) leaves the near object a featureless blur, whereas our average-contribution criterion ($3.1$M Gaussians) reconstructs the detail well.}
    \label{fig:exp:distancescale}
\end{figure}

The distance-scaling guard rail of PixelGS~\cite{zhang2024pixelgs} suppresses the densification gradient of near Gaussians quadratically with depth, scaling it by $\min\!\big((z/(\beta E))^2, 1\big)$ with camera-space depth $z$, constant $\beta$, and scene extent $E$; the wide Zip-NeRF camera orbits inflate $E$, so close-up detail is systematically under-densified and left blurry (\cref{fig:exp:distancescale})---the same weak scene-extent estimate we argue against for cloning and pruning. 
Our criterion requires no such term: unlike a pixel count, its per-view mean contribution is independent of the on-screen footprint, so near views are never up-weighted in the first place, and near and far balance without a distance term. 
The pixel-count criterion thus over-densifies without the guard (\cref{fig:exp:largescale-count}) and suppresses fine detail with it, whereas ours is stable and detail-preserving with no guard at all (per-view breakdown in \cref{sec:appendix:perview}). 

\subsection{Per-View Analysis} \label{sec:appendix:perview}
\begin{figure}[t]
    \centering
    \setlength{\tabcolsep}{1pt}
    \begin{tabular}{ccc}
        \scriptsize Ground truth & \scriptsize PixelGS + DS & \scriptsize Ours \\
        \includegraphics[width=0.325\linewidth]{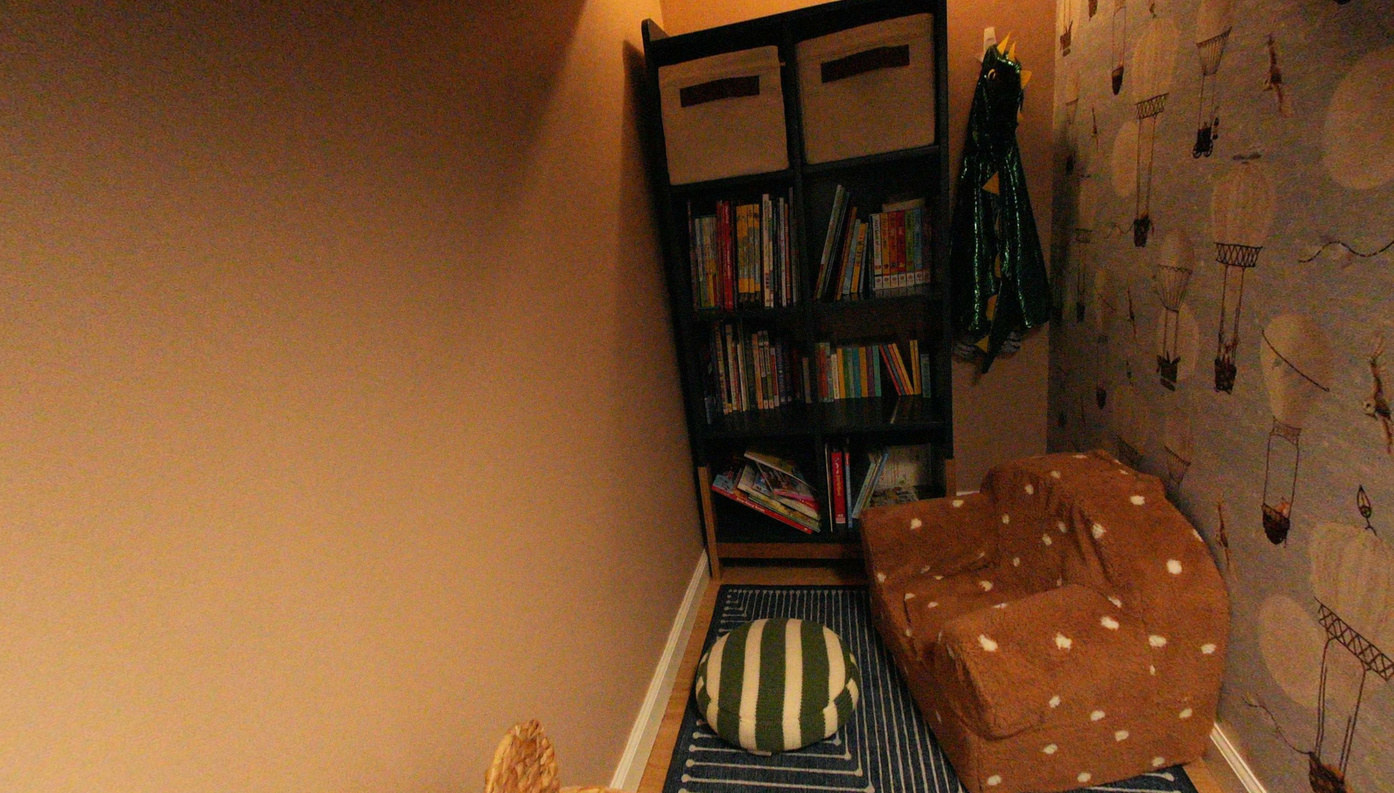} &
        \includegraphics[width=0.325\linewidth]{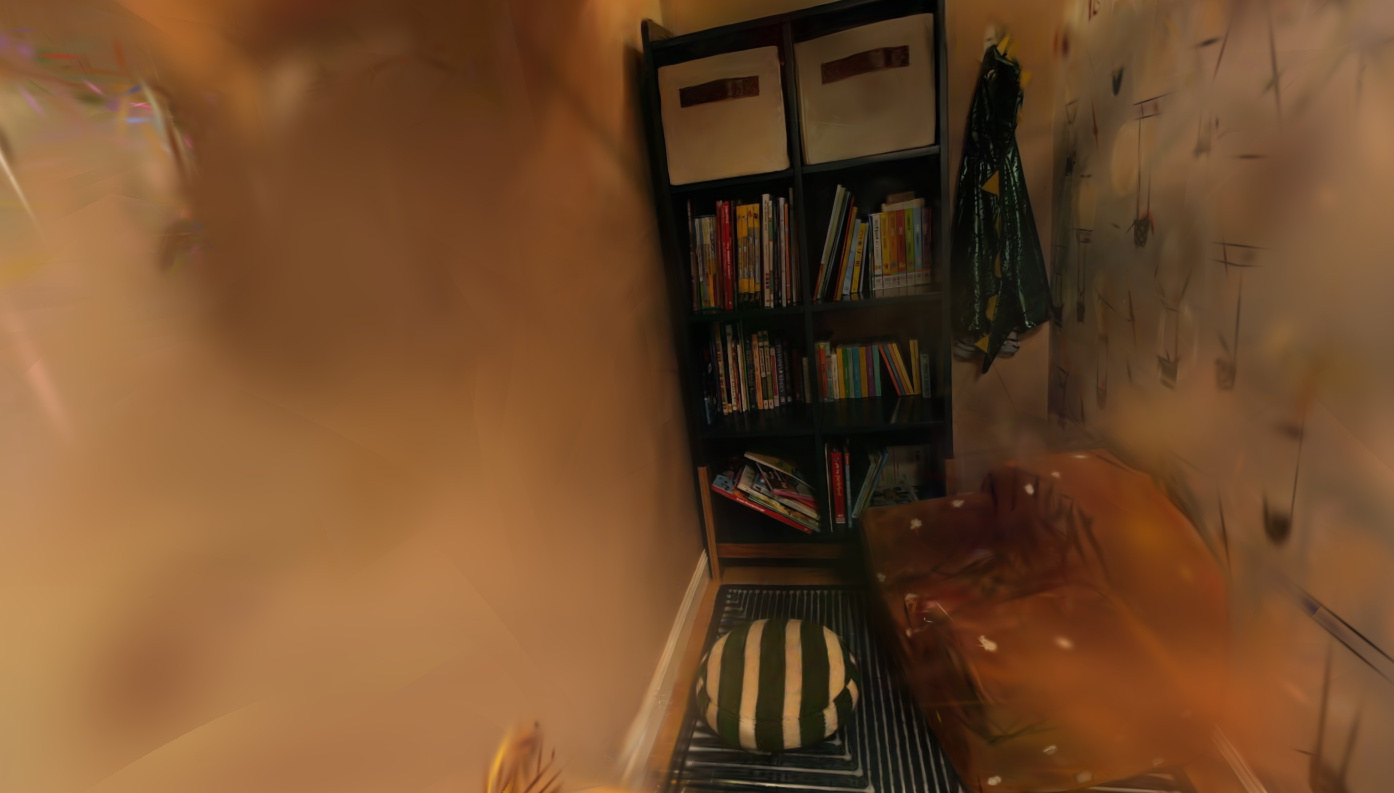} &
        \includegraphics[width=0.325\linewidth]{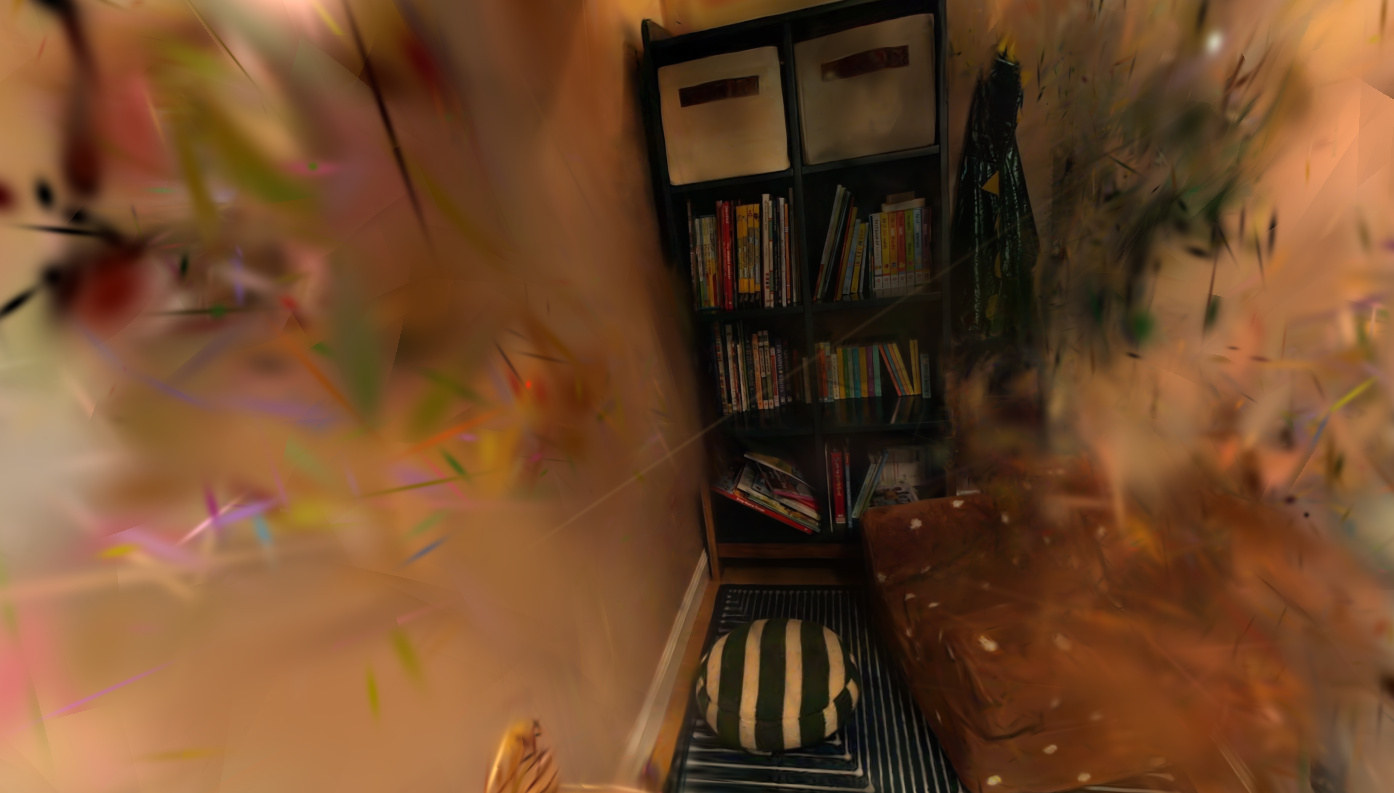} \\
    \end{tabular}
    \caption{An under-observed corner view (alameda test view 00070, $300$k schedule). The region is not intrinsically difficult but instead under-observed. The distance-scaled PixelGS renders an inoffensive smooth blur. Ours densifies to reconstruct the pixel-level color artifacts present in the ground truth, which it renders as colorful speckle.}
    \label{fig:appendix:perview-corner}
\end{figure}

Averaged over full test sets the distance-scaling asymmetry appears modest (Tab.~2 of the main paper) because it is spatially concentrated: 
on alameda at $210$k, ours improves LPIPS over PixelGS + DS on $133$ of $217$ test views by more than $0.005$, with the largest margins on the near-object close-ups (up to $0.10$ LPIPS and $2.4$\,dB better on the Lego toy).
The few views favoring PixelGS + DS by that margin ($27$ of $217$, not surpassing $0.05$ LPIPS) are severely under-observed corners, where its near suppression yields an inoffensive smooth blur, while the densification ours enables appears as floater speckle (\cref{fig:appendix:perview-corner}). 
We note that the ground truth images show pixel-level color artifacts, which explain the colorful speckles. 
Ours attempts to reconstruct this although it might be impossible to reconstruct this high-frequency noise that also changes between different viewpoints. 
Here, PixelGS is rewarded for its blurred reconstruction. 
Our criterion's lowest-quality views are intrinsically difficult views on which every method degrades together: 
on Mip-NeRF360 only a single test view falls below $18$\,dB and none exceed $0.45$ LPIPS. 
At the default schedules all three criteria place nearly the same number of views below $18$\,dB (one on Mip-NeRF360; $17$--$18$ on Zip-NeRF~\cite{barron2023zipnerf} at $30$k). 
At $210$k the occlusion-aware count instead collapses to a cluster reaching $12.5$\,dB on berlin and london, where ours leaves only a single view below $18$\,dB (minimum $16.0$\,dB). 
The $30$k per-view statistics are averaged per view over the three repetitions, while the extended-schedule statistics come from single training runs.

\paragraph{Per-view behavior.} 
Our criterion also avoids method-specific worst-case failures, whereas the count-based criteria's over-densification produces isolated severe failures (\eg $12$--$15$\,dB view clusters on london at $210$k where ours holds $27$--$33$\,dB).

\begin{figure}[t]
    \centering
    \includegraphics[width=0.80\linewidth]{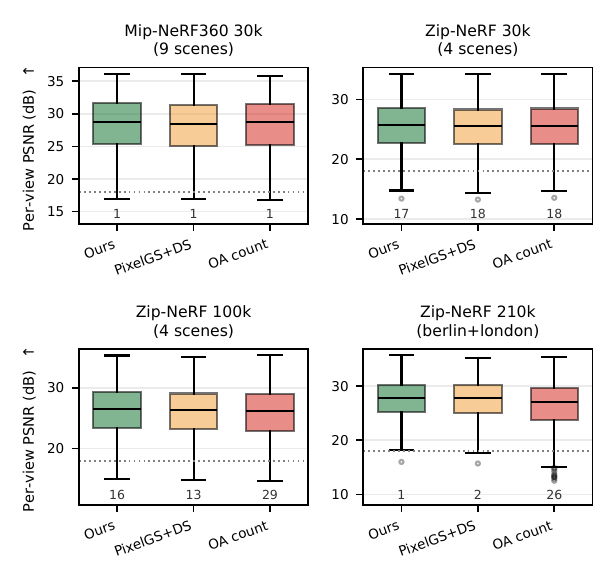}
    \caption{Per-view PSNR distribution per criterion across four training schedules. Numbers below each box count the views under $18$\,dB. The $30$k results are three-run averages; the extended schedules are single runs. The $210$k panel restricts all criteria to berlin and london, the two scenes on which all criteria complete.}
    \label{fig:robustness-boxplot}
\end{figure}

At the default schedules the three criteria's per-view PSNR distributions are statistically indistinguishable (\cref{fig:robustness-boxplot}). 
The divergence is driven by schedule length: 
as training lengthens the unguarded occlusion-aware count develops a growing low-quality tail ($29$ sub-$18$\,dB views at $100$k against $16$ views for ours), whereas ours stays as bounded as the distance-scaled PixelGS~\cite{zhang2024pixelgs} without needing that guard and with fewer Gaussians.

\section{Additional Qualitative Results} \label{sec:appendix:qualitative}
\begin{figure}[t]
    \centering
    \setlength{\tabcolsep}{1pt}
    \begin{tabular}{ccc}
        \scriptsize Ground truth & \scriptsize StopThePop & \scriptsize Ours \\
        \teaserbox{teasergreen}{\includegraphics[width=\dimexpr0.325\linewidth-2\tbrule\relax]{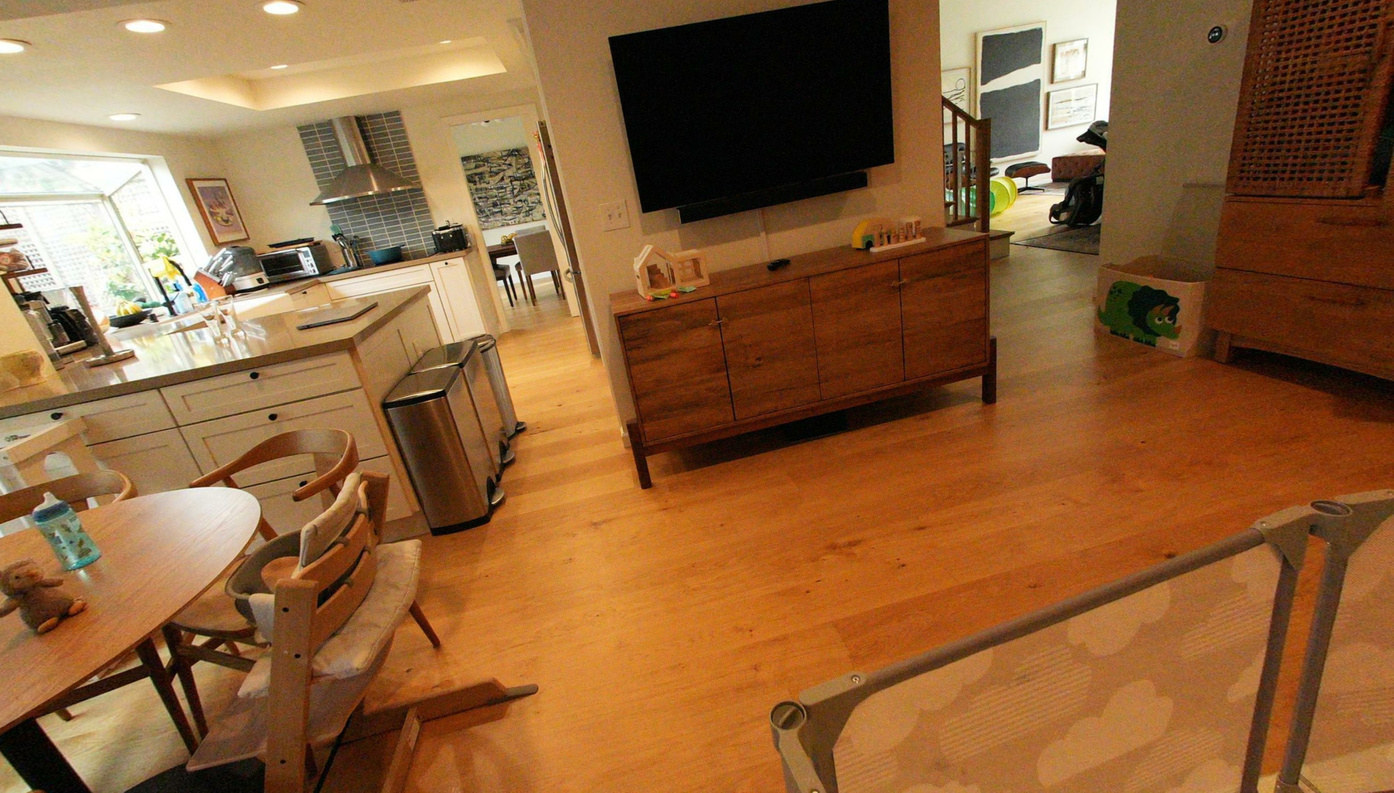}} &
        \teaserbox{teasergreen}{\includegraphics[width=\dimexpr0.325\linewidth-2\tbrule\relax]{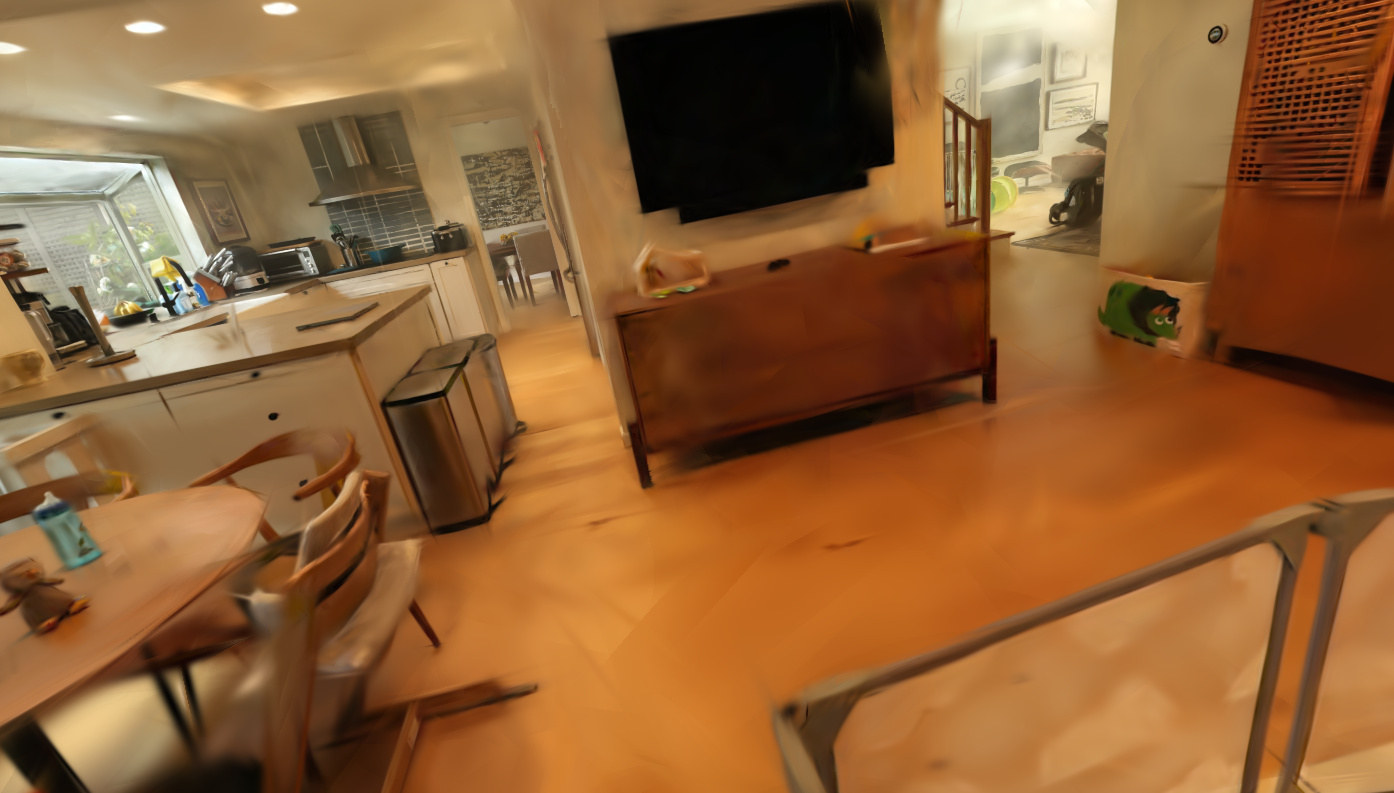}} &
        \teaserbox{teasergreen}{\includegraphics[width=\dimexpr0.325\linewidth-2\tbrule\relax]{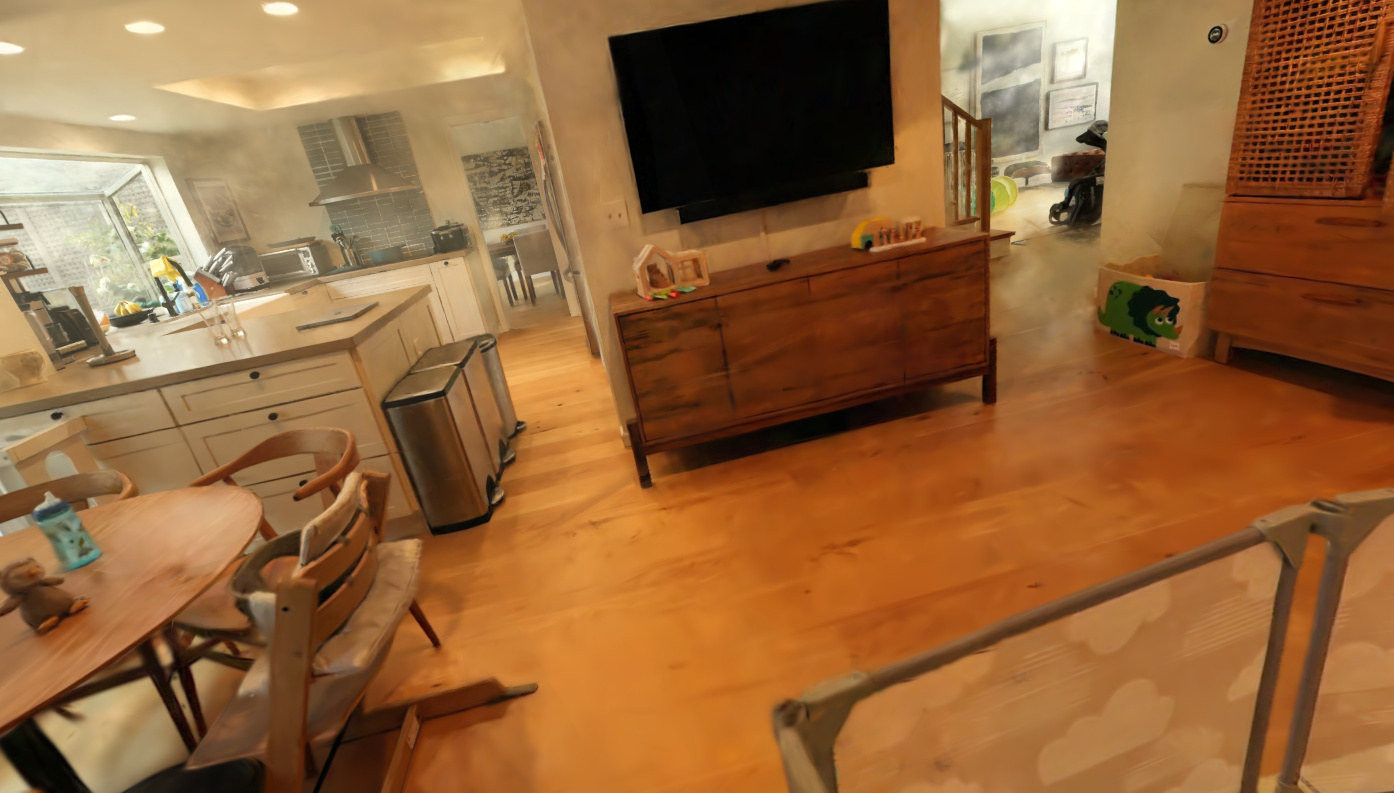}} \\
        \teaserbox{teaserblue}{\includegraphics[width=\dimexpr0.325\linewidth-2\tbrule\relax]{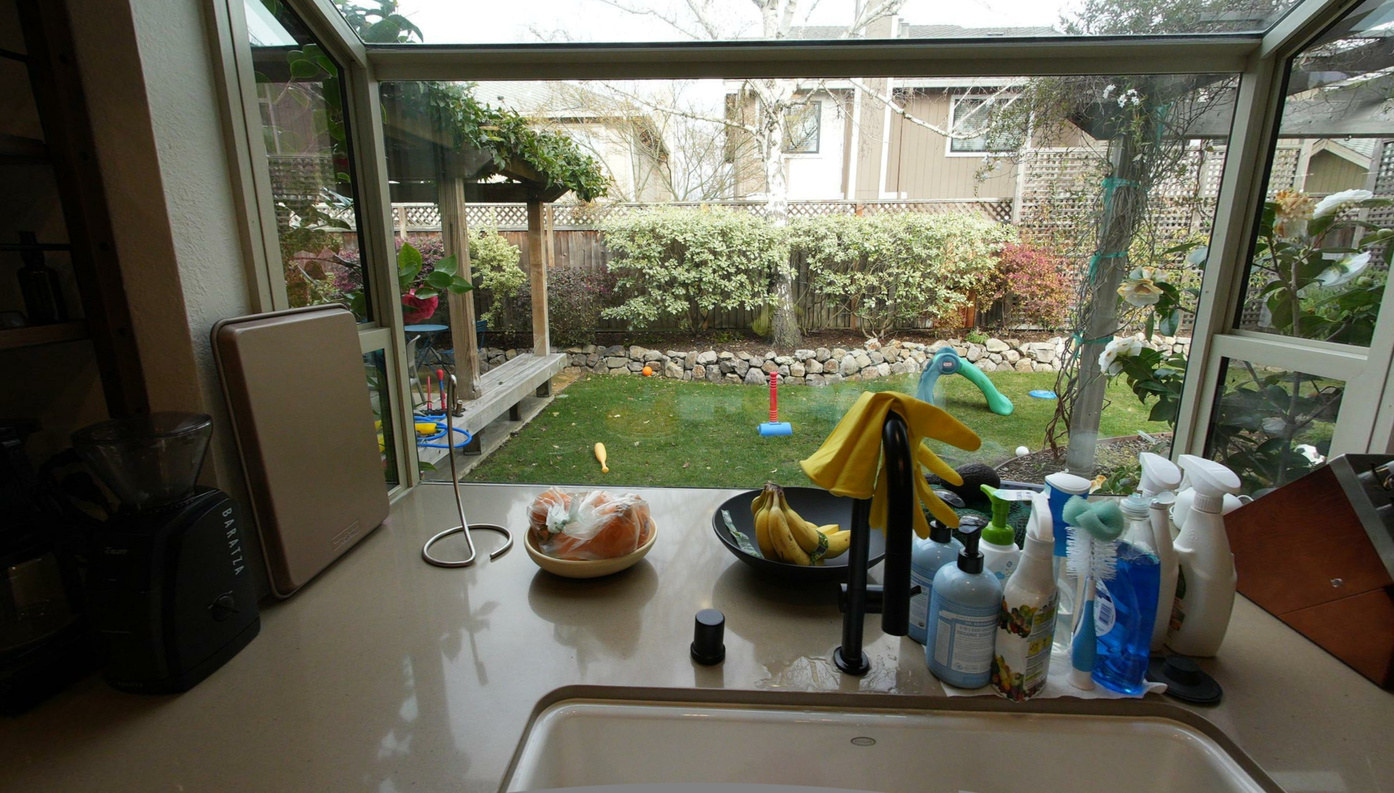}} &
        \teaserbox{teaserblue}{\includegraphics[width=\dimexpr0.325\linewidth-2\tbrule\relax]{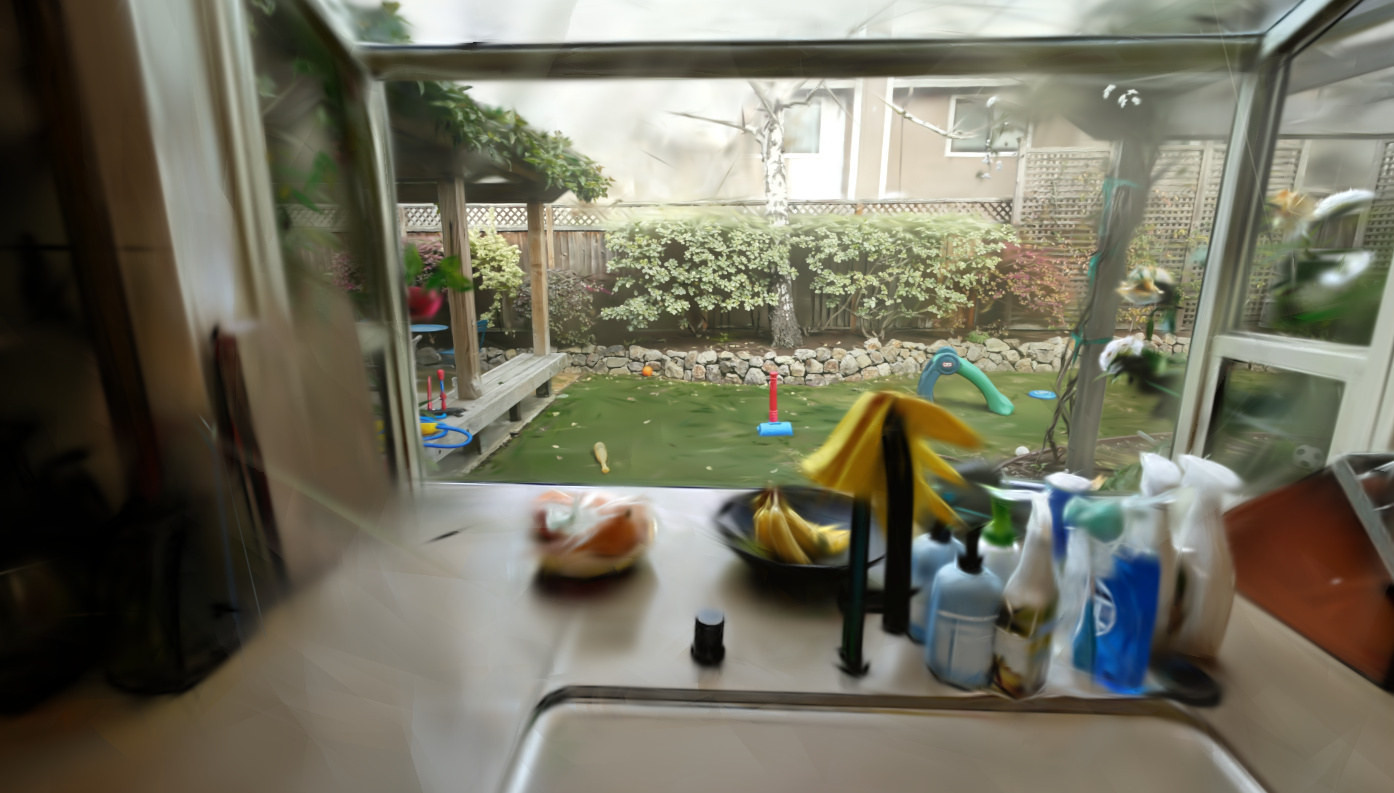}} &
        \teaserbox{teaserblue}{\includegraphics[width=\dimexpr0.325\linewidth-2\tbrule\relax]{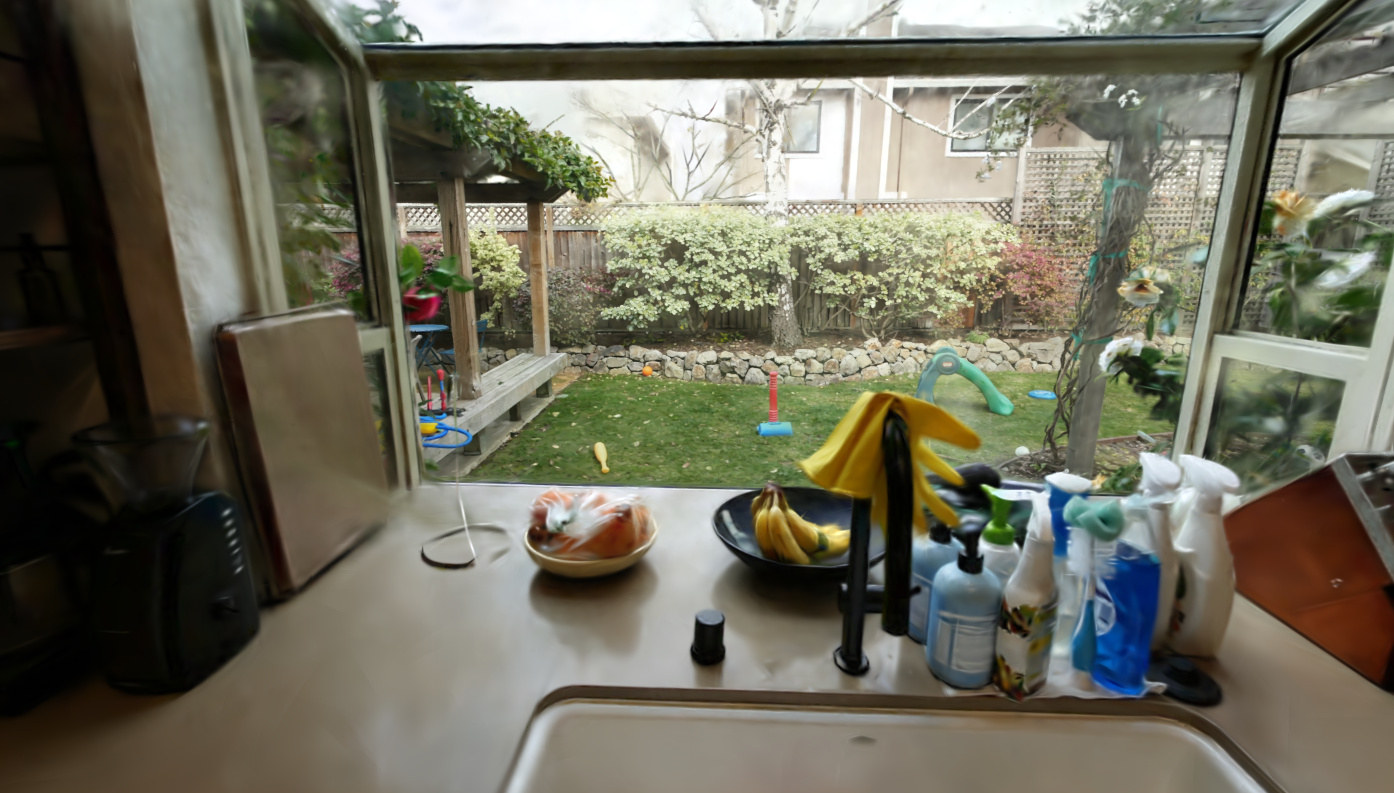}} \\
        \teaserbox{teaserred}{\includegraphics[width=\dimexpr0.325\linewidth-2\tbrule\relax]{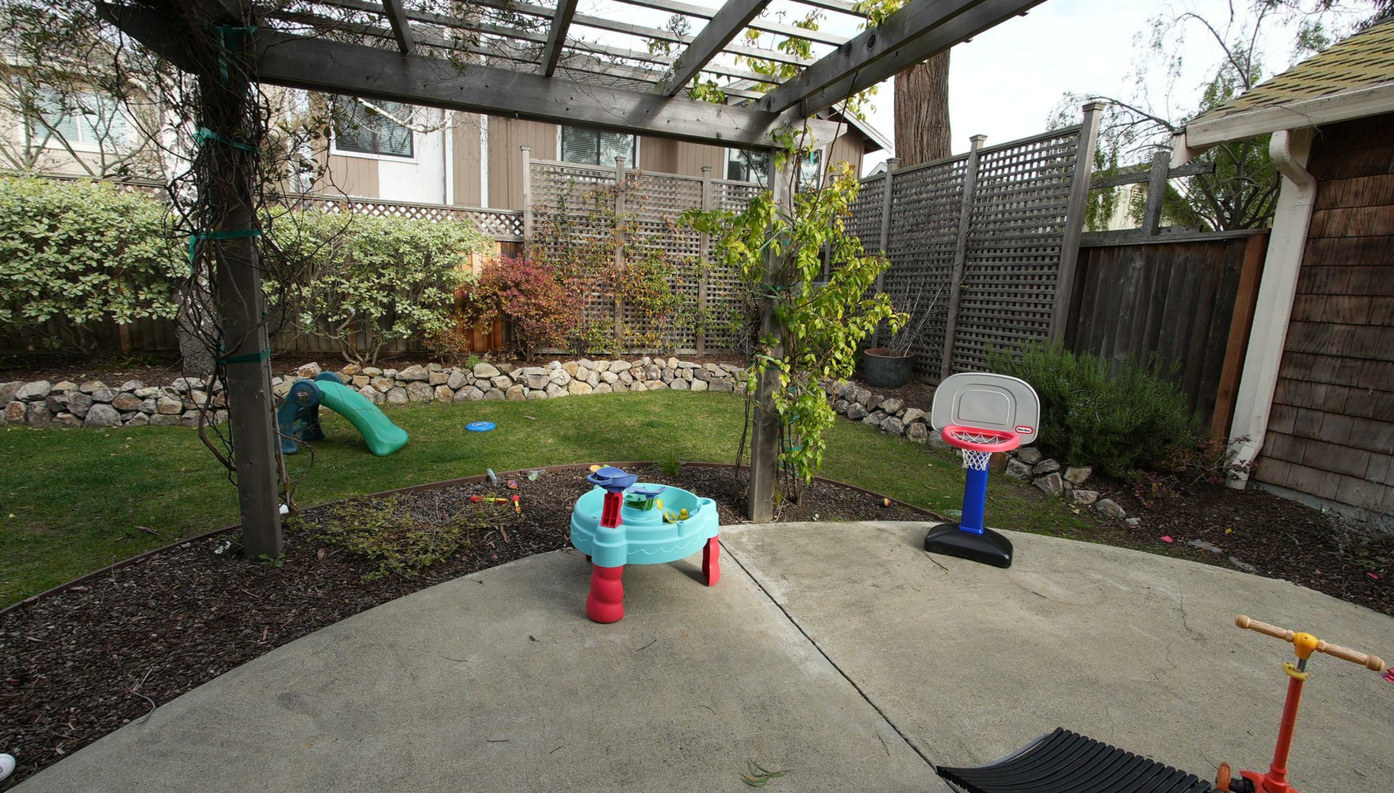}} &
        \teaserbox{teaserred}{\includegraphics[width=\dimexpr0.325\linewidth-2\tbrule\relax]{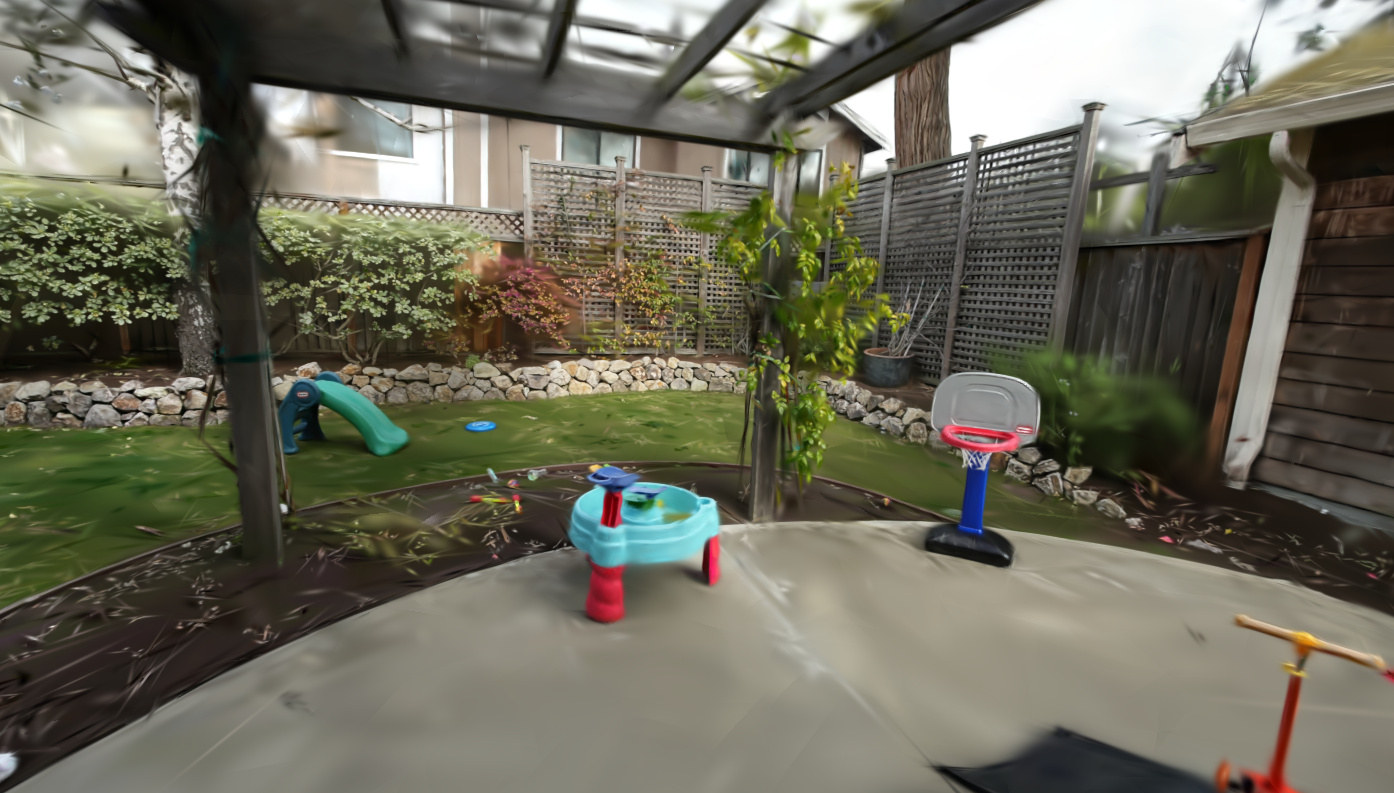}} &
        \teaserbox{teaserred}{\includegraphics[width=\dimexpr0.325\linewidth-2\tbrule\relax]{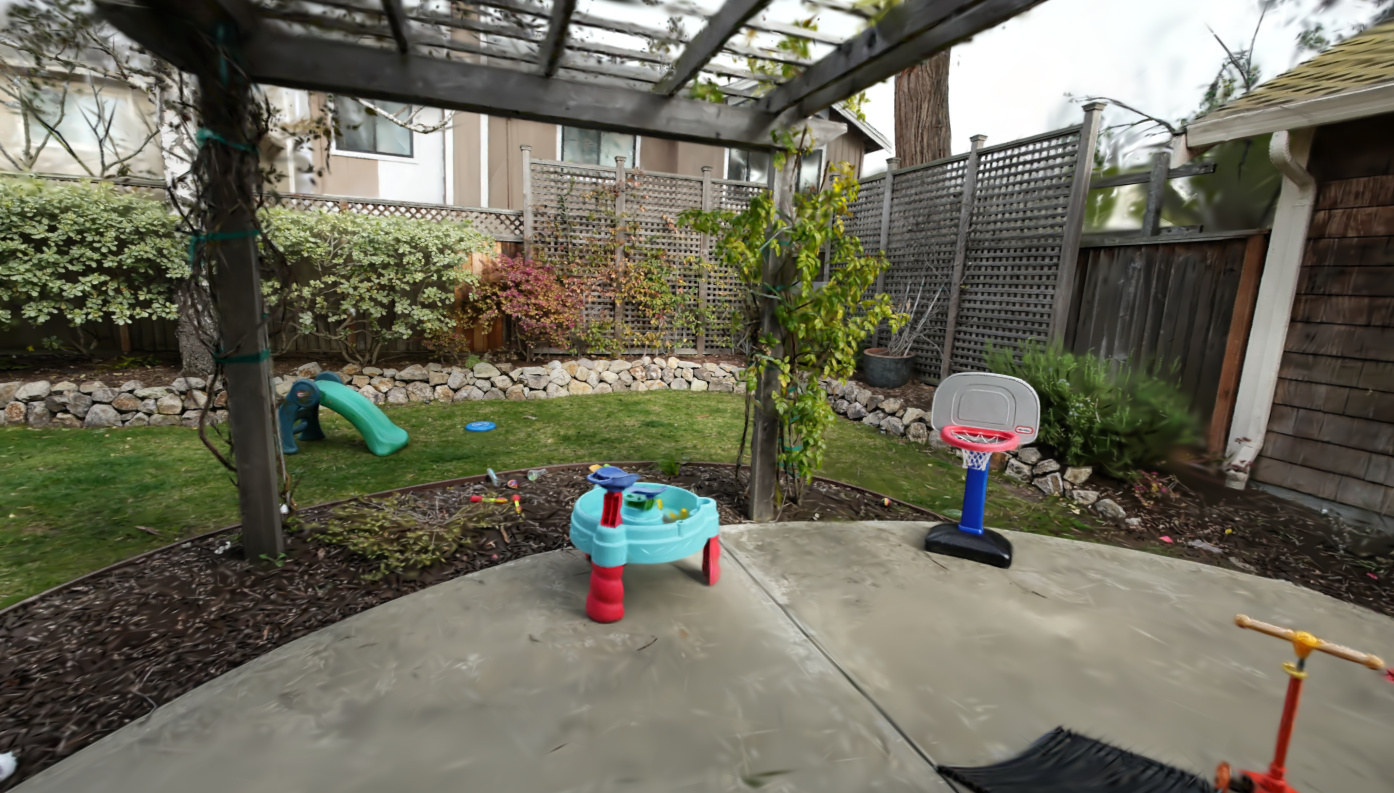}} \\
    \end{tabular}
    \caption{The three teaser locations of the main paper (Fig.~1) as full test frames (ground truth, StopThePop, ours). Frame colors match the location boxes of the teaser.} 
    \label{fig:teaser-fullframe}
\end{figure}

\Cref{fig:teaser-fullframe} shows the three teaser locations of the main paper as full test frames.

\begin{figure}[t]
    \centering
    \scriptsize
    \setlength{\tabcolsep}{0pt}
    \begin{tabular}{@{}c@{}}
        \makebox[0.2\linewidth]{Ground truth}\makebox[0.2\linewidth]{PixelGS}\makebox[0.2\linewidth]{Ours}\makebox[0.2\linewidth]{PixelGS ($1-$GMS)}\makebox[0.2\linewidth]{Ours ($1-$GMS)} \\
        \makebox[\linewidth][l]{berlin $00185$ --- largest LPIPS gain; GMSD $0.127 \rightarrow 0.088$} \\
        \includegraphics[width=\linewidth]{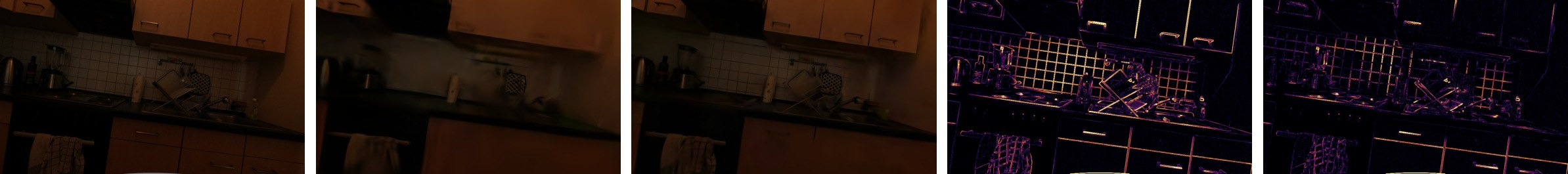} \\[5pt]
        \makebox[\linewidth][l]{berlin $00107$ --- smallest LPIPS gain; GMSD $0.056 \rightarrow 0.059$} \\
        \includegraphics[width=\linewidth]{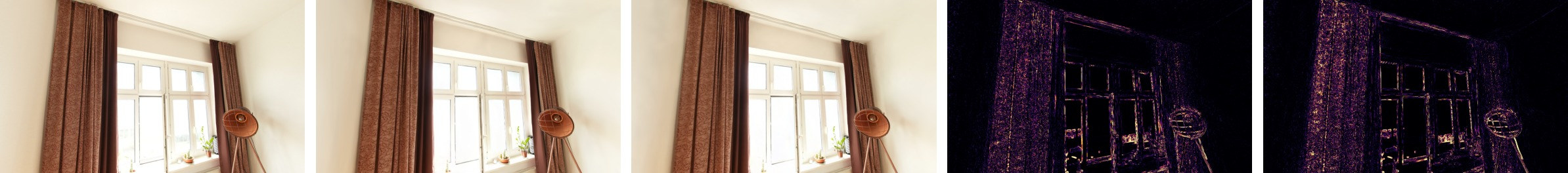} \\[5pt]
        \makebox[\linewidth][l]{london $00184$ --- largest LPIPS gain; GMSD $0.178 \rightarrow 0.129$} \\
        \includegraphics[width=\linewidth]{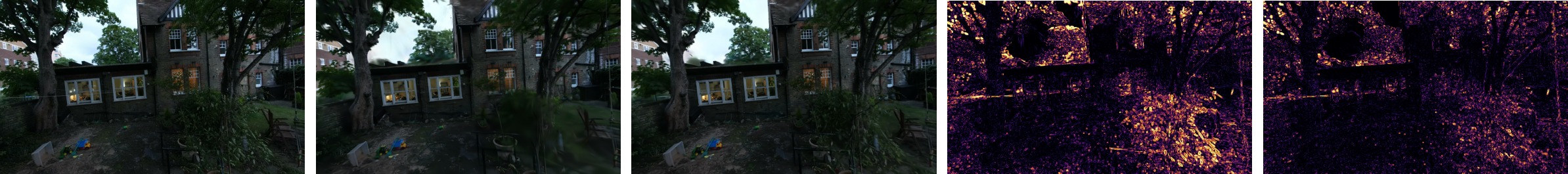} \\[5pt]
        \makebox[\linewidth][l]{london $00195$ --- smallest LPIPS gain; GMSD $0.134 \rightarrow 0.134$} \\
        \includegraphics[width=\linewidth]{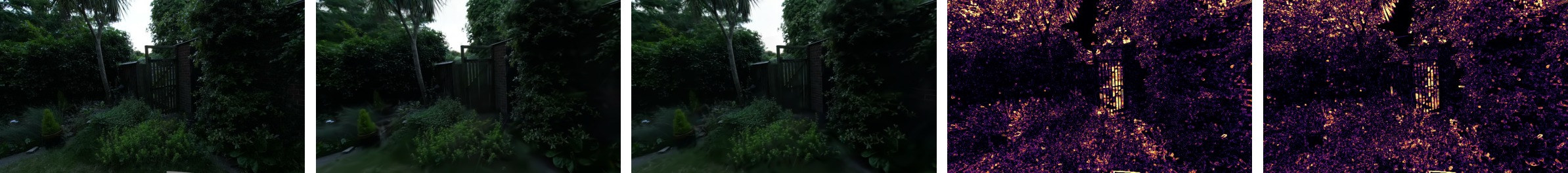} \\[5pt]
        \makebox[\linewidth][l]{nyc $00011$ --- largest LPIPS gain; GMSD $0.159 \rightarrow 0.103$} \\
        \includegraphics[width=\linewidth]{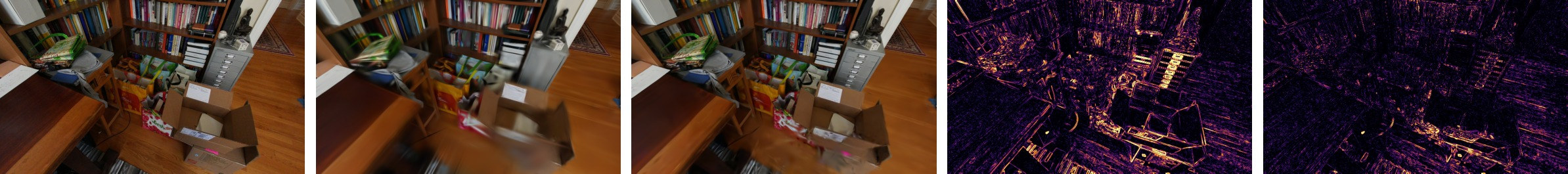} \\[5pt]
        \makebox[\linewidth][l]{nyc $00022$ --- smallest LPIPS gain; GMSD $0.047 \rightarrow 0.044$} \\
        \includegraphics[width=\linewidth]{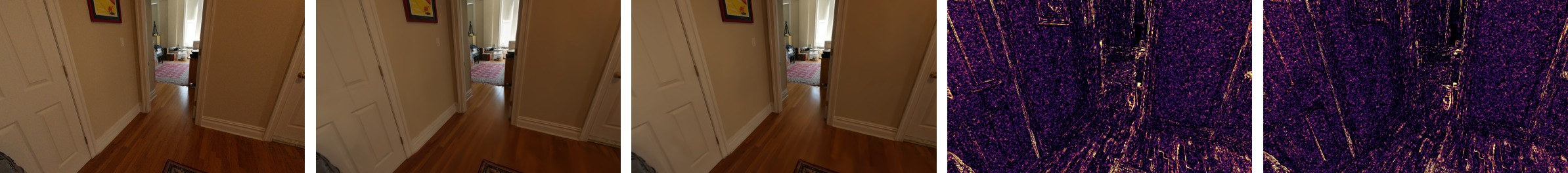} \\
    \end{tabular}
    \caption{Full-frame renders and error maps against vanilla PixelGS~\cite{zhang2024pixelgs} on three Zip-NeRF scenes ($30$k, single runs). Columns: ground truth, PixelGS, ours, and the gradient-magnitude-similarity error map ($1-$GMS, brighter is worse) of each. For every scene, the first row is the test view on which our LPIPS advantage over PixelGS is largest and the second the view on which it is smallest. Row labels give the per-view GMSD as PixelGS $\rightarrow$ ours.}
    \label{fig:appendix:errormaps}
\end{figure}

We show full-frame renders and error maps against vanilla PixelGS~\cite{zhang2024pixelgs} (StopThePop base) on three Zip-NeRF scenes in \cref{fig:appendix:errormaps}. 
We visualize gradient-magnitude similarity ($1-\text{GMS}$) rather than absolute error, because absolute error barely separates the two methods: 
the main failure is blur caused by insufficient reconstruction, which is favored by per-pixel metrics. 
Gradient-magnitude similarity instead responds to lost structure, which is where the two methods differ. 
For each scene we show the test views on which our LPIPS advantage over PixelGS is largest and smallest, respectively. 
Where the advantage is largest, PixelGS visibly loses fine structure such as tiles, foliage, and clutter. 
Its error map is correspondingly brighter: 
its GMSD is $0.127$, $0.178$, and $0.159$ against our $0.088$, $0.129$, and $0.103$. 
Where the advantage is smallest, the two error maps are nearly indistinguishable. 
On nyc our LPIPS is in fact better on every test view. 
Our weakest views relative to PixelGS are therefore not views where it recovers more detail, but views where the two reconstructions are structurally equivalent.

\bibliographystyle{splncs04}
\bibliography{main}

\end{document}